\pdfoutput=1
\documentclass[11pt]{article}

\usepackage[preprint]{acl}

\usepackage{times}
\usepackage{latexsym}
\usepackage{booktabs}
\usepackage{threeparttable}
\usepackage{hyperref}

\usepackage[T1]{fontenc}
\usepackage{amssymb}
\usepackage{paralist}
\usepackage{amsmath} 
\usepackage{adjustbox}
\usepackage[T1]{fontenc}
\usepackage{float}
\usepackage{soul}

\usepackage[utf8]{inputenc}
\usepackage{multirow}

\usepackage{microtype}
\usepackage{enumitem}

\usepackage{inconsolata}

\usepackage{graphicx}
\usepackage{subcaption} 

\usepackage[framemethod=TikZ]{mdframed}

\newmdenv[
  roundcorner=8pt,
  linecolor=black,
  linewidth=0.8pt,
  innertopmargin=6pt,
  innerbottommargin=6pt,
  innerleftmargin=6pt,
  innerrightmargin=6pt
]{RoundedBox}

\title{Where to show Demos in Your Prompt:\\A Positional Bias of In-Context Learning}

\author{Kwesi Cobbina \and Tianyi Zhou \\
        University of Maryland, College Park\\
        \texttt{kcobbina@umd.edu, tianyi.david.zhou@gmail.com}\\
        }

\begin{document}
\maketitle
\captionsetup{font=small}
\begin{abstract}

In-context learning (ICL) is a critical emerging capability of large language models (LLMs), enabling few-shot learning during inference by including a few demonstrations (demos) in the prompt. 
However, it has been found that ICL's performance can be sensitive to the choices of demos and their order.
This paper investigates an unexplored new positional bias of ICL for the first time: we observe that the predictions and accuracy can drift drastically when the positions of demos, system prompt, and user message in LLM input are varied. This bias, we refer to as \textsc{Demos' Position in Prompt} bias (\texttt{DPP} bias). We design a systematic evaluation pipeline to study this type of positional bias across classification, QA, summarization, and reasoning tasks. We introduce two metrics, \textsc{accuracy-change} and \textsc{prediction-change}, to quantify net gains and output volatility induced by demos' position change. Extensive experiments on ten LLMs from four open-source model families (\textsc{Qwen}, \textsc{Llama3}, \textsc{Mistral}, \textsc{Cohere}) verify that the bias significantly affects their accuracy and predictions: 
placing demos at the start of prompt yields the most stable and accurate outputs with gains of up to +6 points. In contrast, placing demos at the end of the user message flips over 30\% of predictions without improving correctness in QA tasks. Smaller models are most affected by this sensitivity, though even large models do remain marginally affected on complex tasks.

\end{abstract}

\section{Introduction}
\label{sec:intro}
The rapid evolution of large language models (LLMs) has redefined the boundaries of machine learning, enabling unprecedented few-shot and zero-shot generalization across tasks like classification, question answering, and summarization (\citealp{NEURIPS2020_1457c0d6, radford2019language}). Central 

\begin{figure}[!ht]
\centering
  \includegraphics[width=\linewidth]{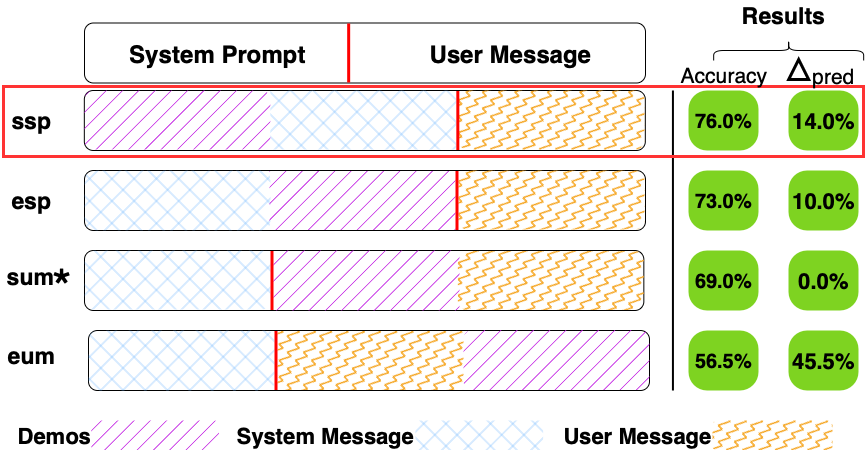}
  \caption{\textbf{Four configurations of demos' position in prompt (\texttt{DPP})} from \autoref{sec:methodology}: \textit{ssp} (Start of System Prompt), \textit{esp} (End of System Prompt), \textit{sum}
  (Start of User Message, \textbf{default}), and \textit{eum} (End of User Message). 
  Their results with \textsc{Qwen-1.5B} on \texttt{AG news} datasets are reported on the right: Their accuracies vary drastically and the percentage of changed predictions (compared to default \textit{sum}) can be up to \textit{45.5\%}.} 
  \label{fig:promptlayouts}
\end{figure}

\noindent to this paradigm shift is in-context learning (ICL), where models dynamically adapt to new tasks by processing demos embedded directly in the input prompt.
Recent work has exposed critical vulnerabilities: minor perturbations to demo ordering or demo count \cite{lu-etal-2022-fantastically} can degrade performance unpredictably. This brittleness not only undermines reproducibility but also challenges assumptions about LLMs’ capacity for systematic reasoning, raising urgent questions about whether current models truly learn from context or merely exploit superficial patterns.

We discover a \textbf{novel positional bias} in \emph{in-context learning} (ICL): \texttt{DPP} bias, in which moving an \emph{unchanged} block of demos from the start of a prompt to the end can swing task accuracy by \textbf{up to 20~percents} and flip almost half of a model’s predictions (see Fig.~\ref{fig:promptlayouts}).  
This phenomenon, purely spatial, independent of demo content, challenges the widespread assumption that large language models learn robustly from any properly formatted context.

Despite growing awareness of prompt sensitivity, the role of demo positioning where demos are placed relative to instructions, queries, or other contextual elements remains underexplored. Prior studies have focused primarily on demo selection \cite{liu-etal-2022-makes}, or template phrasing (\citealp{cho-etal-2024-tutor, voronov-etal-2024-mind}), leaving a gap in understanding how spatial arrangements modulate ICL efficacy. This paper addresses this gap through a systematic investigation of positional effects across eight tasks spanning classification, reasoning, and generation. By conducting controlled studies on models like \textsc{Llama-3} (1B, 3B, 8B, 70B) and \textsc{Mixtral\_8x7B}, we demonstrate that strategic placement (e.g., clustering critical demos near task instructions) can yield performance swings, even when demo content remains identical.

Our work makes \textbf{five complementary contributions}. \textbf{1. }We first uncover and quantify a previously unreported positional bias (\texttt{DPP} bias) in in-context learning, showing that simply relocating an identical demo block within the prompt can shift accuracy by up to 50 percentage points while flipping nearly half of a model’s predictions. \textbf{2. }Building on this insight, we design a controlled evaluation pipeline that isolates four canonical demo placements, at the start or end of the system prompt and at the start or end of the user message, so that any performance change is attributable purely to position. \textbf{3. }To capture both net performance shifts and output volatility, we introduce two task-agnostic metrics, \emph{accuracy-change} and \emph{prediction-change}. Using this framework, \textbf{4. }we conduct the first large-scale empirical study of positional effects across eight tasks and ten state-of-the-art LLMs, revealing a consistent primacy bias that becomes less severe as model size grows. \textbf{5. }Finally, we translate these findings into practical guidelines.

\section{Related Work}
\label{sec:related_works}
In this section, we review existing literature on positional biases in in-context learning (ICL). We organize the discussion into three subsections: internal demo-order bias, mechanistic hypothesis, and the role level gap spatial placement.

\subsection{Internal Demonstration-Order Bias}
Prompt-order sensitivity is a well-established phenomenon in in-context learning (ICL). \citet{lu-etal-2022-fantastically} demonstrated that merely permuting the order of demonstrations can lead to accuracy fluctuations of approximately ±15\% in reasoning tasks, such as arithmetic and commonsense question-answering. Similarly, \citet{min-etal-2022-rethinking} found that large language models (LLMs) frequently exploit superficial lexical overlaps between demonstrations and queries rather than learning robust semantic mappings. \citet{zhao2021calibrateuseimprovingfewshot} further showed that demonstration order significantly impacts few-shot outcomes and this was also supported by \citet{wang-etal-2023-primacy} who found that ChatGPT predominantly favors earlier listed labels in classification tasks, while \citet{10.5555/3600270.3602070} indicated that reasoning gains from Chain-of-Thought (CoT) rationales heavily depend on their positioning within prompts. These studies underscore the fragility of ICL to superficial prompt characteristics, motivating further exploration into position-related biases. \textit{Our study departs from these works by holding the \emph{internal} order fixed and relocating the entire demo block to different prompt regions.}


\subsection{Mechanistic Hypothesis}
Recent research attributes positional bias in transformer-based models to intrinsic architectural tendencies, notably primacy bias and induction heads. \citet{olsson2022incontextlearninginductionheads} and \citet{chan2022datadistributionalpropertiesdrive} highlight that transformers disproportionately emphasize early tokens due to induction head mechanisms, causing initial context to steer subsequent predictions significantly. Similarly, \citet{xiao2024efficientstreaminglanguagemodels} note sequential processing biases towards earlier context, which impact performance when crucial information appears later in the sequence. Additionally, \citet{liu2023lostmiddlelanguagemodels} observed that tokens in the middle positions of sequences receive less attention, leading to performance degradation. 


\noindent\citet{bietti2023birthtransformermemoryviewpoint} further supports this by linking primacy bias to underlying transformer memory mechanisms. While these hypotheses illuminate \emph{why} order matters, empirical work on how they interact with \emph{prompt roles} (system vs.\ user) is scarce.  
\textit{We provide the first role-aware stress test of these mechanisms.}


\subsection{Spatial Placement (Role-Level) Gap}
While prior ICL research extensively explores the selection of demonstrations, relatively little attention has been paid to their precise spatial placement within prompts. Studies such as \citet{cho-etal-2024-tutor}, \citet{reynolds2021promptprogramminglargelanguage}, and \citet{webson-pavlick-2022-prompt} prioritize choosing semantically relevant demonstrations and designing tailored prompt templates but overlook how the exact location of demonstration blocks, particularly relative to system and user roles, might independently affect model outcomes. \citet{beck-etal-2024-sensitivity} introduce the metrics of “sensitivity” (output flip rate) and “performance” (accuracy delta) when swapping sociodemographic personas in prompts—metrics formally equivalent to our \textbf{Prediction-$\Delta$} and \textbf{Accuracy-$\Delta$}.  However, their experiments hold the prompt’s structural layout constant and only vary \textit{which} persona is inserted, not \textit{where} the block appears. Our study addresses this gap by explicitly varying demonstration placement across prompt roles, highlighting an overlooked but critical dimension of prompt structuring for achieving reliable ICL performance.

\section{Methodology}
\label{sec:methodology}
We present a systematic framework to investigate how the position of in-context demos within a prompt affects model performance. Our approach formalizes the problem of \texttt{DPP} bias, defines the range of demonstration placements considered, and outlines an evaluation pipeline for measuring performance variations.

\subsection{Problem Formulation}
We focus on the classical in-context learning scenario, where an LLM is given a small set of demonstrations along with a query, all concatenated into a single prompt. Formally, let $\mathcal{T}$ be a set of tasks (e.g. sentiment classification, QA, etc.), and for each task $\tau \in \mathcal{T}$, let $D_{\tau}$ be a set of $N$ demonstrations and $Q_{\tau}$ a set of evaluation queries. For a given query $q \in Q_{\tau}$, we construct a prompt $P$ that combines some or all examples from $D_{\tau}$ with $q$. Crucially, our study keeps the content of $P$ (the instruction, the examples in $D_{\tau}$, and the query $q$) fixed, and manipulates only the structural position of the demonstration block within the prompt. We define positional bias (or spatial confounder effect) as any change in the model’s performance on the query set $Q_{\tau}$ that arises solely from where the demonstrations appear in $P$, rather than which demonstrations are provided. Essentially isolating how the different structural positions affect the model output.

\subsection{Demo Positions: Definitions}\label{meth:dpd}
In many recent instruction tuned LLMs, a prompt can include a system prompt, which is then followed by the user message (chat-style format). We leverage this structure to define four distinct canonical demonstration positions where a block of $k$ demos can be inserted in the prompt. These four configurations, illustrated in Figure~\ref{fig:promptlayouts} are defined as followed:

\begin{itemize}
    \item Start of System Prompt (\textit{ssp}): The demos block is placed at the very beginning of the system message, before any instructional content.
    \item End of System Prompt (\textit{esp}): The demos block is placed at the end of the system message, after any general instructions but still before the user’s query.
    \item Start of User Message (\textit{sum}): The demos block is inserted at the beginning of the user message, before the actual query text. 
    \item End of User Message (\textit{eum}): The demonstration block is appended at the very end of the user message, after the query.
\end{itemize}

Figure ~\ref{fig:promptlayouts} provides a schematic diagram of these four positions. It shows whether the demos reside in the system vs. user section of the prompt and whether they appear at the start or end. Intuitively, \textit{ssp} and \textit{esp} represent placing demonstrations before the user’s question, whereas \textit{sum} and \textit{eum} place them before and after the user’s question respectively.

\subsection{Evaluation Metrics}
We report the task‑specific metrics recommended by prior work: \textbf{Accuracy} for multiple‑choice (MCQ) problems, \textbf{F$_1$} and \textbf{Exact Match} for extractive question answering (QA), and \textbf{ROUGE}–L and \textbf{BERTScore} for summarization. Aside from the suggested metrics, to understand the per question by position transitions, we also report other metrics:

\paragraph{Accuracy Change.}
Accuracy Change $\Delta_{\mathrm{metric}}$ directly quantifies how adding demonstrations at a given position influences the model’s overall task performance relative to zero-shot. Formally,
\begin{equation}
    \Delta_{\text{\$metric}} = \text{Metric}_{\text{position}} - \text{Metric}_{\text{zero-shot}}
\end{equation}
 A positive $\Delta_{\mathrm{\$metric}}$\footnote{metric = Accuracy, Exact Match, ROUGE-L
} indicates that placing demos in that location helps the model make more correct predictions, while a negative value means the demonstrations actually degrade performance. By isolating the net gain or loss in accuracy, this metric cleanly attributes performance differences to spatial placement of the same content, enabling fair comparison across positions, models, and tasks.

\paragraph{Prediction Change.}
Prediction Change $\Delta_{\mathrm{pred}}$ measures the volatility of individual model outputs induced by demonstration placement. It is defined as

\begin{equation}
    \Delta_{\text{pred}} = \frac{\#\textit{answer flips}}{\#\mathcal{Q}}
\end{equation}

where $\# \mathcal{Q}$ is the total number of evaluation queries, and \#\textit{answer flips} counts all instances whose predicted outputs flips when going from the default ICL position(\textit{sum}) to the other in-context positions. High $\Delta_{\mathrm{pred}}$ reveals that demonstration placement strongly perturbs the model’s decision boundary, even if net metric gains are small.



\paragraph{Datasets with free-form answers.}
For free-form outputs, we treat two answers as a \emph{prediction flip} when their faithfulness to the gold answer \(y\) diverges meaningfully. 
Concretely, we compute ROUGE-L scores for answers generated from the \texttt{sum} and \texttt{esp} positions and declare a flip if
\[
\bigl|\text{ROUGE-L}(y_{\texttt{sum}},\,y) - \text{ROUGE-L}(y_{\texttt{esp}},\,y)\bigr| > 0.05 .
\]
The same 0.05 threshold is used for XSUM and CNN/DailyMail throughout our experiments.

\noindent\textbf{Remarks~}
We propose a systematic framework to investigate how the structural position of in-context demonstrations affects LLM performance. Our study isolates positional effects by controlling for prompt content while varying the location of a fixed demonstration block. We define four canonical positions within a prompt, \textit{ssp}, \textit{esp}, \textit{sum}, and \textit{eum}, which differ in whether demos are placed within the system or user section, and whether they precede or follow the query. These positions are visualized in Figure~\ref{fig:promptlayouts}.\looseness-1


\section{Empirical Results}
\label{sec:results}
We evaluate how demonstration position affects model performance both in terms of net accuracy change relative to zero-shot, and in terms of answer volatility (prediction flips)

\subsection{Positional Bias across Tasks}
A consistent and pronounced pattern emerges across our benchmark datasets: demonstrations positioned at the beginning of prompts (\textit{ssp} or \textit{esp}) reliably outperform placements later in the prompt (\textit{eum}) and frequently surpass the default ICL position (\textit{sum}).
Throughout our experiments, we set the number of demos to five. 
We keep the demos in the demos block and identical across these conditions, so that any performance differences can be attributed purely to positional effects. (Any additional prompt formatting details and exact templates used for each position are provided in the Appendix. \autoref{apdx:pmtllms} \& \autoref{apdx:fp}) 

\begin{figure}[h]
    \centering
    \includegraphics[width=\linewidth]{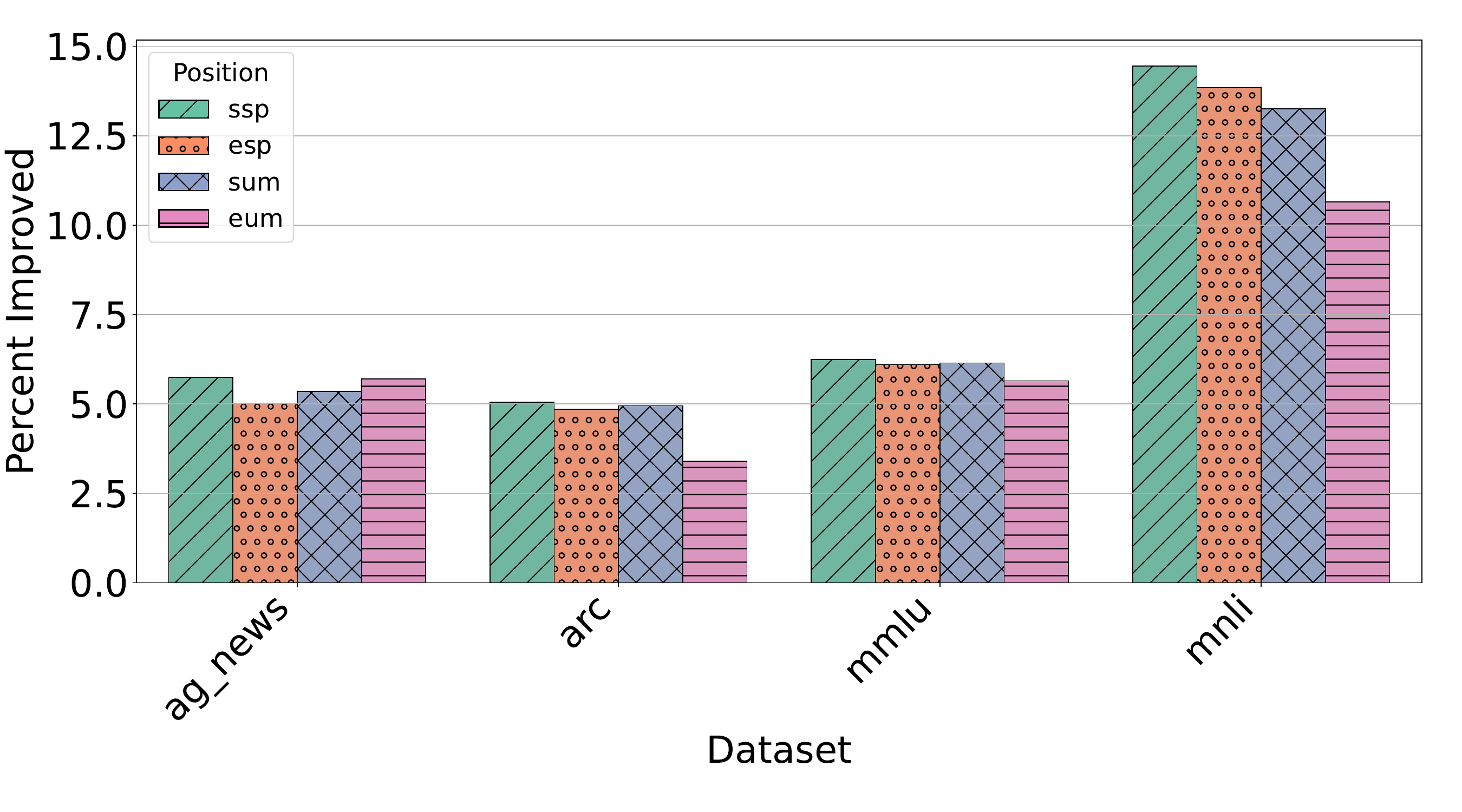}
    \vspace{-2.5em}
    \caption{
        \textbf{Accuracy change (comparing to zero-shot) of the four \texttt{DPP}s across four datasets, averaged over all models.} 
        The \texttt{ssp} achieves the greatest improvement over zero-shot across all four datasets (note the winner may vary for different models as shown in Fig.~\ref{fig:qween1d5bwl}-\ref{fig:lma70b}).
        }
        \vspace{-1em}
    \label{fig:accuracy-change-all}
\end{figure}

\begin{figure}[!ht]
    \centering
    \includegraphics[width=\linewidth]{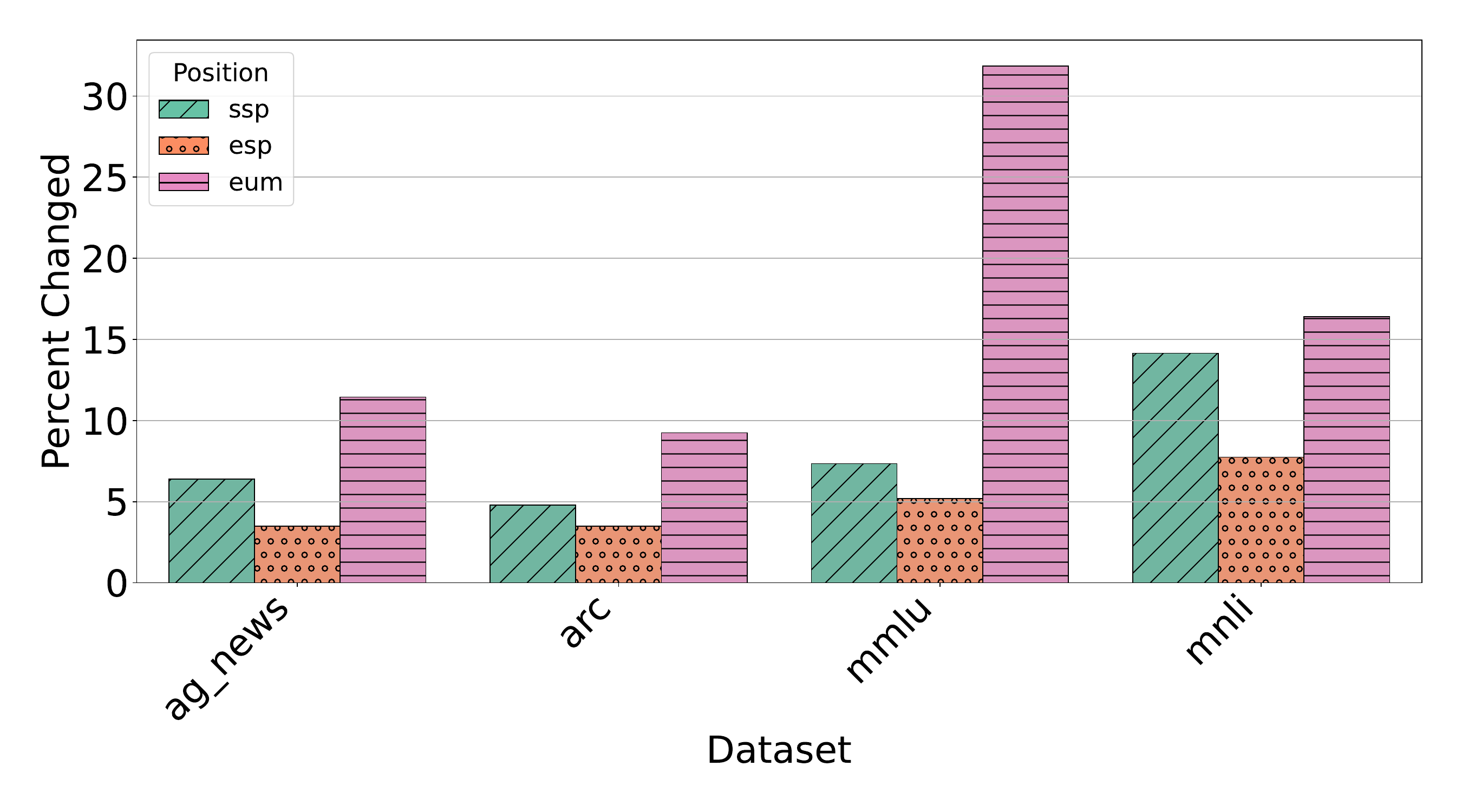}
    \vspace{-2.5em}
    \caption{
        \textbf{Prediction change (comparing to \texttt{sum}) ratios of the three \texttt{DPP}s (excluding \texttt{sum}) across four datasets.} 
        The \texttt{eum} position shows the largest variability on the \texttt{mmlu} dataset.
        }
        \vspace{-1em}
    \label{fig:pred-change-all}
\end{figure}

\noindent \textbf{Classification and QA Tasks.}
Across \textsc{MNLI}, \textsc{AG News}, \textsc{ARC}, and \textsc{MMLU}, placing demonstrations at \texttt{ssp} yields the most consistent accuracy improvements (\autoref{fig:accuracy-change-all} ; \autoref{fig:pred-change-all}). Notably, \textsc{MMLU} shows a \textit{+18\%} gain in accuracy over the zero-shot baseline under \texttt{ssp}. For QA tasks like \textsc{SQuAD}, \texttt{ssp} similarly outperforms later placements, while \texttt{eum} consistently underperforms.

\vspace{0.5em}
\noindent \textbf{Arithmetic Tasks.}
Arithmetic reasoning exhibits scale-sensitive trends. When evaluated, models with smaller parameter sizes (1.5B - 8B) are consistent in preferring demos being placed in the \textit{ssp, esp} positions. For \textsc{LLAMA3 3B}, moving demos from \texttt{ssp} to \texttt{eum} causes a drop in improved prediction rate: \textsc{GSM8K} falls from \textit{42.0\%} to \textit{11\%}, and \textsc{SQuAD} from \textit{41.0\%} to \textit{26.5\%}. Conversely, \textsc{LLAMA3 70B} benefits from \texttt{eum}, improving from \textit{21.5\%} to \textit{88\%} on \textsc{GSM8K}, suggesting that model capacity modulates the effect of position.

\vspace{0.5em}
\noindent \textbf{Generative Summarization.} Performance volatility is most severe in generation tasks. On \textsc{LLAMA3 3B}, \textsc{XSum} sees a drop from \textit{82.5\%} to \textit{27.5\%} improved predictions when shifting from \texttt{ssp} to \texttt{eum},
while \textsc{CNN/DailyMail} drops from \textit{49\%} to a mere \textit{1\%}. These effects percist even in large models, albeit with reduced severity.

\begin{figure*}[h]
  \centering
  \begin{subfigure}[t]{\linewidth}
    \centering
    \includegraphics[width=\linewidth]{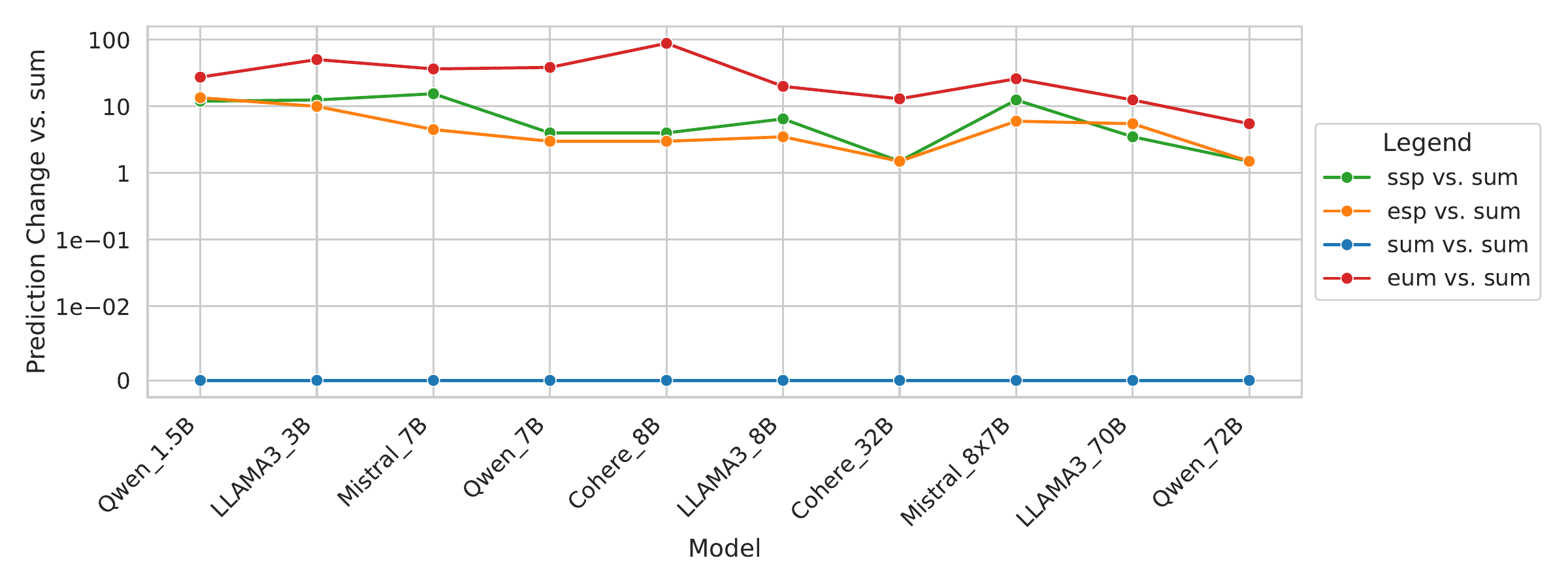}
    \caption{Prediction change (vs.\ \texttt{sum}) of the four \texttt{DPP}s on MMLU: it declines as model scale increases.}
    \label{fig:mmlu_vs_sum}
  \end{subfigure}

  \vspace{1em} 

  \begin{subfigure}[t]{\linewidth}
    \centering
    \includegraphics[width=\linewidth]{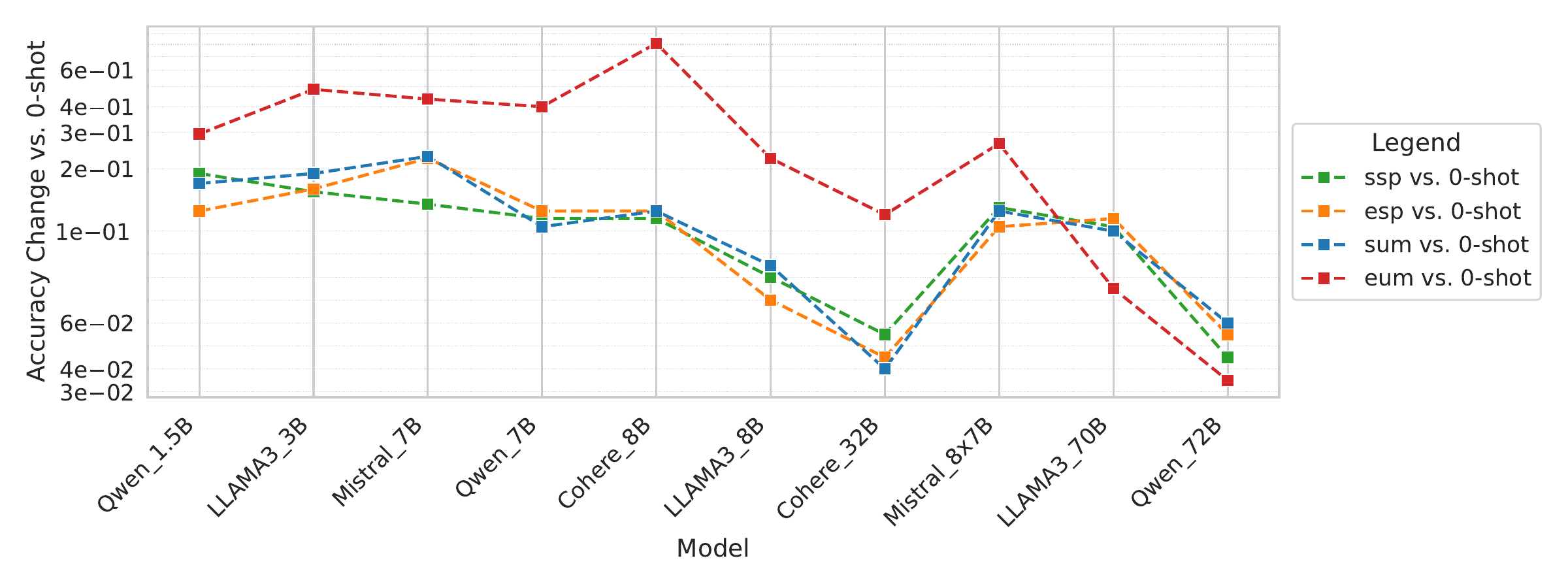}
    \caption{Accuracy change (improvement over \ zero‐shot) of the four \texttt{DPP}s on MMLU: it declines as model scale increases.}
    \label{fig:mmlu_vs_0shot}
  \end{subfigure}

  \vspace{-0.5em} 

  \caption{\textbf{Scaling behavior of \texttt{DPP}s on MMLU.}  
    (a) Prediction‐level shifts relative to \texttt{sum}, and (b) accuracy shifts relative to zero‐shot, both across 10 model sizes (1.5B–72B). Both metrics reveal a weak scaling law: as the model scale increases, the variations caused by \texttt{DPP}s in accuracy and prediction from baselines gradually decline.}
  \label{fig:mmlu_transition_two_panel}
\end{figure*}
\subsection{Scaling Law of Performance Robustness}
To better understand how positional robustness varies with model scale, \autoref{fig:mmlu_transition_two_panel} summarizes the percentage of changed predictions and the accuracy deltas as we analyze the percentage of changed and improved predictions across the four prompt positions. Across all tasks, we observe that larger models generally exhibit reduced prediction volatility (\textit{\% changed}) and enhanced performance stability, but the degree of robustness is task-dependent and not uniformly monotonic with size. \looseness-1

\vspace{0.5em}
\noindent\textbf{Stability Trends Across Tasks.}
On \textbf{classification tasks} such as \textit{AG News}, \textit{MNLI}, and \textit{ARC}, larger models (e.g., \textsc{Qwen 72B}, \textsc{LLaMA3 70B}) exhibit reduced sensitivity to prompt position changes, especially for early-positioned demonstrations (\texttt{ssp}, \texttt{esp}). For example, on \textit{MNLI}, the percentage of predictions that change when moving from \texttt{sum} to \texttt{ssp} drops below \textit{10\%} for \textsc{LLaMA3 70B}, compared to over \textit{20\%} for \textsc{LLaMA3 3B}. Meanwhile, accuracy improvements over zero-shot are consistently higher for early positions but show greater spread across mid-sized models (e.g., 7B–32B). This indicates that while small models benefit from positional tuning, they are also more fragile to changes.

 \noindent On \textbf{question answering tasks} like \textit{SQuAD} and \textit{GSM8K}, the pattern is more nuanced. For \textit{GSM8K}, the change rate remains above \textit{90\%} across nearly all models and positions, indicating high sensitivity to demonstration placement. However, the percentage improvement fluctuates non-monotonically: models like \textsc{Mistral 8x7B} under-perform with \texttt{ssp} placement relative to both smaller and larger models, and \textsc{LLaMA3 70B} shows a complete collapse in improvement under \texttt{ssp}, contrasting its robustness on other tasks. This suggests arithmetic reasoning requires specialized inductive biases that do not scale uniformly with size.

\noindent In \textbf{summarization tasks} such as \textsc{XSum} and \textsc{CNN/DailyMail}, the percentage of prediction changes is consistently near \textit{100\%} for the \texttt{eum} position, even in the largest models. This reflects that downstream text generation is highly susceptible to positional shifts. Notably, larger models like \textsc{Qwen-72B} still exhibit drops in \textit{\% improved} when moving from \texttt{ssp} to \texttt{eum}, albeit less drastically than smaller counterparts. On \textit{CNN/DailyMail}, \texttt{eum} improves only 1\% of predictions for \textsc{LLaMA3-3B}, compared to \textit{49\%} under \texttt{ssp}, while \textsc{LLaMA3-70B} narrows that gap considerably.



\subsection{Analysis of \texttt{DPP} induced Transitions}
While accuracy-based evaluations reveal global trends in positional effectiveness, they can obscure local instability in model behavior. To uncover finer-grained effects, we visualize the answer transitions between correct and incorrect predictions using Sankey diagrams.

\begin{figure}[!ht]
    \centering
    \includegraphics[width=\linewidth]{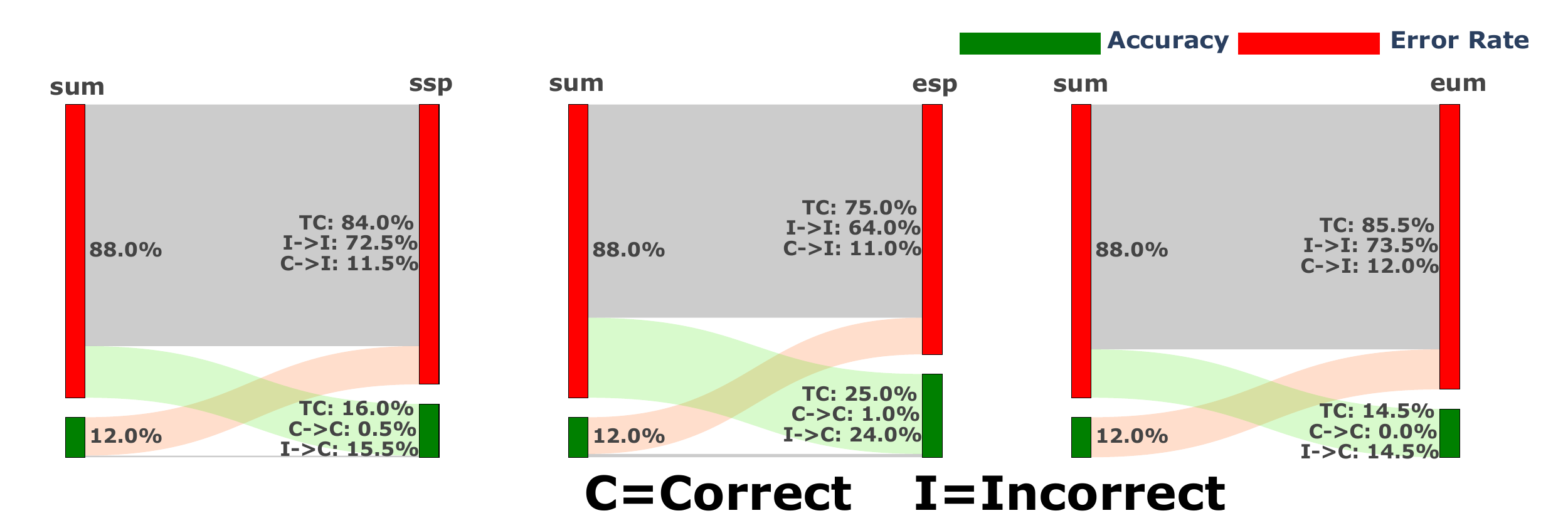}
    \caption{\textbf{Correct-Incorrect Transition} from the default baseline \texttt{DPP} ``\texttt{sum}'' to \texttt{ssp}, \texttt{esp}, and \texttt{eum} when applied to \textsc{Llama3-3B} model on \textit{XSUM} benchmark. Green and red bars denote the accuracy and error rate, respectively. Left and right bars are associated with the baseline and a specific \texttt{DPP}. We also report the percentage of examples that change from Incorrect→Correct (\textbf{I -> C}) and Correct→Incorrect (\textbf{C -> I}). 
    }
    \label{fig:sankey_llama3_mmlu}
\end{figure}
\vspace{0.5em}
\noindent\textbf{Volatility Patterns Across Tasks.}
Across the eight evaluated tasks, we observe a recurring pattern: later-positioned demos (\texttt{eum}) cause significantly more answer flips than earlier positions (\texttt{ssp}, \texttt{esp}). This suggests that placing demonstrations after the query can inject instability into model decision-making, especially in models with fewer inductive biases or weaker context modeling capabilities.

\begin{figure}[!ht]
\vspace{-1em}
    \centering
    \includegraphics[width=\linewidth]{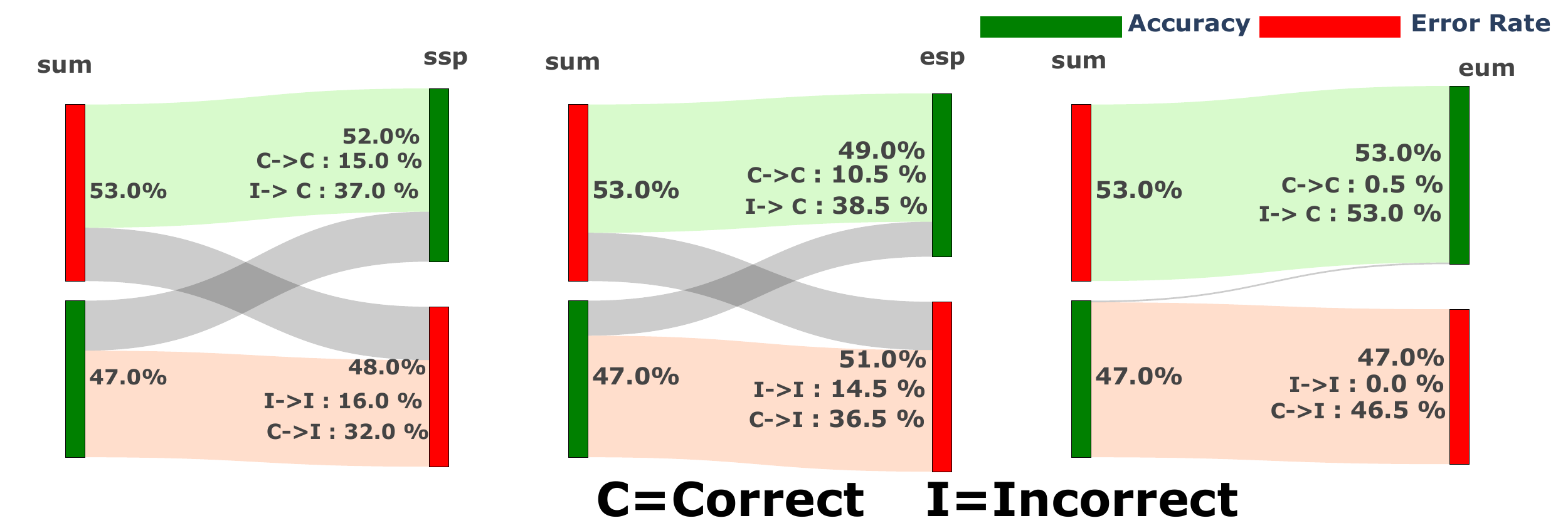}
    \caption{\textbf{Correct-Incorrect Transition} on \textsc{CNN/ DailyMail} for \textsc{Mistral-8x7B}. The high transition ratios between incorrect and correct samples indicate the sensitivity to the change of \texttt{DPP}.\looseness-1  
    }
    \vspace{-1em}
    \label{fig:sankey_mistral_cnn}
\end{figure}

\begin{figure}[!ht]
 \vspace{-1em}
    \centering
    \includegraphics[width=\linewidth]{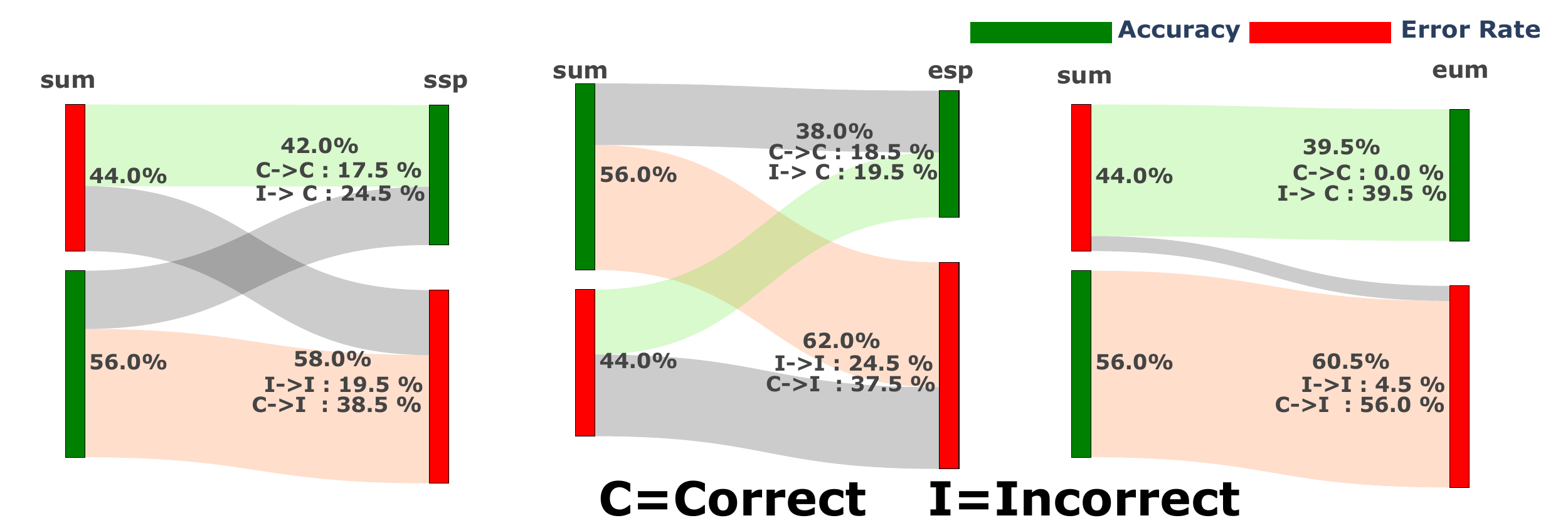}
    \caption{\textbf{Correct-Incorrect Transition} on \textsc{GSM8K} for \textsc{Qwen-72B}. Even for the largest model evaluated in this paper, $>$50\% predictions are changed when using different \texttt{DPP}.\looseness-1}
    \vspace{-1em}
    \label{fig:sankey_qwen72b_gsm}
\end{figure}

In Figure~\ref{fig:sankey_llama3_mmlu}, we see this volatility concretely for \textsc{LLaMA3 3B} on \textit{MMLU}, where moving from \texttt{ssp} to \texttt{eum} causes a large number of transitions from correct to incorrect answers. Similar patterns are seen on:

\begin{itemize}
    \item \textbf{\textit{AG News}}: Smaller models like \textsc{Qwen 1.5B} exhibit over 40\% incorrect-to-correct transitions under \texttt{ssp}, which plummet under \texttt{eum}.
    \item \textbf{\textit{CNN/DailyMail}}: \textsc{Mistral 8x7B} shows one of the most volatile behaviors, with many correct answers flipping to incorrect under late-positioned demos (Fig.~\ref{fig:sankey_mistral_cnn}).
    \item \textbf{\textit{GSM8K}}: Predictions by models like \textsc{Qwen 72B} and \textsc{LLAMA3 70B} still flip a lot across positions, despite their scales (Fig.~\ref{fig:sankey_qwen72b_gsm}).\looseness-1
\end{itemize}

Together, these transition plots reveal that \textit{the same input content,when moved across prompt sections can yield drastically different outputs}. The effect persists across models and tasks, underlining that prompt formatting is not merely stylistic, but functionally consequential. This volatility is especially concerning in high-stakes domains like QA or summarization, where reliability is paramount.

\vspace{0.5em}
\noindent\textbf{Scale-Driven Shifts in Optimal Position.}
Importantly, the position yielding the best improvement is not consistent across model sizes. On \textit{ARC}, \texttt{ssp} dominates for smaller models (\textsc{Qwen 1.5B} to \textsc{Mistral 7B}), whereas \texttt{eum} unexpectedly overtakes \texttt{ssp} in \textsc{Qwen 72B} albeit marginally. Similarly, on \textit{AG News}, while \texttt{ssp} yields the best result for \textsc{LLaMA3 3B}, \texttt{esp} becomes the strongest position in \textsc{LLaMA3 70B}.

\subsection{Winning \texttt{DPP} is Task and Model Specific}
While general trends suggest that early demonstration positions (\texttt{ssp}, \texttt{esp}) often outperform later ones (\texttt{sum}, \texttt{eum}), our analysis reveals that this preference is not consistent across all models or tasks. To understand this heterogeneity, we conduct a win–tie–loss analysis across tasks, identifying which demo position performs best for each task–model pair.

\begin{figure}[!ht]
\centering
\vspace{-1em}
  \includegraphics[width=\linewidth]{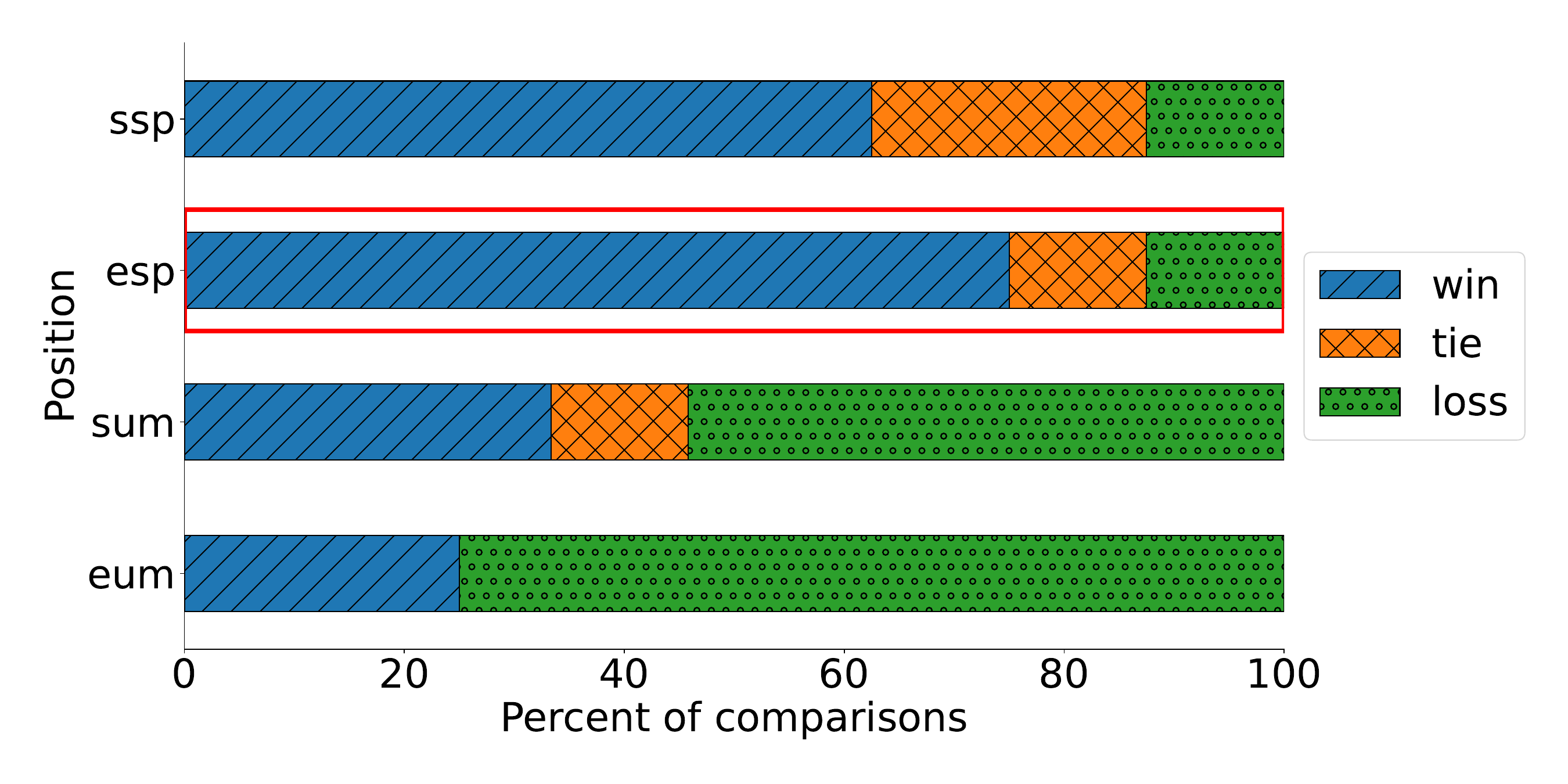}
  \vspace{-2em}
  \caption{Win–loss–tie of each \texttt{DPP} vs. zero-shot on \textsc{Qwen 1.5B} (averaged over all the eight benchmarks).\looseness-1} 
  \vspace{-1em}
  \label{fig:qween1d5bwl}
\end{figure}

\begin{figure}[!ht]
\centering
  \includegraphics[width=\linewidth]{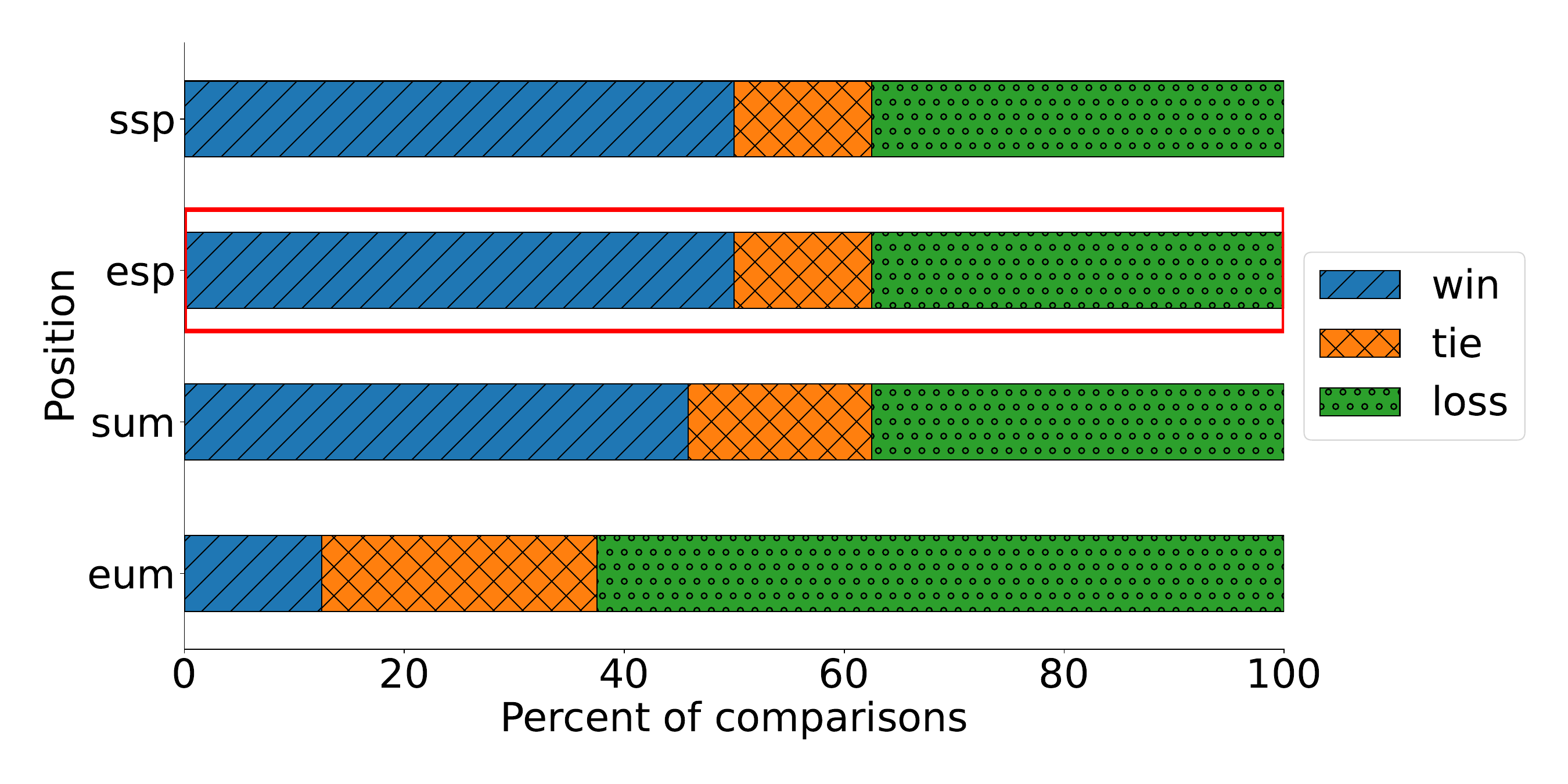}
  \vspace{-2em}
  \caption{Win–loss–tie of each \texttt{DPP} vs. zero-shot on \textsc{Cohere 8B} (averaged over all the eight benchmarks).} 
  \vspace{-1em}
  \label{fig:cohr8bwl}
\end{figure}

\begin{figure}[!ht]
\centering
  \includegraphics[width=\linewidth]{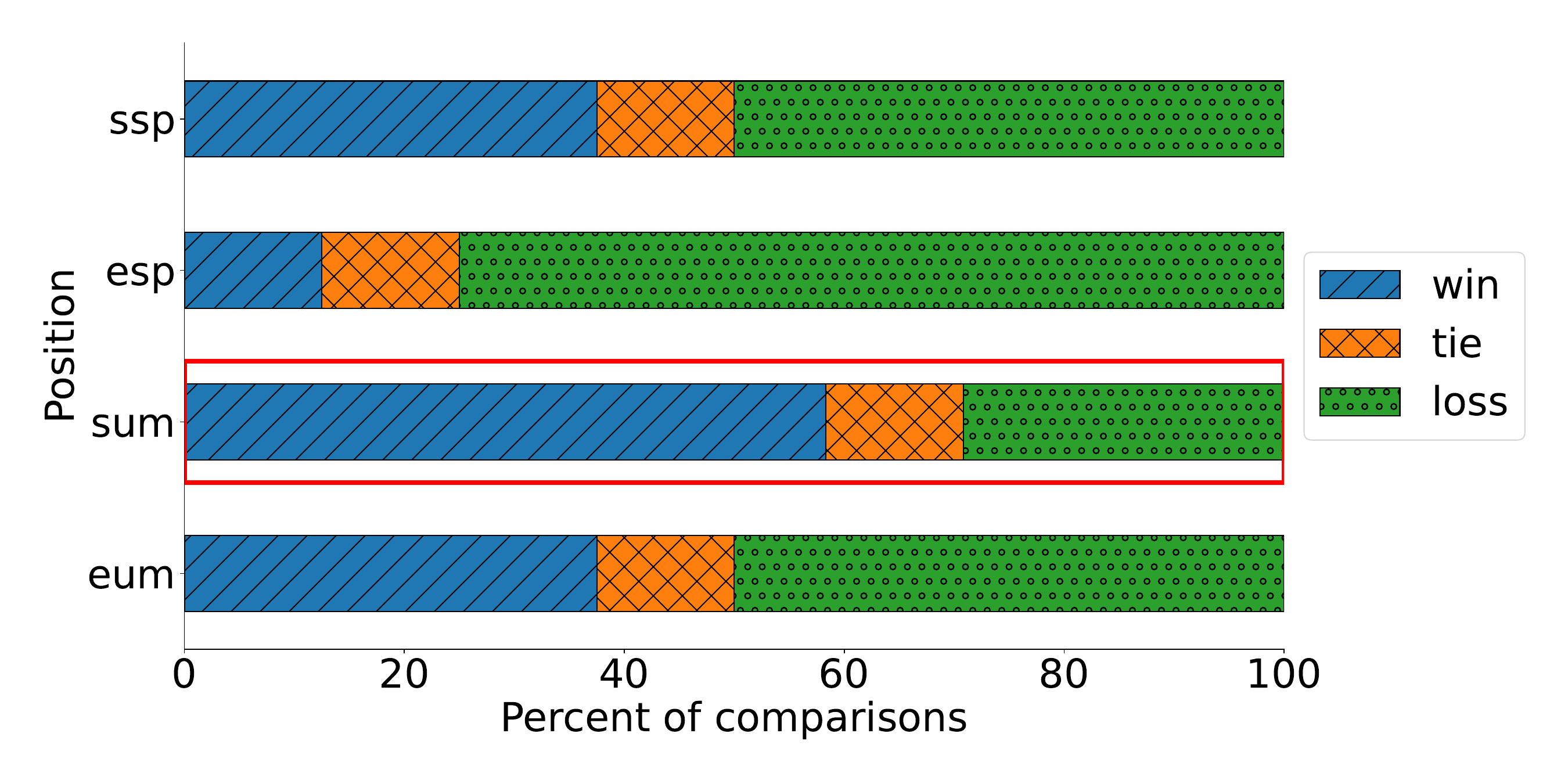}
  \vspace{-2em}
  \caption{Win–loss–tie of each \texttt{DPP} vs. zero-shot on \textsc{Llama3 70B} (averaged over all the eight benchmarks).} 
  \vspace{-1em}
  \label{fig:lma70b}
\end{figure}

Figures~\ref{fig:qween1d5bwl}, \ref{fig:cohr8bwl}, and \ref{fig:lma70b} illustrate this breakdown for three representative models at different scales: \textsc{Qwen-1.5B}, \textsc{Cohere-8B}, and \textsc{LLaMA3-70B}. These win–loss–tie plots display, for each position, the number of tasks where it yielded the best performance (win), tied for the best (tie), or was outperformed by the zero-shot baseline (loss).

\vspace{1em}
\textbf{\textsc{Qwen 1.5B}} (\autoref{fig:qween1d5bwl}): As the smallest model in our suite, \textsc{Qwen 1.5B} strongly prefers placing demos at the \texttt{esp} and \texttt{ssp} position. It wins on most tasks with \texttt{esp} and \texttt{ssp}, rarely losing. This suggests that smaller models are especially sensitive to how demonstrations are front-loaded in the prompt, likely due to limitations in long-range context integration.

\vspace{1em}
\textbf{\textsc{Cohere 8B}} (\autoref{fig:cohr8bwl}): At 8B parameters, Cohere shows moderate flexibility. While \texttt{ssp} still wins most often, \texttt{sum} begin to win on some tasks, particularly \textsc{XSum} and \textsc{SQuAD} indicating that as model capacity grows, preferences start to shift depending on task format and type (classification vs. QA vs. generation).

\vspace{1em}
\textbf{\textsc{LLaMA3 70B}} (\autoref{fig:lma70b}): In contrast to smaller models, \textsc{LLaMA3 70B} shows a consistent preference for placing demonstrations at the \texttt{sum} position, that is, at the start of the user message. Across multiple tasks, \texttt{sum} outperforms all other configurations, including \texttt{ssp} and \texttt{esp}, which dominate in earlier models. This suggests that larger models like \textsc{LLaMA3 70B} may benefit from having demonstrations placed in closer proximity to the query, perhaps due to their greater ability to retain relevant context across longer input sequences.

\vspace{0.5em}
\noindent \textbf{Emergent Observation: No Universally Best Position.} Our results demonstrate that 
early positions dominate on average but exceptions emerge for arithmetic tasks.
Instead, the optimal position varies by both model architecture and task category. For example, in generative summarization tasks, later positions (\texttt{sum}, \texttt{eum}) occasionally outperform early ones, whereas in classification and reasoning tasks, early positions (\texttt{ssp}, \texttt{esp}) are generally more reliable.

\noindent For completeness, we provide win–loss–tie plots for all remaining models and also task specific plots in the \autoref{apdx:fwtl} (Figures~\ref{fig:models-side-by-side1}--\ref{fig:models-side-by-side7}). Collectively, they confirm the absence of a universally optimal position and highlight the need for model-specific prompt tuning.

\subsection{Statistical Test of Performance Difference between zero-shot vs. ICL with each \texttt{DPP}}
\begin{table}[h]
    \centering
    {\tiny
    \begin{tabular}{lccccc}
        \toprule
        \textbf{Position} & 
        \textbf{0-shot Accuracy} & 
        \textbf{ICL Accuracy} & 
        \textbf{p-value} & 
        \textbf{Effect Size} \\
        \midrule
        \textit{ssp} & 0.3364 & 0.6885 & 0.0022** & 1.7193 \\
        \textit{esp} & 0.3364 & 0.6950 & 0.0022** & 1.7000 \\
        \textit{sum} & 0.3364 & 0.6869 & 0.0022** & 1.7254 \\
        \textit{eum} & 0.3364 & 0.4519 & 0.1659 & 0.4140 \\
        \bottomrule
    \end{tabular}
    }
    \caption{Comparing zero-shot vs. the four \texttt{DPP}s on MMLU dataset (averaged over all models) via one-sided Wilcoxon signed-rank test. **--statistical significance at 1\%.}
    \label{tab:stats}
\end{table}

 To quantify the reliability of performance differences across demonstration positions, we conduct a paired statistical analysis comparing each of the four \texttt{DPP}s to the zero-shot baseline. (See \autoref{tab:stats})
 

 For each dataset and \texttt{DPP}, we form paired samples across the available models. We then perform a one-sided Wilcoxon signed-rank test to assess whether the positional condition of the ICL improves over baseline. Specifically, we test the null hypothesis \textbf{$H_0$: the median difference between the \texttt{DPP} and the baseline is zero}, against the alternative hypothesis \textbf{$H_1$: the median difference is greater than zero}, indicating that the \texttt{DPP} outperforms the baseline. The effect sizes are calculated as the standardized mean difference of paired differences. In addition, we apply a multiple comparisons correction (using the FDR Benjamini–Hochberg procedure at $\alpha = 0.05$) to account for the fact that multiple hypotheses are tested simultaneously. 
This analysis provides statistical rigor to our evaluation, helping us determine not just whether differences exist, but whether they are consistently positive across models. By quantifying both the statistical significance and effect size, we can better assess the reliability and practical importance of each \texttt{DPP}.

\begin{table}[!ht]
  \centering
  \small
  \begin{tabular}{llcccc}
    \toprule
    \textbf{Pos1} & \textbf{Pos2} & \(\Delta\) & \(W\)   & \(r\)    & \(p\) (FDR)\\
    \midrule
    SSP & ESP & \(-0.023\) & 9    & \(-0.465\) & 0.428 \\
    SSP & SUM & \(-0.026\) & 14.5 & \(-0.511\) & 0.413 \\
    SSP & EUM & \(+0.182\) & 0    & \(0.905\)  & \textbf{0.023*} \\
    ESP & SUM & \(-0.003\) & 17.5 & \(-0.193\) & 1.000 \\
    ESP & EUM & \(+0.205\) & 3    & \(1.002\)  & \textbf{0.042*} \\
    SUM & EUM & \(+0.209\) & 0    & \(1.038\)  & \textbf{0.023*} \\
    \bottomrule
  \end{tabular}
  \caption{Pairwise Wilcoxon-signed-rank tests among DPP positions on MMLU (10 models).  \(\Delta\)=mean accuracy difference (Pos1 -- Pos2), \(W\)=Wilcoxon statistic, \(r\)=effect size, \(p\)=FDR-corrected.  \(*p<.05\).}
  \label{tab:pairwise-mmlu}
\end{table}

In \autoref{tab:stats}, the three \texttt{DPPs} (SSP, ESP, SUM) each yield dramatic and highly significant improvements over zero‐shot on MMLU, increasing accuracy from 0.3364 to approximately 0.69 (Wilcoxon $W\leq 2$, $r\approx1.7$, FDR‐corrected $p<.01$), whereas EUM only raises accuracy to 0.4519 ($W=10$, $r=0.414$, $p=0.1659$), a non‐significant gain.  The deeper pairwise analysis in \autoref{tab:pairwise-mmlu} confirms that EUM is significantly worse than each of the other \texttt{DPPs}, SSP vs. EUM ($\Delta=+0.182$, $W=0$, $r=0.905$, $p=0.023$), ESP vs. EUM ($\Delta=+0.205$, $W=3$, $r=1.002$, $p=0.043$), and SUM vs. EUM ($\Delta=+0.209$, $W=0$, $r=1.038$, $p=0.023$).  In contrast, all tests among SSP, ESP, and SUM produce trivial mean differences ($|\Delta|<0.03$) and non‐significant $p$‐values ($p>0.4$).  Together, these results demonstrate that while placing demonstrations before the user query consistently delivers robust few‐shot gains, embedding them at the end of the user message severely degrades performance. 

\section{Discussion}
\label{sec:discussion}
\subsection{Why Does DPP Bias Arise?}
We hypothesise that \textbf{\texttt{DPP} bias} has two complementary roots.  
\textbf{(i) Architectural:}  Causal-decoder LLMs are trained with autoregressive masking, so earlier tokens exert disproportionate influence on the hidden state that conditions all subsequent predictions.  Although the system, demonstration, and user segments are \emph{logically} exchangeable, the underlying optimisation objective is not.  Mechanistic-interpretability work on ``induction heads’’ \citep{olsson2022incontextlearninginductionheads}, for instance, has shown that attention weights concentrate on early and sink tokens—behaviour consonant with the empirical trends we observe for DPP.  
\textbf{(ii) Data:}  Instruction-tuning corpora themselves exhibit positional regularities (e.g., demonstrations in fixed slots), thereby imprinting a distributional prior into the model.  Domain-specific training sets may amplify or attenuate this effect, explaining why the optimal demo position varies across tasks and models.

\subsection{Mitigating DPP Bias}
We outline two directions for future work aimed at reducing this positional pathology.  

\noindent\textbf{Test-time calibration.}  
Given the task- and model-specific nature of the bias, we propose a retrieval-based calibration: for each unseen instance, find its $k$ nearest neighbours in an annotated reference set (using an external embedding model), then \emph{majority-vote} over their labelled best positions to select the demo slot for the query instance.  This lightweight procedure incurs no fine-tuning cost and adapts dynamically to input distribution shifts.

\noindent\textbf{Post-training on randomly permuted contexts.}  
Alternatively, one can train an \emph{unbiased} in-context learning corpus by randomly permuting the demonstration positions in every training example. Fine-tuning (or continued pre-training) on this new dataset should encourage position-invariant representations, counteracting the structural preference induced by standard instruction-tuning pipelines.

\section{Conclusion}
\label{sec:conclusion}

This paper introduces and systematically investigates a previously overlooked dimension of in-context learning (ICL): the effects of the positional placement of demonstrations within LLM prompts. Through a large-scale evaluation spanning ten open-source models, eight NLP tasks, and four canonical prompt positions, we uncover a consistent \texttt{DPP}  bias, where demos placed earlier in the prompt (\texttt{ssp}, \texttt{esp}) yield higher accuracy and greater prediction stability than those placed later (\texttt{sum}, \texttt{eum}). These findings persist across both classification and generative tasks and are particularly pronounced in smaller models.

\noindent Our analysis reveals that not only does performance vary substantially by position, but late-placed demonstrations (especially \texttt{eum}) can induce significant prediction volatility flipping model outputs without improving correctness. We further show that positional sensitivity is modulated by both task and model scale: while larger models demonstrate greater robustness, they still exhibit non-trivial instability and shifting optimal positions across tasks.

\noindent We introduce novel diagnostic tools, \textsc{accuracy-change} and \textsc{prediction-change}
to quantify these effects and uncover hidden volatility that standard accuracy metrics obscure. Our win--tie--loss analyses reinforce the key insight: \textbf{no single demonstration position is universally optimal}. Effective prompt design must therefore be both \textit{model-aware} and \textit{task-sensitive}.

\noindent These findings have broad implications for prompting strategies in practice. We recommend that users of instruction-tuned LLMs explicitly evaluate demonstration placement rather than relying on default or ad hoc formats. Furthermore, positional robustness should be considered a core axis in both prompt optimization and instruction fine-tuning pipelines.

\vspace{0.5em}
\noindent \textbf{Future Work.} Our study opens up several avenues for follow-up research. First, deeper interpretability work could investigate \emph{why} certain positions are privileged, whether due to attention initialization, decoder primacy, instruction tuning templates or training corpus conventions. Second, extending this analysis to few-shot chain-of-thought prompts and real-world instruction datasets (e.g., HELM, BIG-Bench) could help generalize these insights. Finally, developing automated demo-placement optimization routines that adapt position jointly with content could offer a principled pathway toward more robust ICL systems.

\section{Ethics Statement}
\label{sec:ethics}
Our work focuses on the technical aspects of prompt design and does not directly engage with potentially sensitive content or private data. However, the following ethical considerations are relevant:

\begin{enumerate}
    \item \textbf{Misuse of Prompt Engineering:} Enhanced control over LLM behavior through strategic demonstration placement could be exploited to generate deceptive or harmful content more effectively. We encourage researchers to incorporate content filtering and moderation frameworks when deploying these methods.
    \item \textbf{Bias and Fairness:} If demonstrations carry implicit biases (e.g., skewed label distributions or stereotypical examples), placing them early in the prompt may amplify such biases in model outputs. Practitioners should carefully curate demonstration sets and validate outputs for unintended bias.
\end{enumerate}

We believe that increasing awareness of spatial effects in prompts will ultimately aid in designing safer, more reliable LLM-based systems while mitigating misuse and bias.

\section{Limitations} \label{sec:limitations}
While our experiments reveal robust trends in how demonstration placement impacts LLM performance, several limitations remain:

\begin{itemize}
    \item \textbf{Model Diversity:} We evaluated only a small subset of model sizes and architectures (e.g., 7B, 13B). Larger-scale models or different architectures (e.g., those fine-tuned on dialogue) may exhibit different sensitivity patterns.
    
    \item \textbf{Task Coverage:} Though we tested multiple tasks (classification, QA, summarization, reasoning), certain tasks with more complex structures (e.g., multi-hop retrieval or dialogic contexts) were not explored in depth.
    
    \item \textbf{Focus on English:} Our results primarily focus on English data. Cross-lingual variations in grammar, morphology, and script may lead to different positional biases and should be investigated further.
    
    \item \textbf{Automated Evaluation Metrics:} We relied on standard metrics (accuracy, F1, ROUGE) to quantify performance. These are imperfect proxies for true utility, especially for generative tasks. It’s conceivable that a prompt layout yields a higher ROUGE but lower factuality, for example. We assume the metrics correlate with better quality in our tasks, which is generally accepted, but caution that “better metric” doesn’t always mean strictly better output in all aspects.
\end{itemize}

\noindent Addressing these limitations will be crucial for fully understanding the impact of demonstration placement across diverse LLMs, languages, and application domains. We hope our findings will catalyze more research into robust, spatially aware prompting techniques.



\bibliography{anthology,custom}

\appendix
\newpage
\section{Appendix} \label{sec:appendix}

\subsection{Foundations of In-Context Learning}
The ability of large language models (LLMs) to adapt to novel tasks through in-context learning (ICL)—learning from demonstrations embedded directly in the input prompt—has emerged as a hallmark of their generalization capabilities \cite{NEURIPS2020_1457c0d6}. Early studies underscored the remarkable ability of LLMs to generalize from minimal context, a capability that was later extended to zero-shot settings \cite{radford2019language}. Unlike traditional fine-tuning, ICL requires no gradient updates, enabling rapid task adaptation in zero- and few-shot settings \cite{NEURIPS2022_9d560961}. Recent works, such as \citeauthor{zhang2023speakforeignlanguagesvoice}, synthesize the evolution of ICL, framing it as both a practical tool for task-specific adaptation and a window into understanding emergent behaviors in LLMs. However, these works \citealp[][]{kim2022selfgeneratedincontextlearningleveraging, lu-etal-2022-fantastically, yang2024autoiclincontextlearninghuman, liu2024letslearnstepstep} highlight a critical unresolved challenge: the brittleness of ICL to seemingly minor variations in prompt structure, including the ordering \citep{lu-etal-2022-fantastically, liu2024letslearnstepstep} and formatting \citep{kim2022selfgeneratedincontextlearningleveraging, hao2022structuredpromptingscalingincontext, yang2024autoiclincontextlearninghuman} of demonstrations, as well as the selection of the demonstrations.



\subsection{ Prompting LLMs} \label{apdx:pmtllms}
\label{ssec:model-interaction}
\paragraph{Prompt Format and Instruction-Tuning.}
The model families in our study (\textsc{Qwen}, \textsc{Mistral}, \textsc{LLaMA3}, and \textsc{Cohere}) are instruction-tuned using chat-style templates that explicitly separate prompt segments into system instructions, user messages, and assistant responses. These templates are commonly implemented using structured tags (e.g., \texttt{<|system|>}, role delimiters) that guide the model's internal parsing of the prompt.\footnote{See Hugging Face's chat template documentation: \url{https://huggingface.co/docs/transformers/main/chat_templating}, and instruction-tuning frameworks such as LLaMA Factory: \url{https://github.com/hiyouga/LLaMA-Factory}} As a result, demonstration position within these fields (whether they appear in the system prompt versus the user message) interacts with the model's learned formatting biases. Our experiments quantify this interaction and reveal a systematic spatial preference that emerges from instruction-tuned behavior.

\paragraph{Model Instantiation.}
We wrap each LLM in a unified \emph{ChatModel} interface, parameterized by model type (e.g., \textsc{LLAMA3\_8B}, \textsc{LLAMA3\_70B}) and decoding settings. This abstraction ensures consistent usage across tasks. We set the \texttt{temperature} to 0 for deterministic decoding. For multiple-choice tasks, we cap \texttt{max\_new\_tokens} at 50, and for generative tasks, at 500.

\paragraph{Question Processing.}
For each query $q_j$, we:
\begin{enumerate}[]
    \item \emph{Assemble} the prompt: combine the chosen prompt template, the formatted demonstrations (possibly shuffled or ablated), and $q_j$.
    \item \emph{Check length}: as some demonstrations $\mathcal{D}_\tau$ might exceed the model defined token limits, we estimate the token length to ensure we do not exceed model limits (e.g., 8192 tokens).
    \item \emph{Generate response}: feed the prompt into $f_\theta$ via streaming token-by-token output. 
\end{enumerate}

\subsection{Final System Prompts} \label{apdx:fp}
\begin{itemize}
    \item AG News
    \begin{RoundedBox}
    You are a text classification assistant. You will receive a news article and must classify it into one of the following categories: World, Sports, Business, or Sci/Tech. Respond with only the category name. Do not provide any explanations in your response. Provide your answer as a json object with the key 'Answer'.
    \end{RoundedBox}

    \item MNLI
    \begin{RoundedBox}
        You are a multi-genre natural language inference system. When given two sentences (premise and hypothesis), determine whether the relationship is entailment, neutral, or contradiction. Handle diverse domains including fiction, government reports, telephone speech, and more. Do not provide any explanations in your response. Provide your answer as a json object with the key 'Answer'.
    \end{RoundedBox}
    \newpage
    \item ARC
    \begin{RoundedBox}
        You are a science-focused tutor who provides detailed reasoning for multiple-choice questions at the middle-school and high-school level. You excel at scientific reasoning and can clarify your thought process if asked. When given a question with several possible answers, identify the most scientifically accurate choice. Do not provide any explanations in your response. Provide your answer as a json object with the key 'Answer'.
    \end{RoundedBox}
    \item MMLU
    \begin{RoundedBox}
        You are an expert tutor with broad interdisciplinary knowledge. You can answer college-level and advanced high-school multiple-choice questions across numerous subjects, from mathematics and science to humanities and law. When given a question and multiple options, select the best option based on your expertise. Do not provide any explanations in your response. Provide your answer as a json object with the key 'Answer'.

    \end{RoundedBox}
    \item CNN/Dailymail
    \begin{RoundedBox}
        You are a summarization expert for news articles. Given a full news story, produce a concise summary capturing the main points. Avoid adding personal commentary or speculative details. Stick to the facts from the article. Do not provide any explanations in your response. Provide your answer as a json object with the key 'Answer'.
    \end{RoundedBox}
    \item XSUM
    \begin{RoundedBox}
        "You are a summarization expert for news articles. Given a full news story, produce a concise summary capturing the main points. Avoid adding personal commentary or speculative details. Stick to the facts from the article. Do not provide any explanations in your response. Provide your answer as a json object with the key 'Answer'.
    \end{RoundedBox}
    \newpage
    \item SQuAD
    \begin{RoundedBox}
        You are a reading comprehension assistant. Given a passage (context) and a question, you identify the most accurate answer from the passage. You only rely on the provided text and avoid adding extraneous information. Do not provide any explanations in your response. Provide your answer as a json object with the key 'Answer'.
    \end{RoundedBox}
    \item GSM8K
    \begin{RoundedBox}
        You are a math tutor specializing in grade-school arithmetic and algebra word problems. Explain your reasoning step by step (if requested) and provide the final numeric or short answer. Emphasize clarity and correctness in each step. Provide your answer as a json object with the key 'Answer'.
    \end{RoundedBox}
\end{itemize}

\vspace{3.0em}
\noindent \textbf{DPP templates}
\begin{enumerate}

  \item \textit{ssp}
    \begin{RoundedBox}
        \textbf{<system>} \\
        Use the demos below as examples on how to answer the question\\
        \texttt{<DEMOS\_PLACEHOLDER>}\\
        \texttt{<SYSTEM\_PLACEHOLDER>} \\
        \textbf{<end\_of\_system>} \\
        \textbf{<user>} \\
        \texttt{<QUESTION\_PLACEHOLDER>}\\ 
        \textbf{<end\_of\_user>}
    \end{RoundedBox}
    
  \item \textit{esp}
    \begin{RoundedBox}
        \textbf{<system>} \\
        \texttt{<SYSTEM\_PLACEHOLDER>} \\
        Use the demos below as examples on how to answer the question\\
        \texttt{<DEMOS\_PLACEHOLDER>}\\
        \textbf{<end\_of\_system>} \\
        \textbf{<user>} \\
        \texttt{<QUESTION\_PLACEHOLDER>}\\ 
        \textbf{<end\_of\_user>}
    \end{RoundedBox}
    \newpage
  \item \textit{sum}
    \begin{RoundedBox}
        \textbf{<system>} \\
        \texttt{<SYSTEM\_PLACEHOLDER>} \\
        \textbf{<end\_of\_system>} \\
        \textbf{<user>} \\
        Use the demos below as examples on how to answer the question\\
        \texttt{<DEMOS\_PLACEHOLDER>}\\
        \texttt{<QUESTION\_PLACEHOLDER>}\\ 
        \textbf{<end\_of\_user>}
    \end{RoundedBox}
  \item \textit{eum}
    \begin{RoundedBox}
        \textbf{<system>} \\
        \texttt{<SYSTEM\_PLACEHOLDER>} \\
        \textbf{<end\_of\_system>} \\
        \textbf{<user>} \\
        Answer this question
        \texttt{<QUESTION\_PLACEHOLDER>}\\ 
        Use the demos below as examples on how to answer the question\\
        \texttt{<DEMOS\_PLACEHOLDER>}\\
        \textbf{<end\_of\_user>}
    \end{RoundedBox}
\end{enumerate}

\vspace{3.0em}
\subsection{Terms of use}
We adhere to the terms of usage provided by the model/dataset authors. 

\noindent\textbf{Licenses and Citations for Model Families}
\begin{itemize}
    \item Qwen \cite{yang2025qwen3technicalreport} : \url{https://choosealicense.com/licenses/apache-2.0/} 
    \item Cohere \cite{dang2024ayaexpansecombiningresearch} : \url{https://docs.cohere.com/docs/c4ai-acceptable-use-policy} ; \url{https://cohere.com/c4ai-cc-by-nc-license}
    \item Mistral \cite{Jiang2024Mixtral, Jiang2023Mistral7} : \url{https://mistral.ai/terms-of-service/} ; \url{https://choosealicense.com/licenses/apache-2.0/}
    \item LLAMA \cite{grattafiori2024llama3herdmodels} : \url{ai.meta.com/llama/use-policy} ; \url{https://huggingface.co/meta-llama/Meta-Llama-3-8B/blob/main/LICENSE}
\end{itemize}
\vspace{3em}
\noindent\textbf{Licenses and Citations for datasets}
\begin{itemize}
    \item AG News \cite{Zhang2015CharacterlevelCN} : \url{http://groups.di.unipi.it/~gulli/AG_corpus_of_news_articles.html}
    \item MNLI \cite{N18-1101} : \url{https://www.anc.org/OANC/license.txt} ; \url{https://huggingface.co/datasets/choosealicense/licenses/blob/main/markdown/mit.md} ; \url{https://spdx.org/licenses/CC-BY-SA-3.0} ; \url{https://spdx.org/licenses/CC-BY-3.0}
    \item ARC \cite{allenai:arc} : \url{https://huggingface.co/datasets/choosealicense/licenses/blob/main/markdown/cc-by-sa-4.0.md}
    \item MMLU \cite{hendrycks2021ethics, hendryckstest2021} : \url{https://github.com/hendrycks/test/blob/master/LICENSE}
    \item CNN/Dailymail \cite{DBLP-GEKSB15, see-etal-2017-get} : \url{https://huggingface.co/datasets/choosealicense/licenses/resolve/main/markdown/apache-2.0.md}
    \item XSUM \cite{Narayan2018DontGM} : \url{https://github.com/EdinburghNLP/XSum?tab=MIT-1-ov-file}
    \item SQuAD \cite{rajpurkar-etal-2016-squad} : \url{https://huggingface.co/datasets/choosealicense/licenses/resolve/main/markdown/cc-by-sa-4.0.md}
    \item GSM8K \cite{cobbe2021gsm8k} : \url{https://huggingface.co/datasets/choosealicense/licenses/resolve/main/markdown/mit.md}
\end{itemize}
\subsection{Experiment Details}
We discuss below the experiment details of our work. We detail the model sizes and hyperparameters as well as the computational resorces used.
\subsubsection{Model Size and Budget}
The model sized we use are between 1.5B parameters to 72B parameters:

\begin{itemize}
\item \textbf{Llama 3}: 3B, 8B and 70B (4-bit BnB)
\item \textbf{Mistral}: 7B (4-bit BnB) and Mixture-of-Experts 8$\times$7B (4-bit AWQ)
\item \textbf{Qwen}: 1.5B, 7B and 72B (4-bit BnB)
\item \textbf{Cohere}: 8B and 32B (4-bit BnB)
\end{itemize}

All checkpoints are served with vLLM\cite{kwon2023efficient} and loaded in 4-bit weight-only quantization (bitsandbytes \cite{dettmers2022llmint8} or
AWQ \cite{lin2023awq}) with Flash-Attention v2\cite{dao2023flashattention2} and a 1 000-token context window.\footnote{The
Mixture-of-Experts model is served with AWQ because vLLM currently lacks
bitsandbytes support for 8-expert routing.}

\paragraph{Compute budget.}  
Inference is performed on a cluster of A100 80 GB and RTX A4000 16 GB GPUs via
vLLM 0.4.0; tensor parallelism is disabled (1 GPU / model).  A single
8-task $\times$ 5-demo sweep for a 70 B model requires $\approx$\,1 GPU-hour
(temperature 0, no sampling).
\subsubsection{Experimental Setup And Hyperparameters}
\label{sec:appendix-hparams}

\begin{itemize}
    \item \textbf{Prompt structures.}  We cycle through four canonical demo slots
          (\textit{ssp}, \textit{esp}, \textit{sum}, \textit{eum}; see
          \S\ref{sec:methodology}).  Demo counts $k\!\in\!\{1,2,3,4,5\}$ are
          enumerated; ablations drop one demo at a time.
    \item \textbf{Generation parameters.}  Unless stated otherwise we use
          \texttt{temperature = 0.0}, \texttt{top\_p = 1.0}, \texttt{num\_beams =
          1}.  \texttt{max\_new\_tokens} is task-dependent: 50 for
          classification/QA, 500 for open-ended generation
          (\textit{CNN/DailyMail, XSum, GSM8K, Squad}).
    \item \textbf{Seed and reproducibility.}  All experiments use
          \texttt{seed=42}; we fix NumPy, Python and PyTorch RNGs before each
          run.
\end{itemize}

\subsubsection{Answer Extraction}
To robustly map an LLM’s free‐form output to our discrete labels, we implement the following multi‐step pipeline:
\begin{enumerate}
  \item Normalize whitespace and strip punctuation.
  \item Attempt to parse JSON‐like substrings and extract the “answer” field.
  \item Apply multiple‐choice heuristics (letter match or exact option‐text match).
  \item Scan for ``Answer:'' or ``Solution:'' prefixes.
  \item Fallback to returning the cleaned string, then perform an exact or fuzzy match against the label set (otherwise assign “other”).
\end{enumerate}
This ensures that even messy or verbose outputs get reliably converted into our evaluation labels.

\subsubsection{Evaluation Metrics}
\label{sec:appendix-metrics}

\vspace{0.5em}
\begin{table*}[t]
    \begin{center}
    \begin{tabular}{@{}ll@{}}
    \toprule
    \textbf{Task family}        & \textbf{Metrics reported}\\
    \midrule
    Classification (MNLI, ARC, MMLU, AG News) & Accuracy \\[2pt]
    Extractive QA (SQuAD, GSM8K)           & Exact Match, F$_1$ \\[2pt]
    Summarisation (CNN/DailyMail, XSum)               & ROUGE-1/2/L, BERTScore (P/R/F$_1$) \\[2pt]
    \midrule
    \multicolumn{2}{@{}l}{\emph{Auxiliary readability metrics for all tasks:}}
    \\
    \multicolumn{2}{@{}l}{Coleman–Liau, Flesch-Kincaid, Gunning-Fog}\\
    \bottomrule
    \end{tabular}
    \end{center}
\end{table*}

\vspace{2.0em}

\textbf{Additional Transition Metrics:} \\
\noindent \textbf{Improved (\%)} – the percentage of examples that switch from an
        \emph{incorrect} baseline prediction to a \emph{correct} one
        (\textsc{I$\rightarrow$C}).\\
        \textbf{Regressed (\%)} – the percentage that switch from
        \emph{correct} to \emph{incorrect}
        (\textsc{C$\rightarrow$I}).\\
        \textbf{Net~$\Delta_\text{pred}$} – \textit{Improved} minus
        \textit{Regressed}; a positive value indicates a net gain in prediction
        accuracy while a negative value indicates overall degradation.

\newpage

\subsection{Use of AI}
ChatGPT was used in this work to rephrase sentences, and write the code to generate tables. Most captions (Figures and Tables) were refined by AI.

\vspace{3.0em}

\newpage
\subsection{Additions Experimentational Results: Tables}

\begin{table*}[!httbp]
\centering
\footnotesize
\renewcommand{\arraystretch}{0.85} 
\setlength{\tabcolsep}{4pt}       
\begin{tabular}{l@{\hspace{2pt}}cccccccccccccccc}
\toprule
\multirow{3}{*}{\textbf{System}} 
        & \multicolumn{16}{c}{\textbf{Task}} \\
    \cmidrule(lr){2-17}
        & \multicolumn{4}{c}{\textbf{AG News}} & \multicolumn{4}{c}{\textbf{MNLI}} & \multicolumn{4}{c}{\textbf{ARC}} & \multicolumn{4}{c}{\textbf{MMLU}} \\
        \cmidrule(lr){2-5} \cmidrule(lr){6-9} \cmidrule(lr){10-13} \cmidrule(lr){14-17} 
        & \textit{ssp} & \textit{esp} & \textit{sum} & \textit{eum} & \textit{ssp} & \textit{esp} & \textit{sum} & \textit{eum} & \textit{ssp} & \textit{esp} & \textit{sum} & \textit{eum} & \textit{ssp} & \textit{esp} & \textit{sum} & \textit{eum} \\
    \midrule
     \texttt{Qwen\_1.5B} & 0.76 & 0.73 & 0.69 & 0.56 & 0.34 & 0.32 & 0.29 & 0.32 & 0.7 & 0.71 & 0.69 & 0.63 & 0.5 & 0.56 & 0.5 & 0.38 \\
 \texttt{Qwen\_7B} & 0.82 & 0.81 & 0.81 & 0.81 & 0.34 & 0.35 & 0.35 & 0.31 & 0.89 & 0.89 & 0.89 & 0.84 & 0.71 & 0.7 & 0.69 & 0.41 \\
 \texttt{Qwen\_72B} & 0.81 & 0.81 & 0.82 & 0.81 & 0.33 & 0.33 & 0.33 & 0.33 & 0.94 & 0.94 & 0.95 & 0.95 & 0.83 & 0.83 & 0.81 & 0.82 \\
 \texttt{Cohere\_8B} & 0.82 & 0.8 & 0.79 & 0.79 & 0.35 & 0.35 & 0.35 & 0.35 & 0.8 & 0.78 & 0.78 & 0.73 & 0.94 & 0.92 & 0.93 & 0.05 \\
 \texttt{Cohere\_32B} & 0.76 & 0.88 & 0.86 & 0.77 & 0.34 & 0.35 & 0.33 & 0.34 & 0.84 & 0.84 & 0.83 & 0.86 & 0.96 & 0.97 & 0.96 & 0.86 \\
 \texttt{Mistral\_7B} & 0.83 & 0.8 & 0.81 & 0.81 & 0.35 & 0.36 & 0.35 & 0.34 & 0.64 & 0.65 & 0.64 & 0.57 & 0.4 & 0.45 & 0.46 & 0.29 \\
 \texttt{Mistral\_8×7B} & 0.77 & 0.79 & 0.79 & 0.81 & 0.32 & 0.33 & 0.33 & 0.32 & 0.66 & 0.8 & 0.74 & 0.46 & 0.57 & 0.59 & 0.56 & 0.12 \\
 \texttt{LLAMA3\_3B} & 0.76 & 0.73 & 0.72 & 0.7 & 0.33 & 0.32 & 0.3 & 0.32 & 0.77 & 0.78 & 0.74 & 0.69 & 0.59 & 0.58 & 0.57 & 0.23 \\
 \texttt{LLAMA3\_8B} & 0.87 & 0.87 & 0.83 & 0.86 & 0.36 & 0.34 & 0.36 & 0.34 & 0.78 & 0.8 & 0.79 & 0.75 & 0.59 & 0.57 & 0.58 & 0.57 \\
 \texttt{LLAMA3\_70B} & 0.84 & 0.83 & 0.84 & 0.81 & 0.35 & 0.35 & 0.34 & 0.33 & 0.93 & 0.91 & 0.92 & 0.92 & 0.79 & 0.77 & 0.81 & 0.77 \\
\bottomrule\end{tabular}
\caption{Accuracy scores of ten LLMs on AG News, MNLI, ARC, and MMLU benchmarks under four prompting strategies: \textit{ssp} (demos at the start of the system prompt), \textit{esp} (demos at the end of the system prompt), \textit{sum} (demos at the start of the user message), and \textit{eum} (demos at the end of the user message).}\label{tab:my_custom_table1}\end{table*}

\begin{table*}[!httbp]
  \centering
  \footnotesize
  \renewcommand{\arraystretch}{0.85}
  \setlength{\tabcolsep}{4pt}
  \begin{tabular}{l@{\hspace{3pt}}cccccccccccc}
    \toprule
    \multirow{3}{*}{\textbf{System}} 
      & \multicolumn{12}{c}{\textbf{CNN/DailyMail}} \\
    \cmidrule(lr){2-13}
      & \multicolumn{4}{c}{\textbf{ROUGE-1}} 
      & \multicolumn{4}{c}{\textbf{ROUGE-2}} 
      & \multicolumn{4}{c}{\textbf{ROUGE-L}} \\
    \cmidrule(lr){2-5} \cmidrule(lr){6-9} \cmidrule(lr){10-13}
      & \textit{ssp} & \textit{esp} & \textit{sum} & \textit{eum} 
      & \textit{ssp} & \textit{esp} & \textit{sum} & \textit{eum}
      & \textit{ssp} & \textit{esp} & \textit{sum} & \textit{eum} \\
    \midrule
    \texttt{Qwen\_1.5B}   & 0.35 & 0.32 & 0.34 & 0.14 & 0.13 & 0.12 & 0.13 & 0.01 & 0.22 & 0.20 & 0.22 & 0.09 \\
    \texttt{Qwen\_7B}     & 0.38 & 0.38 & 0.38 & 0.23 & 0.13 & 0.13 & 0.13 & 0.06 & 0.24 & 0.24 & 0.24 & 0.15 \\
    \texttt{Qwen\_72B}    & 0.41 & 0.40 & 0.39 & 0.39 & 0.15 & 0.14 & 0.14 & 0.14 & 0.25 & 0.25 & 0.24 & 0.23 \\
    \texttt{Cohere\_8B}   & 0.42 & 0.41 & 0.42 & 0.23 & 0.18 & 0.17 & 0.17 & 0.06 & 0.28 & 0.27 & 0.27 & 0.15 \\
    \texttt{Cohere\_32B}  & 0.43 & 0.43 & 0.44 & 0.37 & 0.19 & 0.20 & 0.20 & 0.15 & 0.29 & 0.30 & 0.30 & 0.24 \\
    \texttt{Mistral\_7B}  & 0.35 & 0.36 & 0.36 & 0.15 & 0.14 & 0.15 & 0.15 & 0.01 & 0.22 & 0.23 & 0.23 & 0.10 \\
    \texttt{Mistral\_8×7B} & 0.35 & 0.33 & 0.32 & 0.35 & 0.13 & 0.12 & 0.12 & 0.13 & 0.22 & 0.20 & 0.20 & 0.21 \\
    \texttt{LLAMA3\_3B}   & 0.40 & 0.39 & 0.39 & 0.14 & 0.15 & 0.14 & 0.14 & 0.01 & 0.25 & 0.25 & 0.24 & 0.10 \\
    \texttt{LLAMA3\_8B}   & 0.39 & 0.39 & 0.40 & 0.38 & 0.15 & 0.15 & 0.15 & 0.15 & 0.24 & 0.24 & 0.25 & 0.23 \\
    \texttt{LLAMA3\_70B}  & 0.41 & 0.42 & 0.41 & 0.41 & 0.16 & 0.16 & 0.16 & 0.17 & 0.26 & 0.26 & 0.26 & 0.26 \\
    \bottomrule
  \end{tabular}

  \vspace{1ex}

  \begin{tabular}{l@{\hspace{3pt}}cccccccccccc}
    \toprule
    \multirow{3}{*}{\textbf{System}} 
      & \multicolumn{12}{c}{\textbf{XSUM}} \\
    \cmidrule(lr){2-13}
      & \multicolumn{4}{c}{\textbf{ROUGE-1}} 
      & \multicolumn{4}{c}{\textbf{ROUGE-2}} 
      & \multicolumn{4}{c}{\textbf{ROUGE-L}} \\
    \cmidrule(lr){2-5} \cmidrule(lr){6-9} \cmidrule(lr){10-13}
      & \textit{ssp} & \textit{esp} & \textit{sum} & \textit{eum} 
      & \textit{ssp} & \textit{esp} & \textit{sum} & \textit{eum}
      & \textit{ssp} & \textit{esp} & \textit{sum} & \textit{eum} \\
    \midrule
    \texttt{Qwen\_1.5B}   & 0.19 & 0.19 & 0.20 & 0.12 & 0.04 & 0.04 & 0.05 & 0.01 & 0.13 & 0.13 & 0.14 & 0.09 \\
    \texttt{Qwen\_7B}     & 0.24 & 0.27 & 0.26 & 0.16 & 0.06 & 0.07 & 0.07 & 0.01 & 0.16 & 0.19 & 0.18 & 0.13 \\
    \texttt{Qwen\_72B}    & 0.25 & 0.29 & 0.31 & 0.24 & 0.08 & 0.09 & 0.11 & 0.07 & 0.18 & 0.21 & 0.23 & 0.17 \\
    \texttt{Cohere\_8B}   & 0.32 & 0.37 & 0.38 & 0.17 & 0.12 & 0.16 & 0.16 & 0.04 & 0.24 & 0.28 & 0.29 & 0.12 \\
    \texttt{Cohere\_32B}  & 0.44 & 0.47 & 0.47 & 0.30 & 0.21 & 0.24 & 0.24 & 0.12 & 0.35 & 0.39 & 0.39 & 0.23 \\
    \texttt{Mistral\_7B}  & 0.19 & 0.19 & 0.19 & 0.09 & 0.05 & 0.05 & 0.05 & 0.01 & 0.13 & 0.13 & 0.13 & 0.07 \\
    \texttt{Mistral\_8×7B} & 0.23 & 0.21 & 0.22 & 0.20 & 0.07 & 0.07 & 0.07 & 0.06 & 0.16 & 0.15 & 0.16 & 0.14 \\
    \texttt{LLAMA3\_3B}   & 0.26 & 0.28 & 0.30 & 0.17 & 0.07 & 0.08 & 0.09 & 0.01 & 0.18 & 0.21 & 0.23 & 0.14 \\
    \texttt{LLAMA3\_8B}   & 0.30 & 0.33 & 0.32 & 0.24 & 0.09 & 0.11 & 0.11 & 0.07 & 0.22 & 0.24 & 0.24 & 0.17 \\
    \texttt{LLAMA3\_70B}  & 0.31 & 0.34 & 0.37 & 0.28 & 0.11 & 0.13 & 0.14 & 0.09 & 0.23 & 0.26 & 0.28 & 0.21 \\
    \bottomrule
  \end{tabular}
  \caption{ROUGE-1, ROUGE-2, and ROUGE-L scores for ten LLMs on the CNN/DailyMail and XSUM datasets. We evaluate four prompting strategies: \textit{ssp} (demos at the start of the system prompt), \textit{esp} (demos at the end of the system prompt), \textit{sum} (demos at the start of the user message), and \textit{eum} (demos at the end of the user message).}
  \label{tab:my_custom_table3}
\end{table*}
\begin{table*}[!httbp]
\centering
\footnotesize
\renewcommand{\arraystretch}{0.85}
\setlength{\tabcolsep}{4pt}
\begin{tabular}{l@{\hspace{2pt}}cccccccccccccccc}
\toprule
\multirow{3}{*}{\textbf{System}}
  & \multicolumn{16}{c}{\textbf{Tasks}}\\
\cmidrule(lr){2-17}
  & \multicolumn{8}{c}{\textbf{SQUAD}}
  & \multicolumn{8}{c}{\textbf{GSM8K}}\\
\cmidrule(lr){2-9} \cmidrule(lr){10-17}
  & \multicolumn{4}{c}{\textbf{Exact Match}} & \multicolumn{4}{c}{\textbf{F1}}
  & \multicolumn{4}{c}{\textbf{Exact Match}} & \multicolumn{4}{c}{\textbf{F1}}\\
\cmidrule(lr){2-5}\cmidrule(lr){6-9}\cmidrule(lr){10-13}\cmidrule(lr){14-17}
  & \textit{ssp}&\textit{esp}&\textit{sum}&\textit{eum}
  & \textit{ssp}&\textit{esp}&\textit{sum}&\textit{eum}
  & \textit{ssp}&\textit{esp}&\textit{sum}&\textit{eum}
  & \textit{ssp}&\textit{esp}&\textit{sum}&\textit{eum}\\
\midrule
 \texttt{Qwen\_1.5B} & 50.5 & 56.5 & 54.5 & 16.5 & 64.97 & 71.21 & 67.12 & 25.63 & - & - & - & - & 13 & 16.7 & 13.5 & 0.31 \\
 \texttt{Qwen\_7B} & 66.5 & 68.5 & 65.5 & 53 & 80.39 & 81.9 & 80.5 & 68.51 & - & - & - & - & 24.58 & 43.58 & 42.76 & 41.74 \\
 \texttt{Qwen\_72B} & 68.5 & 69.5 & 69.5 & 68 & 83.26 & 83.82 & 84.02 & 82.62 & - & - & - & - & 45.56 & 45.68 & 45.95 & 46.97 \\
 \texttt{Cohere\_8B} & 72 & 69 & 68.5 & 7 & 84.34 & 83.69 & 82.86 & 10.9 & - & - & - & - & 39.02 & 45.82 & 45.87 & 17.67 \\
 \texttt{Cohere\_32B} & 63 & 64.5 & 67 & 58 & 80.66 & 81.66 & 82.83 & 77.45 & - & - & - & - & 34.59 & 47.85 & 48.33 & 47.21 \\
 \texttt{Mistral\_7B} & 57 & 52.5 & 49 & 41 & 74.55 & 70.64 & 67.87 & 54.01 & - & - & - & - & 32.19 & 40.11 & 39.63 & 31.14 \\
 \texttt{Mistral\_8×7B} & 51.5 & 47 & 44.5 & 33.5 & 69.18 & 65.19 & 63.84 & 56.29 & - & - & - & - & 24.21 & 27.71 & 27.71 & 35.75 \\
 \texttt{LLAMA3\_3B} & 62 & 63.5 & 58 & 58.5 & 77.12 & 78.45 & 74.35 & 73.22 & - & - & - & - & 34.76 & 33.73 & 36.52 & 11.5 \\
 \texttt{LLAMA3\_8B} & 68 & 68 & 68.5 & 63 & 82.28 & 82.66 & 83.16 & 78.95 & - & - & - & - & 38.45 & 40.06 & 39.72 & 42.85 \\
 \texttt{LLAMA3\_70B} & 68 & 67.5 & 69 & 68 & 82.66 & 82.7 & 84.09 & 82.28 & - & - & - & - & 5.94 & 5.78 & 12.07 & 41.93 \\
\bottomrule\end{tabular}
\caption{Exact Match and F1 scores of ten LLMs on SQuAD and GSM8K benchmarks under four prompting strategies: \textit{ssp} (demos at the start of the system prompt), \textit{esp} (demos at the end of the system prompt), \textit{sum} (demos at the start of the user message), and \textit{eum} (demos at the end of the user message).}\label{tab:my_custom_table4}\end{table*}

\begin{table*}[!httbp]
    \centering
    \renewcommand{\arraystretch}{0.85}
    \setlength{\tabcolsep}{4pt}
    \begin{tabular}{l@{\hspace{4pt}}cccccc}
    \toprule
    \textbf{DPP Position} & \textbf{Qwen 1.5B} & \textbf{Qwen 7B} & \textbf{Qwen 72B} & \textbf{LLAMA3 3B} & \textbf{LLAMA3 8B} & \textbf{LLAMA3 70B} \\
    \midrule
        \texttt{ssp}      & \textbf{0.131}  & 0.099  & 0.120  & 0.163 & \textbf{0.186} & 0.076 \\
        \texttt{esp}      & 0.112  & 0.089  & 0.106  & \textbf{0.181} & 0.160 & 0.078 \\
        \texttt{sum}      & 0.128  & 0.105  & 0.099  & 0.175 & 0.168 & 0.091 \\
        \texttt{eum}      & 0.099  & \textbf{0.112}  & \textbf{0.169}  & 0.102 & 0.124 & \textbf{0.134} \\
    \bottomrule
    \end{tabular}
    \caption{Positional-bias persists on \texttt{Booksum}, a long-context summarization benchmark (chapters approx. 5 K tokens; 5-demo prompts is approx. 23 K tokens). We report ROUGE-L on six representative models. Bold numbers mark the best \texttt{DPP} position per model.}
\end{table*}

\begin{table*}[!httbp]
  \centering
  \begin{subtable}[t]{0.48\textwidth}
    \centering
    \begin{tabular}{@{}lcccc@{}}
      \toprule
      \textbf{Dataset} & \textbf{ssp} & \textbf{esp} & \textbf{sum} & \textbf{eum} \\
      \midrule
      \multicolumn{5}{c}{\texttt{Qwen\_1.5B}} \\
      \midrule
      \texttt{mnli}    & 0.0171  & 0.0108  & 0.0046  & 0.0124  \\
      \texttt{ag\_news}& -0.2592 & -0.2448 & -0.2256 & -0.1656 \\
      \texttt{arc}     & -0.1596 & -0.1610 & -0.1554 & -0.1386 \\
      \texttt{mmlu}    & -0.0155 & -0.0185 & -0.0155 & -0.0095 \\
      \midrule
      \multicolumn{5}{c}{\texttt{Qwen\_7B}} \\
      \midrule
      \texttt{mnli}    & -0.0048 &  0.0024 &  0.0048 & -0.0168 \\
      \texttt{ag\_news}& -0.0943 & -0.0918 & -0.0918 & -0.0909 \\
      \texttt{arc}     &  0.1370 &  0.1360 &  0.1370 &  0.1280 \\
      \texttt{mmlu}    &  0.0053 &  0.0052 &  0.0051 &  0.0023 \\
      \midrule
      \multicolumn{5}{c}{\texttt{Qwen\_72B}} \\
      \midrule
      \texttt{mnli}    & -0.0048 & -0.0064 & -0.0064 & -0.0048 \\
      \texttt{ag\_news}&  0.0756 &  0.0763 &  0.0777 &  0.0763 \\
      \texttt{arc}     & -0.5180 & -0.5180 & -0.5215 & -0.5215 \\
      \texttt{mmlu}    & -0.3965 & -0.3965 & -0.3873 & -0.3904 \\
      \midrule
      \multicolumn{5}{c}{\texttt{Cohere\_8B}} \\
      \midrule
      \texttt{mnli}    & -0.0030 & -0.0030 & -0.0030 & -0.0030 \\
      \texttt{ag\_news}&  0.5952 &  0.5712 &  0.5616 &  0.5616 \\
      \texttt{arc}     & -0.1728 & -0.1674 & -0.1688 & -0.1553 \\
      \texttt{mmlu}    & -0.4125 & -0.4042 & -0.4070 &  0.0743 \\
      \midrule
      \multicolumn{5}{c}{\texttt{Cohere\_32B}} \\
      \midrule
      \texttt{mnli}    & -0.0158 & -0.0040 & -0.0237 & -0.0158 \\
      \texttt{ag\_news}&  0.3941 &  0.4828 &  0.4722 &  0.4012 \\
      \texttt{arc}     & -0.3699 & -0.3699 & -0.3672 & -0.3807 \\
      \texttt{mmlu}    & -0.4882 & -0.4946 & -0.4914 & -0.4284 \\
      \bottomrule
    \end{tabular}
    \subcaption{Qwen\_1.5B, Qwen\_7B, Qwen\_72B, Cohere\_8B, Cohere\_32B}
    \label{tab:cohere32b_transition_a}
  \end{subtable}
  \hfill
  \begin{subtable}[t]{0.48\textwidth}
    \centering
    \begin{tabular}{@{}lcccc@{}}
      \toprule
      \textbf{Dataset} & \textbf{ssp} & \textbf{esp} & \textbf{sum} & \textbf{eum} \\
      \midrule
      \multicolumn{5}{c}{\texttt{Mistral\_7B}} \\
      \midrule
      \texttt{mnli}    &  0.0450 &  0.0550 &  0.0450 &  0.0350 \\
      \texttt{ag\_news}&  0.4209 &  0.4002 &  0.4071 &  0.4105 \\
      \texttt{arc}     &  0.4361 &  0.4450 &  0.4406 &  0.3827 \\
      \texttt{mmlu}    &  0.1420 &  0.1775 &  0.1846 &  0.0674 \\
      \midrule
      \multicolumn{5}{c}{\texttt{Mistral\_8×7B}} \\
      \midrule
      \texttt{mnli}    &  0.0041 &  0.0061 &  0.0061 &  0.0041 \\
      \texttt{ag\_news}& -0.1580 & -0.1653 & -0.1638 & -0.1696 \\
      \texttt{arc}     &  0.4845 &  0.6223 &  0.5653 &  0.2993 \\
      \texttt{mmlu}    &  0.1687 &  0.1778 &  0.1620 & -0.0337 \\
      \midrule
      \multicolumn{5}{c}{\texttt{LLAMA3\_3B}} \\
      \midrule
      \texttt{mnli}    &  0.0018 & -0.0018 & -0.0090 & -0.0018 \\
      \texttt{ag\_news}& -0.1100 & -0.1056 & -0.1023 & -0.0990 \\
      \texttt{arc}     & -0.3186 & -0.3213 & -0.3024 & -0.2754 \\
      \texttt{mmlu}    & -0.0675 & -0.0660 & -0.0637 & -0.0135 \\
      \midrule
      \multicolumn{5}{c}{\texttt{LLAMA3\_8B}} \\
      \midrule
      \texttt{mnli}    &  0.0126 &  0.0054 &  0.0126 &  0.0036 \\
      \texttt{ag\_news}& -0.4536 & -0.4536 & -0.4248 & -0.4464 \\
      \texttt{arc}     & -0.2644 & -0.2706 & -0.2685 & -0.2521 \\
      \texttt{mmlu}    &  0.0160 &  0.0154 &  0.0156 &  0.0152 \\
      \midrule
      \multicolumn{5}{c}{\texttt{LLAMA3\_70B}} \\
      \midrule
      \texttt{mnli}    &  0.0068 &  0.0068 &  0.0051 & -0.0017 \\
      \texttt{ag\_news}&  0.0540 &  0.0535 &  0.0544 &  0.0508 \\
      \texttt{arc}     & -0.6715 & -0.6545 & -0.6630 & -0.6630 \\
      \texttt{mmlu}    & -0.3540 & -0.3393 & -0.3629 & -0.3422 \\
      \bottomrule
    \end{tabular}
    
    \subcaption{Mistral\_7B, Mistral\_8×7B, LLAMA3\_3B, LLAMA3\_8B, LLAMA3\_70B}
    \label{tab:cohere32b_transition_b}
  \end{subtable}
  \caption{Transition metrics for four benchmarks (\texttt{MNLI, AG News, ARC}, and \texttt{MMLU}) across ten LLMs under different in‐context demonstration placements. For each model and dataset, the entry shows the performance delta (relative to the zero‐shot baseline) under each placement strategy: \texttt{ssp}, \texttt{esp}, \texttt{sum}, and \texttt{eum}.}
  \label{tab:cohere32b_transition}
\end{table*}

\begin{table*}[!httbp]
  \centering
  \renewcommand{\arraystretch}{0.85}
  \begin{subtable}[t]{0.48\textwidth}
    \centering
    {\small
    \begin{tabular}{@{}lcccc@{}}
      \toprule
      \textbf{Dataset}           & \textbf{ssp} & \textbf{esp} & \textbf{sum} & \textbf{eum} \\
      \midrule
      \multicolumn{5}{c}{\texttt{Qwen\_1.5B}} \\
      \midrule
      \texttt{mnli}              & 0.4488  & 0.4426  & 0.4364  & 0.4442  \\
      \texttt{xsum}              & 0.0167  & 0.0139  & 0.0236  & -0.0263 \\
      \texttt{squad}             & 0.0089  & 0.1059  & 0.0422  & -0.6020 \\
      \texttt{gsm8k}             & 9.2178  & 12.1266 & 9.6114  & -0.7547 \\
      \texttt{ag\_news}          & 0.3752  & 0.3896  & 0.4088  & 0.4688  \\
      \texttt{cnn\_dailymail}    & -0.0136 & -0.0319 & -0.0114 & -0.1755 \\
      \texttt{arc}               & 0.4426  & 0.4412  & 0.4468  & 0.4636  \\
      \texttt{mmlu}              & 0.5000  & 0.4970  & 0.5000  & 0.5060  \\
      \midrule
      \multicolumn{5}{c}{\texttt{Qwen\_7B}} \\
      \midrule
      \texttt{mnli}              & 0.4208  & 0.4280  & 0.4304  & 0.4088  \\
      \texttt{xsum}              & 0.0344  & 0.0614  & 0.0581  & -0.0081 \\
      \texttt{squad}             & 0.0646  & 0.0846  & 0.0660  & -0.0927 \\
      \texttt{gsm8k}             & 7.7581  & 14.5307 & 14.2375 & 13.8742 \\
      \texttt{ag\_news}          & 0.4447  & 0.4473  & 0.4473  & 0.4482  \\
      \texttt{cnn\_dailymail}    & 0.0164  & 0.0170  & 0.0199  & -0.0956 \\
      \texttt{arc}               & 0.5780  & 0.5770  & 0.5780  & 0.5690  \\
      \texttt{mmlu}              & 0.5021  & 0.5020  & 0.5019  & 0.4991  \\
      \midrule
      \multicolumn{5}{c}{\texttt{Qwen\_72B}} \\
      \midrule
      \texttt{mnli}              & 0.4456  & 0.4440  & 0.4440  & 0.4456  \\
      \texttt{xsum}              & 0.0440  & 0.0767  & 0.0999  & 0.0345  \\
      \texttt{squad}             & 0.0420  & 0.0489  & 0.0515  & 0.0340  \\
      \texttt{gsm8k}             & 9.7468  & 9.7755  & 9.8384  & 10.0803 \\
      \texttt{ag\_news}          & 0.5434  & 0.5441  & 0.5455  & 0.5441  \\
      \texttt{cnn\_dailymail}    & 0.0299  & 0.0238  & 0.0173  & 0.0088  \\
      \texttt{arc}               & 0.1885  & 0.1885  & 0.1850  & 0.1850  \\
      \texttt{mmlu}              & 0.2987  & 0.2987  & 0.3079  & 0.3048  \\
      \midrule
      \multicolumn{5}{c}{\texttt{Cohere\_8B}} \\
      \midrule
      \texttt{mnli}              & 0.4116  & 0.4116  & 0.4116  & 0.4116  \\
      \texttt{xsum}              & 0.1092  & 0.1615  & 0.1727  & -0.0324 \\
      \texttt{squad}             & 0.0502  & 0.0421  & 0.0318  & -0.8643 \\
      \texttt{gsm8k}             & 44.3362 & 52.2347 & 52.2888 & 19.5249 \\
      \texttt{ag\_news}          & 0.8072  & 0.7832  & 0.7736  & 0.7736  \\
      \texttt{cnn\_dailymail}    & 0.0442  & 0.0321  & 0.0340  & -0.1237 \\
      \texttt{arc}               & 0.4204  & 0.4258  & 0.4244  & 0.4379  \\
      \texttt{mmlu}              & 0.2607  & 0.2690  & 0.2663  & 0.7475  \\
      \midrule
      \multicolumn{5}{c}{\texttt{Cohere\_32B}} \\
      \midrule
      \texttt{mnli}              & 0.3736  & 0.3854  & 0.3657  & 0.3736  \\
      \texttt{xsum}              & 0.1149  & 0.1607  & 0.1659  & -0.0528 \\
      \texttt{squad}             & 0.0129  & 0.0254  & 0.0401  & -0.0274 \\
      \texttt{gsm8k}             & 16.7731 & 23.5865 & 23.8346 & 23.2584 \\
      \texttt{ag\_news}          & 0.6811  & 0.7698  & 0.7591  & 0.6882  \\
      \texttt{cnn\_dailymail}    & 0.0069  & 0.0239  & 0.0336  & -0.0546 \\
      \texttt{arc}               & 0.3164  & 0.3164  & 0.3191  & 0.3056  \\
      \texttt{mmlu}              & 0.2102  & 0.2039  & 0.2070  & 0.2701  \\
      \bottomrule
    \end{tabular}
    }
    \subcaption{Qwen\_1.5B, Qwen\_7B, Qwen\_72B, Cohere\_8B, \\ Cohere\_32B}
    \label{tab:transition_a}
  \end{subtable}
  \hfill
  \begin{subtable}[t]{0.48\textwidth}
    \centering
    {\small
    \begin{tabular}{@{}lcccc@{}}
      \toprule
      \textbf{Dataset}           & \textbf{ssp} & \textbf{esp} & \textbf{sum} & \textbf{eum} \\
      \midrule
      \multicolumn{5}{c}{\texttt{Mistral\_7B}} \\
      \midrule
      \texttt{mnli}              & 0.3550  & 0.3650  & 0.3550  & 0.3450  \\
      \texttt{xsum}              & 0.0182  & 0.0240  & 0.0212  & -0.0473 \\
      \texttt{squad}             & 0.1141  & 0.0557  & 0.0143  & -0.1928 \\
      \texttt{gsm8k}             & 15.4169 & 19.4560 & 19.2118 & 14.8845 \\
      \texttt{ag\_news}          & 0.7277  & 0.7070  & 0.7139  & 0.7173  \\
      \texttt{cnn\_dailymail}    & -0.0015 & 0.0064  & 0.0025  & -0.1621 \\
      \texttt{arc}               & 0.6202  & 0.6291  & 0.6246  & 0.5667  \\
      \texttt{mmlu}              & 0.4290  & 0.4645  & 0.4716  & 0.3544  \\
      \midrule
      \multicolumn{5}{c}{\texttt{Mistral\_8×7B}} \\
      \midrule
      \texttt{mnli}              & 0.4262  & 0.4283  & 0.4283  & 0.4262  \\
      \texttt{xsum}              & 0.0411  & 0.0313  & 0.0329  & 0.0155  \\
      \texttt{squad}             & 0.1719  & 0.1043  & 0.0814  & -0.0465 \\
      \texttt{gsm8k}             & 6.6051  & 7.7051  & 7.7045  & 10.2305 \\
      \texttt{ag\_news}          & 0.4232  & 0.4159  & 0.4174  & 0.4116  \\
      \texttt{cnn\_dailymail}    & 0.0196  & 0.0002  & -0.0071 & 0.0144  \\
      \texttt{arc}               & 0.6473  & 0.7850  & 0.7280  & 0.4620  \\
      \texttt{mmlu}              & 0.5337  & 0.5428  & 0.5270  & 0.3313  \\
      \midrule
      \multicolumn{5}{c}{\texttt{LLAMA3\_3B}} \\
      \midrule
      \texttt{mnli}              & 0.4388  & 0.4352  & 0.4280  & 0.4352  \\
      \texttt{xsum}              & 0.0299  & 0.0605  & 0.0842  & -0.0204 \\
      \texttt{squad}             & -0.0249 & -0.0080 & -0.0599 & -0.0742 \\
      \texttt{gsm8k}             & 6.0728  & 5.8630  & 6.4302  & 1.3402  \\
      \texttt{ag\_news}          & 0.4439  & 0.4483  & 0.4516  & 0.4549  \\
      \texttt{cnn\_dailymail}    & 0.0070  & 0.0080  & 0.0051  & -0.1837 \\
      \texttt{arc}               & 0.3542  & 0.3515  & 0.3704  & 0.3974  \\
      \texttt{mmlu}              & 0.4858  & 0.4872  & 0.4895  & 0.5397  \\
      \midrule
      \multicolumn{5}{c}{\texttt{LLAMA3\_8B}} \\
      \midrule
      \texttt{mnli}              & 0.4514  & 0.4442  & 0.4514  & 0.4424  \\
      \texttt{xsum}              & 0.0743  & 0.1038  & 0.0994  & 0.0212  \\
      \texttt{squad}             & 0.0394  & 0.0442  & 0.0504  & -0.0027 \\
      \texttt{gsm8k}             & 25.1482 & 26.2461 & 26.0143 & 28.1443 \\
      \texttt{ag\_news}          & 0.2336  & 0.2336  & 0.2624  & 0.2408  \\
      \texttt{cnn\_dailymail}    & -0.0004 & 0.0035  & 0.0074  & -0.0125 \\
      \texttt{arc}               & 0.3852  & 0.3791  & 0.3811  & 0.3975  \\
      \texttt{mmlu}              & 0.5036  & 0.5030  & 0.5032  & 0.5028  \\
      \midrule
      \multicolumn{5}{c}{\texttt{LLAMA3\_70B}} \\
      \midrule
      \texttt{mnli}              & 0.4490  & 0.4490  & 0.4473  & 0.4405  \\
      \texttt{xsum}              & 0.0922  & 0.1280  & 0.1531  & 0.0664  \\
      \texttt{squad}             & 0.0172  & 0.0176  & 0.0347  & 0.0125  \\
      \texttt{gsm8k}             & 0.0691  & 0.0406  & 1.1734  & 6.5517  \\
      \texttt{ag\_news}          & 0.5306  & 0.5302  & 0.5311  & 0.5274  \\
      \texttt{cnn\_dailymail}    & 0.0156  & 0.0172  & 0.0158  & 0.0210  \\
      \texttt{arc}               & 0.1387  & 0.1557  & 0.1472  & 0.1472  \\
      \texttt{mmlu}              & 0.3289  & 0.3437  & 0.3200  & 0.3407  \\
      \bottomrule
    \end{tabular}
    }
    \subcaption{Mistral\_7B, Mistral\_8×7B, LLAMA3\_3B, LLAMA3\_8B, LLAMA3\_70B}
    \label{tab:transition_b}
  \end{subtable}
  \caption{Comprehensive transition metrics for eight benchmarks (\emph{MNLI, XSUM, SQuAD, GSM8K, AG News, CNN/DailyMail, ARC}, and \emph{MMLU}) across ten LLMs and four demonstration placements. Each cell reports the change in performance relative to zero‐shot when demos are placed at the start/end of the system prompt or the start/end of the user message (\texttt{ssp}, \texttt{esp}, \texttt{sum}, \texttt{eum}).}
  \label{tab:full_transition}
\end{table*}

\begin{table*}[!httbp]
\centering
\footnotesize
\setlength{\tabcolsep}{3pt}
\renewcommand{\arraystretch}{1.1}
\begin{tabular}{lcccccccc}
\toprule
\multicolumn{9}{c}{\textbf{MNLI}} \\
\midrule
\multirow{2}{*}{\textbf{Position}} & \textbf{$\Delta_\text{pred}$ (\%)} & \textbf{Improved (\%)} & \textbf{Regressed (\%)} & \textbf{Net $\Delta_\text{pred}$}
& \textbf{$\Delta_\text{pred}$ (\%)} & \textbf{Improved (\%)} & \textbf{Regressed (\%)} & \textbf{Net $\Delta_\text{pred}$} \\
\cmidrule(lr){2-5} \cmidrule(lr){6-9}
\multicolumn{4}{c}{\texttt{Qwen\_1.5B}} & \multicolumn{4}{c}{\texttt{Qwen\_7B}} \\
\midrule
ssp & 38.50 & 19.00 & 19.50 & -1 & 37.00 & 18.50 & 18.50 & 0 \\
esp & 34.50 & 16.00 & 18.50 & -5 & 34.50 & 18.00 & 16.50 & 3 \\
sum & 38.50 & 17.00 & 21.50 & -9 & 37.00 & 19.50 & 17.50 & 4 \\
eum & 6.00 & 2.00 & 4.00 & -4 & 29.50 & 13.50 & 16.00 & -5 \\
\midrule
\multicolumn{4}{c}{\texttt{Qwen\_72B}} & \multicolumn{4}{c}{\texttt{Cohere\_8B}} \\
\midrule
ssp & 9.00 & 6.00 & 3.00 & 6 & 13.00 & 8.50 & 4.50 & 8 \\
esp & 9.50 & 6.00 & 3.50 & 5 & 19.00 & 11.50 & 7.50 & 8 \\
sum & 9.50 & 6.00 & 3.50 & 5 & 14.00 & 9.00 & 5.00 & 8 \\
eum & 8.00 & 5.50 & 2.50 & 6 & 12.00 & 8.00 & 4.00 & 8 \\
\midrule
\multicolumn{4}{c}{\texttt{Cohere\_32B}} & \multicolumn{4}{c}{\texttt{Mistral\_7B}} \\
\midrule
ssp & 13.00 & 6.50 & 6.50 & 0 & 34.00 & 20.00 & 14.00 & 12 \\
esp & 15.50 & 8.50 & 7.00 & 3 & 33.00 & 20.00 & 13.00 & 14 \\
sum & 10.00 & 4.50 & 5.50 & -2 & 32.00 & 19.00 & 13.00 & 12 \\
eum & 12.00 & 6.00 & 6.00 & 0 & 26.00 & 15.50 & 10.50 & 10 \\
\midrule
\multicolumn{4}{c}{\texttt{Mistral\_8x7B}} & \multicolumn{4}{c}{\texttt{LLAMA3\_3B}} \\
\midrule
ssp & 5.00 & 2.00 & 3.00 & -2 & 41.50 & 20.50 & 21.00 & -1 \\
esp & 4.50 & 2.00 & 2.50 & -1 & 30.50 & 14.50 & 16.00 & -3 \\
sum & 4.50 & 2.00 & 2.50 & -1 & 18.50 & 7.50 & 11.00 & -7 \\
eum & 9.50 & 4.50 & 5.00 & -1 & 24.50 & 11.50 & 13.00 & -3 \\
\midrule
\multicolumn{4}{c}{\texttt{LLAMA3\_8B}} & \multicolumn{4}{c}{\texttt{LLAMA3\_70B}} \\
\midrule
ssp & 23.00 & 13.50 & 9.50 & 8 & 58.50 & 30.00 & 28.50 & 3 \\
esp & 24.00 & 13.00 & 11.00 & 4 & 56.50 & 29.00 & 27.50 & 3 \\
sum & 33.00 & 18.50 & 14.50 & 8 & 58.00 & 29.50 & 28.50 & 2 \\
eum & 20.50 & 11.00 & 9.50 & 3 & 59.00 & 29.00 & 30.00 & -2 \\
\bottomrule
\end{tabular}
\caption{Delta metrics on the \texttt{MNLI} benchmark across ten LLMs and four \texttt{DPP}s. For each \texttt{DPP}, we report: (1) the percentage of examples whose predicted answer changed, (2) the percentage that improved, (3) the percentage that regressed, and (4) the Net $\Delta_\text{pred}$ (cnt.) (Total Count Improved – Total Count Regressed), all measured relative to the \textit{sum} configuration.}
\label{tab:mnli_delta_table}
\end{table*}

\begin{table*}[!httbp]
\centering
\footnotesize
\setlength{\tabcolsep}{3pt}
\renewcommand{\arraystretch}{1.1}
\begin{tabular}{lcccccccc}
\toprule
\multicolumn{9}{c}{\textbf{XSUM}} \\
\midrule
\multirow{2}{*}{\textbf{Position}} & \textbf{$\Delta_\text{pred}$ (\%)} & \textbf{Improved (\%)} & \textbf{Regressed (\%)} & \textbf{Net $\Delta_\text{pred}$}
& \textbf{$\Delta_\text{pred}$ (\%)} & \textbf{Improved (\%)} & \textbf{Regressed (\%)} & \textbf{Net $\Delta_\text{pred}$} \\
\cmidrule(lr){2-5} \cmidrule(lr){6-9}
\multicolumn{4}{c}{\texttt{Qwen\_1.5B}} & \multicolumn{4}{c}{\texttt{Qwen\_7B}} \\
\midrule
ssp & 0.00 & 92.50 & 0.00 & 185 & 0.00 & 99.00 & 0.00 & 198 \\
esp & 0.00 & 90.50 & 0.50 & 180 & 0.00 & 99.00 & 0.00 & 198 \\
sum & 0.00 & 91.50 & 0.00 & 183 & 0.00 & 99.00 & 0.00 & 198 \\
eum & 0.00 & 69.00 & 0.00 & 138 & 0.00 & 92.50 & 0.00 & 185 \\
\midrule
\multicolumn{4}{c}{\texttt{Qwen\_72B}} & \multicolumn{4}{c}{\texttt{Cohere\_8B}} \\
\midrule
ssp & 0.00 & 99.00 & 0.00 & 198 & 0.00 & 99.00 & 0.00 & 198 \\
esp & 0.00 & 98.50 & 0.00 & 197 & 0.00 & 97.50 & 0.50 & 194 \\
sum & 0.00 & 98.50 & 0.00 & 197 & 0.00 & 98.50 & 0.00 & 197 \\
eum & 0.00 & 98.00 & 0.50 & 195 & 0.00 & 78.00 & 1.00 & 154 \\
\midrule
\multicolumn{4}{c}{\texttt{Cohere\_32B}} & \multicolumn{4}{c}{\texttt{Mistral\_7B}} \\
\midrule
ssp & 0.00 & 99.00 & 0.00 & 198 & 0.00 & 92.00 & 0.00 & 184 \\
esp & 0.00 & 99.00 & 0.00 & 198 & 0.00 & 90.50 & 0.00 & 181 \\
sum & 0.00 & 99.50 & 0.00 & 199 & 0.00 & 88.50 & 0.00 & 177 \\
eum & 0.00 & 88.50 & 0.00 & 177 & 0.00 & 42.00 & 0.50 & 83 \\
\midrule
\multicolumn{4}{c}{\texttt{Mistral\_8x7B}} & \multicolumn{4}{c}{\texttt{LLAMA3\_3B}} \\
\midrule
ssp & 0.00 & 94.00 & 0.00 & 188 & 0.00 & 99.50 & 0.00 & 199 \\
esp & 0.00 & 94.00 & 0.00 & 188 & 0.00 & 97.00 & 0.00 & 194 \\
sum & 0.00 & 95.00 & 0.00 & 190 & 0.00 & 98.50 & 0.00 & 197 \\
eum & 0.00 & 91.50 & 0.00 & 183 & 0.00 & 90.50 & 0.00 & 181 \\
\midrule
\multicolumn{4}{c}{\texttt{LLAMA3\_8B}} & \multicolumn{4}{c}{\texttt{LLAMA3\_70B}} \\
\midrule
ssp & 0.00 & 98.00 & 0.00 & 196 & 0.00 & 99.50 & 0.00 & 199 \\
esp & 0.00 & 98.50 & 0.00 & 197 & 0.00 & 99.50 & 0.00 & 199 \\
sum & 0.00 & 97.50 & 0.00 & 195 & 0.00 & 99.50 & 0.00 & 199 \\
eum & 0.00 & 97.50 & 0.00 & 195 & 0.00 & 99.50 & 0.00 & 199 \\
\bottomrule
\end{tabular}

\caption{Delta metrics on the \texttt{XSUM} benchmark across ten LLMs and four \texttt{DPP}s. For each \texttt{DPP}, we report: (1) the percentage of examples whose predicted answer changed, (2) the percentage that improved, (3) the percentage that regressed, and (4) the Net $\Delta_\text{pred}$ (cnt.) (Total Count Improved – Total Count Regressed), all measured relative to the \textit{sum} configuration.}
\label{tab:xsum_delta_table}
\end{table*}

\begin{table*}[!httbp]
\centering
\footnotesize
\setlength{\tabcolsep}{3pt}
\renewcommand{\arraystretch}{1.1}
\begin{tabular}{lcccccccc}
\toprule
\multicolumn{9}{c}{\textbf{SQUAD}} \\
\midrule
\multirow{2}{*}{\textbf{Position}} & \textbf{$\Delta_\text{pred}$ (\%)} & \textbf{Improved (\%)} & \textbf{Regressed (\%)} & \textbf{Net $\Delta_\text{pred}$}
& \textbf{$\Delta_\text{pred}$ (\%)} & \textbf{Improved (\%)} & \textbf{Regressed (\%)} & \textbf{Net $\Delta_\text{pred}$} \\
\cmidrule(lr){2-5} \cmidrule(lr){6-9}
\multicolumn{4}{c}{\texttt{Qwen\_1.5B}} & \multicolumn{4}{c}{\texttt{Qwen\_7B}} \\
\midrule
ssp & 0.00 & 23.00 & 26.50 & -7 & 0.00 & 19.50 & 7.00 & 25 \\
esp & 0.00 & 27.00 & 36.50 & -19 & 0.00 & 20.00 & 3.50 & 33 \\
sum & 0.00 & 24.50 & 27.50 & -6 & 0.00 & 21.00 & 5.50 & 31 \\
eum & 0.00 & 13.00 & 56.50 & -87 & 0.00 & 19.50 & 20.00 & -1 \\
\midrule
\multicolumn{4}{c}{\texttt{Qwen\_72B}} & \multicolumn{4}{c}{\texttt{Cohere\_8B}} \\
\midrule
ssp & 0.00 & 16.50 & 4.50 & 24 & 0.00 & 20.50 & 6.50 & 28 \\
esp & 0.00 & 16.00 & 4.00 & 24 & 0.00 & 16.00 & 9.00 & 14 \\
sum & 0.00 & 17.00 & 4.50 & 25 & 0.00 & 15.50 & 8.00 & 15 \\
eum & 0.00 & 13.50 & 5.50 & 16 & 0.00 & 9.50 & 75.50 & -132 \\
\midrule
\multicolumn{4}{c}{\texttt{Cohere\_32B}} & \multicolumn{4}{c}{\texttt{Mistral\_7B}} \\
\midrule
ssp & 0.00 & 13.00 & 7.00 & 12 & 0.00 & 43.00 & 6.00 & 74 \\
esp & 0.00 & 16.00 & 8.00 & 16 & 0.00 & 42.00 & 7.00 & 70 \\
sum & 0.00 & 16.50 & 7.00 & 19 & 0.00 & 38.50 & 9.00 & 59 \\
eum & 0.00 & 16.00 & 15.00 & 2 & 0.00 & 32.00 & 20.00 & 24 \\
\midrule\multicolumn{4}{c}{\texttt{Mistral\_8x7B}} & \multicolumn{4}{c}{\texttt{LLAMA3\_3B}} \\
\midrule
ssp & 0.00 & 42.50 & 3.00 & 79 & 0.00 & 19.50 & 13.00 & 13 \\
esp & 0.00 & 34.50 & 5.50 & 58 & 0.00 & 22.50 & 9.00 & 27 \\
sum & 0.00 & 37.00 & 8.00 & 58 & 0.00 & 22.00 & 18.50 & 7 \\
eum & 0.00 & 27.00 & 10.50 & 33 & 0.00 & 22.50 & 18.50 & 8 \\
\midrule\multicolumn{4}{c}{\texttt{LLAMA3\_8B}} & \multicolumn{4}{c}{\texttt{LLAMA3\_70B}} \\
\midrule
ssp & 0.00 & 16.50 & 5.50 & 22 & 0.00 & 14.50 & 5.00 & 19 \\
esp & 0.00 & 17.00 & 5.00 & 24 & 0.00 & 15.00 & 5.50 & 19 \\
sum & 0.00 & 19.50 & 6.50 & 26 & 0.00 & 17.00 & 5.00 & 24 \\
eum & 0.00 & 15.00 & 12.50 & 5 & 0.00 & 18.00 & 8.00 & 20 \\
\bottomrule
\end{tabular}
\caption{Delta metrics on the \texttt{SQUAD} benchmark across ten LLMs and four \texttt{DPP}s. For each \texttt{DPP}, we report: (1) the percentage of examples whose predicted answer changed, (2) the percentage that improved, (3) the percentage that regressed, and (4) the Net $\Delta_\text{pred}$ (cnt.) (Total Count Improved – Total Count Regressed), all measured relative to the \textit{sum} configuration.}
\label{tab:squad_delta_table}
\end{table*}

\begin{table*}[!httbp]
\centering
\footnotesize
\setlength{\tabcolsep}{3pt}
\renewcommand{\arraystretch}{1.1}
\begin{tabular}{lcccccccc}
\toprule
\multicolumn{9}{c}{\textbf{GSM8K}} \\
\midrule
\multirow{2}{*}{\textbf{Position}} & \textbf{$\Delta_\text{pred}$ (\%)} & \textbf{Improved (\%)} & \textbf{Regressed (\%)} & \textbf{Net $\Delta_\text{pred}$}
& \textbf{$\Delta_\text{pred}$ (\%)} & \textbf{Improved (\%)} & \textbf{Regressed (\%)} & \textbf{Net $\Delta_\text{pred}$} \\
\cmidrule(lr){2-5} \cmidrule(lr){6-9}
\multicolumn{4}{c}{\texttt{Qwen\_1.5B}} & \multicolumn{4}{c}{\texttt{Qwen\_7B}} \\
\midrule
ssp & 0.00 & 35.50 & 9.00 & 53 & 0.00 & 62.00 & 0.50 & 123 \\
esp & 0.00 & 42.50 & 5.50 & 74 & 0.00 & 100.00 & 0.00 & 200 \\
sum & 0.00 & 34.50 & 9.50 & 50 & 0.00 & 100.00 & 0.00 & 200 \\
eum & 0.00 & 0.50 & 15.50 & -30 & 0.00 & 95.00 & 0.00 & 190 \\
\midrule\multicolumn{4}{c}{\texttt{Qwen\_72B}} & \multicolumn{4}{c}{\texttt{Cohere\_8B}} \\
\midrule
ssp & 0.00 & 100.00 & 0.00 & 200 & 0.00 & 91.50 & 0.50 & 182 \\
esp & 0.00 & 100.00 & 0.00 & 200 & 0.00 & 100.00 & 0.00 & 200 \\
sum & 0.00 & 100.00 & 0.00 & 200 & 0.00 & 100.00 & 0.00 & 200 \\
eum & 0.00 & 100.00 & 0.00 & 200 & 0.00 & 53.50 & 4.00 & 99 \\
\midrule\multicolumn{4}{c}{\texttt{Cohere\_32B}} & \multicolumn{4}{c}{\texttt{Mistral\_7B}} \\
\midrule
ssp & 0.00 & 73.00 & 1.50 & 143 & 0.00 & 96.50 & 0.50 & 192 \\
esp & 0.00 & 98.00 & 0.00 & 196 & 0.00 & 99.50 & 0.00 & 199 \\
sum & 0.00 & 99.50 & 0.00 & 199 & 0.00 & 99.50 & 0.00 & 199 \\
eum & 0.00 & 99.50 & 0.00 & 199 & 0.00 & 99.50 & 0.00 & 199 \\
\midrule\multicolumn{4}{c}{\texttt{Mistral\_8x7B}} & \multicolumn{4}{c}{\texttt{LLAMA3\_3B}} \\
\midrule
ssp & 0.00 & 62.00 & 0.00 & 124 & 0.00 & 97.50 & 0.00 & 195 \\
esp & 0.00 & 73.50 & 0.50 & 146 & 0.00 & 95.50 & 2.00 & 187 \\
sum & 0.00 & 73.00 & 2.00 & 142 & 0.00 & 100.00 & 0.00 & 200 \\
eum & 0.00 & 91.00 & 0.50 & 181 & 0.00 & 73.50 & 3.00 & 141 \\
\midrule\multicolumn{4}{c}{\texttt{LLAMA3\_8B}} & \multicolumn{4}{c}{\texttt{LLAMA3\_70B}} \\
\midrule
ssp & 0.00 & 100.00 & 0.00 & 200 & 0.00 & 12.00 & 2.00 & 20 \\
esp & 0.00 & 99.50 & 0.00 & 199 & 0.00 & 13.00 & 1.50 & 23 \\
sum & 0.00 & 100.00 & 0.00 & 200 & 0.00 & 32.00 & 1.00 & 62 \\
eum & 0.00 & 100.00 & 0.00 & 200 & 0.00 & 92.50 & 0.00 & 185 \\
\bottomrule
\end{tabular}
\caption{Delta metrics on the \texttt{Gsm8k} benchmark across ten LLMs and four \texttt{DPP}s. For each \texttt{DPP}, we report: (1) the percentage of examples whose predicted answer changed, (2) the percentage that improved, (3) the percentage that regressed, and (4) the Net $\Delta_\text{pred}$ (cnt.) (Total Count Improved – Total Count Regressed), all measured relative to the \textit{sum} configuration.}
\label{tab:gsm8k_delta_table}
\end{table*}

\begin{table*}[!httbp]
\centering
\footnotesize
\setlength{\tabcolsep}{3pt}
\renewcommand{\arraystretch}{1.1}
\begin{tabular}{lcccccccc}
\toprule
\multicolumn{9}{c}{\textbf{AG\_NEWS}}\\
\midrule
\multirow{2}{*}{\textbf{Position}} & \textbf{$\Delta_\text{pred}$ (\%)} & \textbf{Improved (\%)} & \textbf{Regressed (\%)} & \textbf{Net $\Delta_\text{pred}$} & \textbf{$\Delta_\text{pred}$ (\%)} & \textbf{Improved (\%)} & \textbf{Regressed (\%)} & \textbf{Net $\Delta_\text{pred}$}\\
\cmidrule(lr){2-5} \cmidrule(lr){6-9}
\multicolumn{4}{c}{\texttt{Qwen\_1.5B}} & \multicolumn{4}{c}{\texttt{Qwen\_7B}}\\
\midrule
ssp & 7.50 & 3.50 & 4.00 & -1 & 5.00 & 1.50 & 3.50 & -4\\
esp & 10.50 & 3.50 & 7.00 & -7 & 5.50 & 1.00 & 4.50 & -7\\
sum & 19.50 & 6.00 & 13.50 & -15 & 6.50 & 1.50 & 5.00 & -7\\
eum & 46.00 & 13.00 & 33.00 & -40 & 7.00 & 1.50 & 5.50 & -8\\
\midrule
\multicolumn{4}{c}{\texttt{Qwen\_72B}} & \multicolumn{4}{c}{\texttt{Cohere\_8B}}\\
\midrule
ssp & 3.50 & 1.00 & 2.50 & -3 & 11.00 & 7.00 & 4.00 & 6\\
esp & 2.00 & 0.50 & 1.50 & -2 & 11.50 & 6.00 & 5.50 & 1\\
sum & 2.00 & 1.00 & 1.00 & 0 & 11.50 & 5.50 & 6.00 & -1\\
eum & 2.00 & 0.50 & 1.50 & -2 & 17.50 & 8.50 & 9.00 & -1\\
\midrule
\multicolumn{4}{c}{\texttt{Cohere\_32B}} & \multicolumn{4}{c}{\texttt{Mistral\_7B}}\\
\midrule
ssp & 20.00 & 5.50 & 14.50 & -18 & 16.50 & 11.50 & 5.00 & 13\\
esp & 8.50 & 6.00 & 2.50 & 7 & 11.50 & 7.50 & 4.00 & 7\\
sum & 9.00 & 5.50 & 3.50 & 4 & 12.50 & 8.50 & 4.00 & 9\\
eum & 15.00 & 3.50 & 11.50 & -16 & 14.00 & 9.50 & 4.50 & 10\\
\midrule
\multicolumn{4}{c}{\texttt{Mistral\_8x7B}} & \multicolumn{4}{c}{\texttt{LLAMA3\_3B}}\\
\midrule
ssp & 8.50 & 3.50 & 5.00 & -3 & 14.50 & 9.00 & 5.50 & 7\\
esp & 11.00 & 3.50 & 7.50 & -8 & 13.50 & 7.50 & 6.00 & 3\\
sum & 11.00 & 3.50 & 7.50 & -8 & 14.00 & 7.00 & 7.00 & 0\\
eum & 8.50 & 3.00 & 5.50 & -5 & 15.50 & 7.00 & 8.50 & -3\\
\midrule
\multicolumn{4}{c}{\texttt{LLAMA3\_8B}} & \multicolumn{4}{c}{\texttt{LLAMA3\_70B}}\\
\midrule
ssp & 9.00 & 5.50 & 3.50 & 4 & 10.50 & 9.50 & 1.00 & 17\\
esp & 9.00 & 5.50 & 3.50 & 4 & 10.00 & 9.00 & 1.00 & 16\\
sum & 12.00 & 5.00 & 7.00 & -4 & 11.00 & 10.00 & 1.00 & 18\\
eum & 8.00 & 4.50 & 3.50 & 2 & 7.00 & 6.00 & 1.00 & 10\\
\bottomrule
\end{tabular}
\caption{Delta metrics on the \texttt{Ag News} benchmark across ten LLMs and four \texttt{DPP}s. For each \texttt{DPP}, we report: (1) the percentage of examples whose predicted answer changed, (2) the percentage that improved, (3) the percentage that regressed, and (4) the Net $\Delta_\text{pred}$ (cnt.) (Total Count Improved – Total Count Regressed), all measured relative to the \textit{sum} configuration.}
\label{tab:ag_news_delta_table}
\end{table*}
\begin{table*}[!httbp]
\centering
\footnotesize
\setlength{\tabcolsep}{3pt}
\renewcommand{\arraystretch}{1.1}
\begin{tabular}{lcccccccc}
\toprule
\multicolumn{9}{c}{\textbf{CNN\_DAILYMAIL}} \\
\midrule
\multirow{2}{*}{\textbf{Position}} & \textbf{$\Delta_\text{pred}$ (\%)} & \textbf{Improved (\%)} & \textbf{Regressed (\%)} & \textbf{Net $\Delta_\text{pred}$}
& \textbf{$\Delta_\text{pred}$ (\%)} & \textbf{Improved (\%)} & \textbf{Regressed (\%)} & \textbf{Net $\Delta_\text{pred}$} \\
\cmidrule(lr){2-5} \cmidrule(lr){6-9}
\multicolumn{4}{c}{\texttt{Qwen\_1.5B}} & \multicolumn{4}{c}{\texttt{Qwen\_7B}} \\
\midrule
ssp & 0.00 & 86.00 & 5.50 & 161 & 0.00 & 94.50 & 1.50 & 186 \\
esp & 0.00 & 80.00 & 9.50 & 141 & 0.00 & 94.00 & 0.50 & 187 \\
sum & 0.00 & 87.50 & 6.00 & 163 & 0.00 & 92.50 & 1.00 & 183 \\
eum & 0.00 & 0.00 & 13.00 & -26 & 0.00 & 42.50 & 7.00 & 71 \\
\midrule\multicolumn{4}{c}{\texttt{Qwen\_72B}} & \multicolumn{4}{c}{\texttt{Cohere\_8B}} \\
\midrule
ssp & 0.00 & 95.50 & 0.00 & 191 & 0.00 & 94.00 & 1.50 & 185 \\
esp & 0.00 & 95.50 & 0.00 & 191 & 0.00 & 91.50 & 4.00 & 175 \\
sum & 0.00 & 95.50 & 0.00 & 191 & 0.00 & 91.00 & 1.50 & 179 \\
eum & 0.00 & 94.00 & 0.00 & 188 & 0.00 & 31.50 & 7.00 & 49 \\
\midrule\multicolumn{4}{c}{\texttt{Cohere\_32B}} & \multicolumn{4}{c}{\texttt{Mistral\_7B}} \\
\midrule
ssp & 0.00 & 99.00 & 0.00 & 198 & 0.00 & 93.50 & 3.50 & 180 \\
esp & 0.00 & 94.50 & 3.00 & 183 & 0.00 & 94.50 & 2.00 & 185 \\
sum & 0.00 & 95.50 & 3.00 & 185 & 0.00 & 94.50 & 2.50 & 184 \\
eum & 0.00 & 81.00 & 0.50 & 161 & 0.00 & 12.00 & 0.50 & 23 \\
\midrule\multicolumn{4}{c}{\texttt{Mistral\_8x7B}} & \multicolumn{4}{c}{\texttt{LLAMA3\_3B}} \\
\midrule
ssp & 0.00 & 89.00 & 2.50 & 173 & 0.00 & 88.50 & 0.50 & 176 \\
esp & 0.00 & 88.00 & 4.50 & 167 & 0.00 & 91.00 & 1.00 & 180 \\
sum & 0.00 & 86.00 & 6.00 & 160 & 0.00 & 90.50 & 0.50 & 180 \\
eum & 0.00 & 89.95 & 4.02 & 171 & 0.00 & 2.50 & 20.00 & -35 \\
\midrule\multicolumn{4}{c}{\texttt{LLAMA3\_8B}} & \multicolumn{4}{c}{\texttt{LLAMA3\_70B}} \\
\midrule
ssp & 0.00 & 91.50 & 0.00 & 183 & 0.00 & 98.50 & 0.00 & 197 \\
esp & 0.00 & 89.00 & 3.00 & 172 & 0.00 & 98.50 & 0.00 & 197 \\
sum & 0.00 & 90.50 & 1.00 & 179 & 0.00 & 99.00 & 0.00 & 198 \\
eum & 0.00 & 90.00 & 1.00 & 178 & 0.00 & 99.00 & 0.00 & 198 \\
\bottomrule
\end{tabular}

\caption{Delta metrics on the \texttt{Cnn/Dailymail} benchmark across ten LLMs and four \texttt{DPP}s. For each \texttt{DPP}, we report: (1) the percentage of examples whose predicted answer changed, (2) the percentage that improved, (3) the percentage that regressed, and (4) the Net $\Delta_\text{pred}$ (cnt.) (Total Count Improved – Total Count Regressed), all measured relative to the \textit{sum} configuration.}
\label{tab:cnn_dailymail_delta_table}
\end{table*}

\begin{table*}[!httbp]
\centering
\footnotesize
\setlength{\tabcolsep}{3pt}
\renewcommand{\arraystretch}{1.1}
\begin{tabular}{lcccccccc}
\toprule
\multicolumn{9}{c}{\textbf{ARC}} \\
\midrule
\multirow{2}{*}{\textbf{Position}} & \textbf{$\Delta_\text{pred}$ (\%)} & \textbf{Improved (\%)} & \textbf{Regressed (\%)} & \textbf{Net $\Delta_\text{pred}$}
& \textbf{$\Delta_\text{pred}$ (\%)} & \textbf{Improved (\%)} & \textbf{Regressed (\%)} & \textbf{Net $\Delta_\text{pred}$} \\
\cmidrule(lr){2-5} \cmidrule(lr){6-9}
\multicolumn{4}{c}{\texttt{Qwen\_1.5B}} & \multicolumn{4}{c}{\texttt{Qwen\_7B}} \\
\midrule
ssp & 14.00 & 9.50 & 4.50 & 10 & 3.00 & 2.50 & 0.50 & 4 \\
esp & 14.00 & 9.50 & 4.50 & 10 & 3.50 & 2.50 & 1.00 & 3 \\
sum & 14.00 & 8.50 & 5.50 & 6 & 4.00 & 3.00 & 1.00 & 4 \\
eum & 15.50 & 6.50 & 9.00 & -5 & 3.50 & 1.50 & 2.00 & -1 \\
\midrule\multicolumn{4}{c}{\texttt{Qwen\_72B}} & \multicolumn{4}{c}{\texttt{Cohere\_8B}} \\
\midrule
ssp & 1.00 & 0.00 & 1.00 & -2 & 7.50 & 5.00 & 2.50 & 5 \\
esp & 1.00 & 0.00 & 1.00 & -2 & 10.50 & 5.50 & 5.00 & 1 \\
sum & 0.50 & 0.00 & 0.50 & -1 & 13.00 & 7.00 & 6.00 & 2 \\
eum & 0.50 & 0.00 & 0.50 & -1 & 10.00 & 3.50 & 6.50 & -6 \\
\midrule\multicolumn{4}{c}{\texttt{Cohere\_32B}} & \multicolumn{4}{c}{\texttt{Mistral\_7B}} \\
\midrule
ssp & 6.50 & 2.00 & 4.50 & -5 & 15.50 & 9.00 & 6.50 & 5 \\
esp & 5.50 & 1.50 & 4.00 & -5 & 12.00 & 7.50 & 4.50 & 6 \\
sum & 7.00 & 2.00 & 5.00 & -6 & 13.50 & 8.00 & 5.50 & 5 \\
eum & 3.50 & 1.50 & 2.00 & -1 & 12.00 & 4.00 & 8.00 & -8 \\
\midrule\multicolumn{4}{c}{\texttt{Mistral\_8x7B}} & \multicolumn{4}{c}{\texttt{LLAMA3\_3B}} \\
\midrule
ssp & 10.50 & 7.00 & 3.50 & 7 & 17.00 & 11.00 & 6.00 & 10 \\
esp & 10.50 & 7.00 & 3.50 & 7 & 14.50 & 10.00 & 4.50 & 11 \\
sum & 11.00 & 7.50 & 3.50 & 8 & 14.00 & 8.00 & 6.00 & 4 \\
eum & 12.50 & 8.00 & 4.50 & 7 & 16.00 & 6.50 & 9.50 & -6 \\
\midrule\multicolumn{4}{c}{\texttt{LLAMA3\_8B}} & \multicolumn{4}{c}{\texttt{LLAMA3\_70B}} \\
\midrule
ssp & 7.00 & 3.50 & 3.50 & 0 & 2.00 & 1.00 & 1.00 & 0 \\
esp & 8.50 & 5.00 & 3.50 & 3 & 2.00 & 0.00 & 2.00 & -4 \\
sum & 9.00 & 5.00 & 4.00 & 2 & 2.00 & 0.50 & 1.50 & -2 \\
eum & 8.00 & 2.50 & 5.50 & -6 & 1.00 & 0.00 & 1.00 & -2 \\
\bottomrule
\end{tabular}
\caption{Delta metrics on the ARC benchmark across ten LLMs and four \texttt{DPP}s. For each \texttt{DPP}, we report: (1) the percentage of examples whose predicted answer changed, (2) the percentage that improved, (3) the percentage that regressed, and (4) the Net $\Delta_\text{pred}$ (cnt.) (Total Count Improved – Total Count Regressed), all measured relative to the \textit{sum} configuration.}

\caption{Delta metrics on \texttt{arc} across models and \texttt{DPP}s. }
\label{tab:arc_delta_table}
\end{table*}

\begin{table*}[!httbp]
\centering
\footnotesize
\setlength{\tabcolsep}{3pt}
\renewcommand{\arraystretch}{1.1}
\begin{tabular}{lcccccccc}
\toprule
\multicolumn{9}{c}{\textbf{MMLU}} \\
\midrule
\multirow{2}{*}{\textbf{Position}} & \textbf{$\Delta_\text{pred}$ (\%)} & \textbf{Improved (\%)} & \textbf{Regressed (\%)} & \textbf{Net $\Delta_\text{pred}$}
& \textbf{$\Delta_\text{pred}$ (\%)} & \textbf{Improved (\%)} & \textbf{Regressed (\%)} & \textbf{Net $\Delta_\text{pred}$} \\
\cmidrule(lr){2-5} \cmidrule(lr){6-9}
\multicolumn{4}{c}{\texttt{Qwen\_1.5B}} & \multicolumn{4}{c}{\texttt{Qwen\_7B}} \\
\midrule
ssp & 19.00 & 8.00 & 11.00 & -6 & 11.50 & 7.50 & 4.00 & 7 \\
esp & 12.50 & 7.00 & 5.50 & 3 & 12.50 & 7.00 & 5.50 & 3 \\
sum & 17.00 & 7.00 & 10.00 & -6 & 10.50 & 5.50 & 5.00 & 1 \\
eum & 29.50 & 7.00 & 22.50 & -31 & 40.00 & 6.50 & 33.50 & -54 \\
\midrule\multicolumn{4}{c}{\texttt{Qwen\_72B}} & \multicolumn{4}{c}{\texttt{Cohere\_8B}} \\
\midrule
ssp & 4.50 & 2.50 & 2.00 & 1 & 11.50 & 10.00 & 1.50 & 17 \\
esp & 5.50 & 3.00 & 2.50 & 1 & 12.50 & 10.00 & 2.50 & 15 \\
sum & 6.00 & 2.50 & 3.50 & -2 & 12.50 & 10.00 & 2.50 & 15 \\
eum & 3.50 & 1.50 & 2.00 & -1 & 81.00 & 0.50 & 80.50 & -160 \\
\midrule\multicolumn{4}{c}{\texttt{Cohere\_32B}} & \multicolumn{4}{c}{\texttt{Mistral\_7B}} \\
\midrule
ssp & 5.50 & 3.00 & 2.50 & 1 & 13.50 & 5.00 & 8.50 & -7 \\
esp & 4.50 & 3.00 & 1.50 & 3 & 22.50 & 9.00 & 13.50 & -9 \\
sum & 4.00 & 2.50 & 1.50 & 2 & 23.00 & 9.50 & 13.50 & -8 \\
eum & 12.00 & 1.50 & 10.50 & -18 & 43.50 & 11.00 & 32.50 & -43 \\
\midrule\multicolumn{4}{c}{\texttt{Mistral\_8x7B}} & \multicolumn{4}{c}{\texttt{LLAMA3\_3B}} \\
\midrule
ssp & 13.00 & 8.50 & 4.50 & 8 & 15.50 & 9.50 & 6.00 & 7 \\
esp & 10.50 & 6.00 & 4.50 & 3 & 16.00 & 9.50 & 6.50 & 6 \\
sum & 12.50 & 6.00 & 6.50 & -1 & 19.00 & 10.00 & 9.00 & 2 \\
eum & 26.50 & 8.50 & 18.00 & -19 & 48.50 & 8.00 & 40.50 & -65 \\
\midrule\multicolumn{4}{c}{\texttt{LLAMA3\_8B}} & \multicolumn{4}{c}{\texttt{LLAMA3\_70B}} \\
\midrule
ssp & 8.00 & 3.50 & 4.50 & -2 & 10.50 & 5.00 & 5.50 & -1 \\
esp & 7.00 & 2.00 & 5.00 & -6 & 11.50 & 4.50 & 7.00 & -5 \\
sum & 8.50 & 3.00 & 5.50 & -5 & 10.00 & 5.50 & 4.50 & 2 \\
eum & 22.50 & 9.50 & 13.00 & -7 & 7.50 & 2.50 & 5.00 & -5 \\
\bottomrule
\end{tabular}
\caption{Delta metrics on the \texttt{MMLU} benchmark across ten LLMs and four \texttt{DPP}s. For each \texttt{DPP}, we report: (1) the percentage of examples whose predicted answer changed, (2) the percentage that improved, (3) the percentage that regressed, and (4) the Net $\Delta_\text{pred}$ (cnt.) (Total Count Improved – Total Count Regressed), all measured relative to the \textit{sum} configuration.}
\label{tab:mmlu_delta_table}
\end{table*}

\newpage

\clearpage


\subsection{Full Win–Loss–Tie Breakdown by Model}\label{apdx:fwtl}
\begin{figure*}[!httbp]
  \centering
  \begin{subfigure}[b]{0.48\linewidth}
    \includegraphics[width=\linewidth]{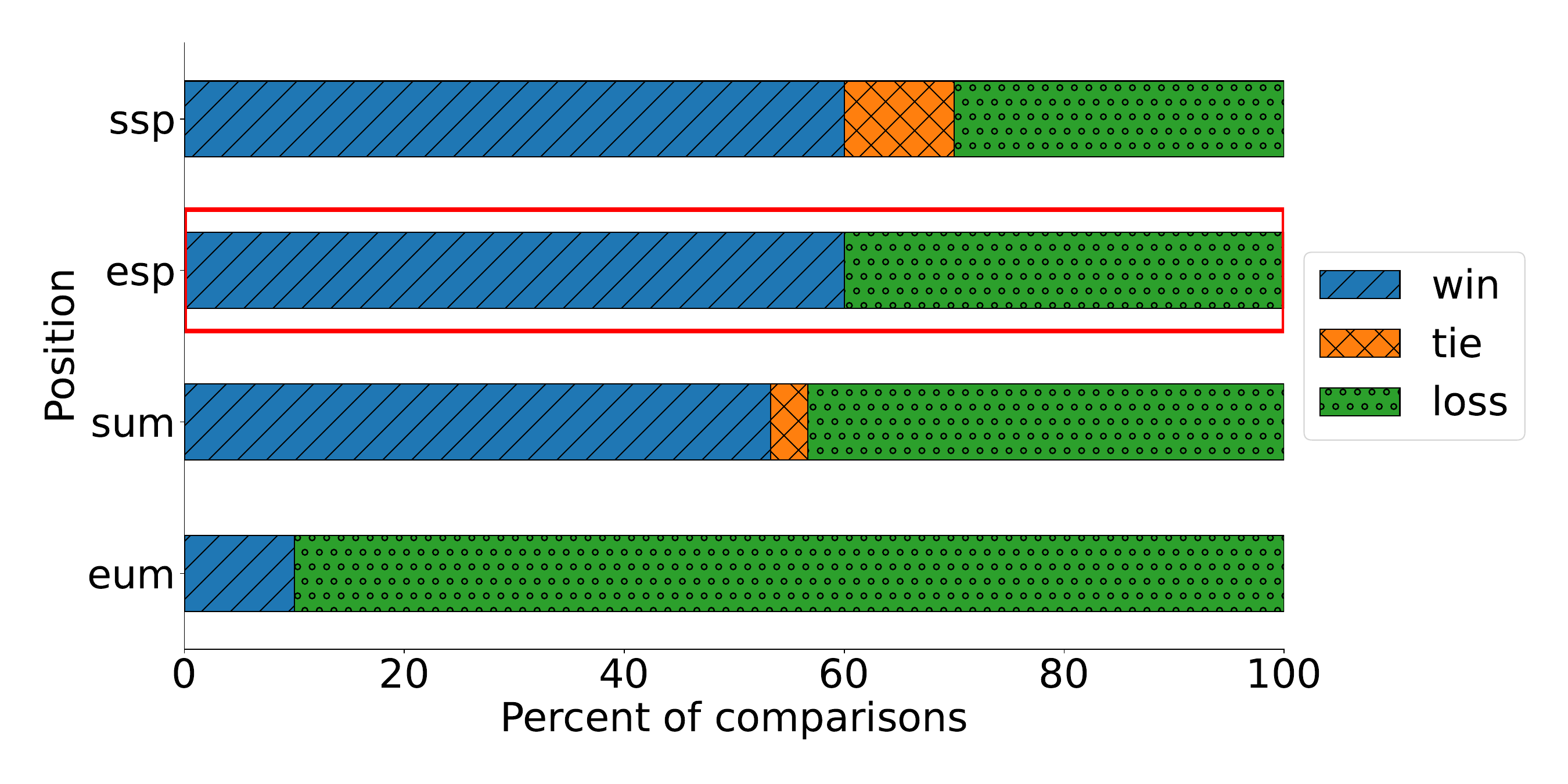}
    \caption{MMLU results.}
    \label{fig:mmluapd}
  \end{subfigure}%
  \hfill
  \begin{subfigure}[b]{0.48\linewidth}
    \includegraphics[width=\linewidth]{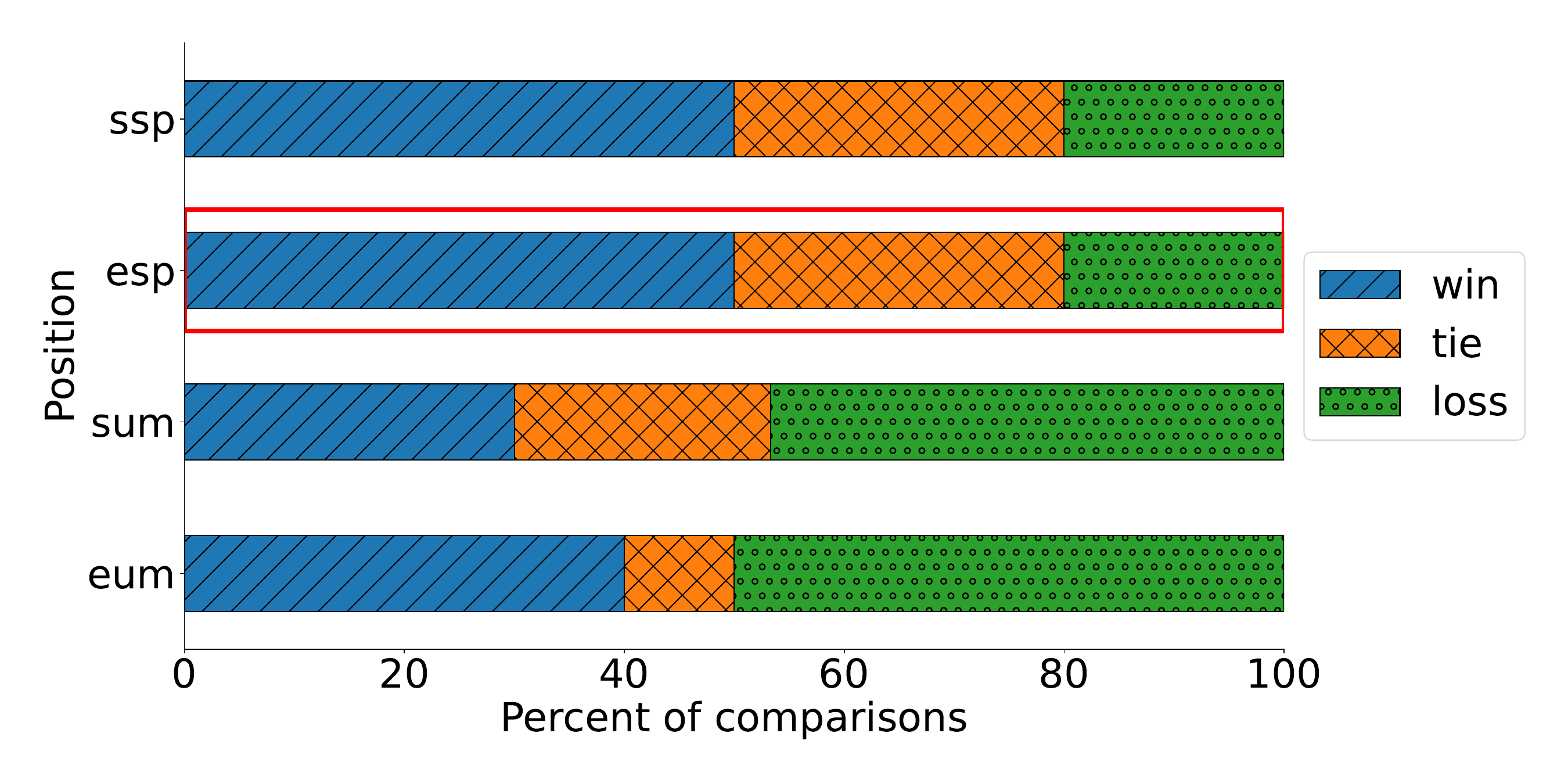}
    \caption{MNLI results.}
    \label{fig:mnliapd}
  \end{subfigure}
  \caption{Win–loss–tie analysis for \textsc{MMLU} and \textsc{MNLI} across all models.}
  \label{fig:models-side-by-side4}
\end{figure*}

\begin{figure*}[!httbp]
  \centering
  \begin{subfigure}[b]{0.48\linewidth}
    \includegraphics[width=\linewidth]{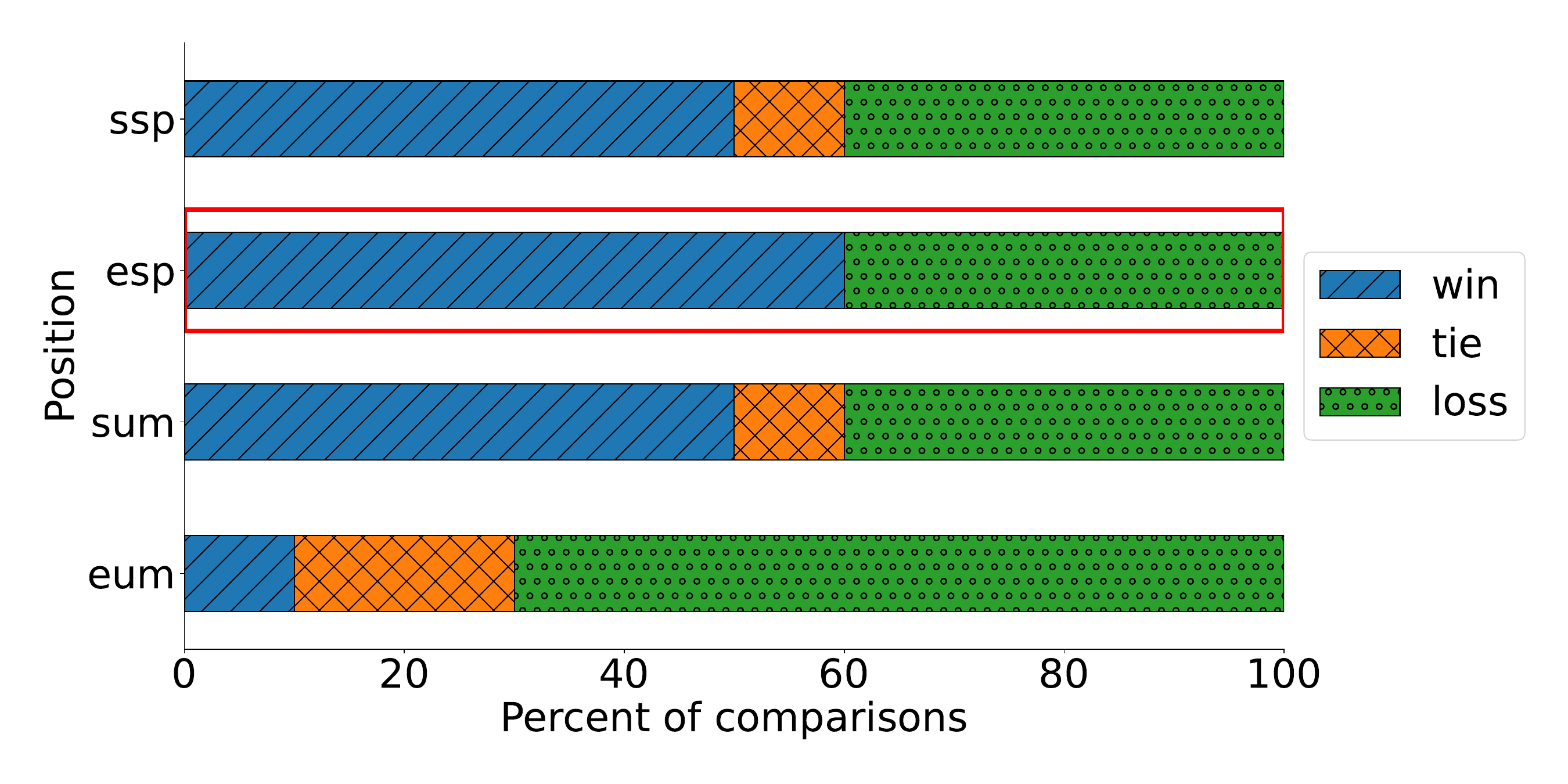}
    \caption{ARC results.}
    \label{fig:arcapd}
  \end{subfigure}%
  \hfill
  \begin{subfigure}[b]{0.48\linewidth}
    \includegraphics[width=\linewidth]{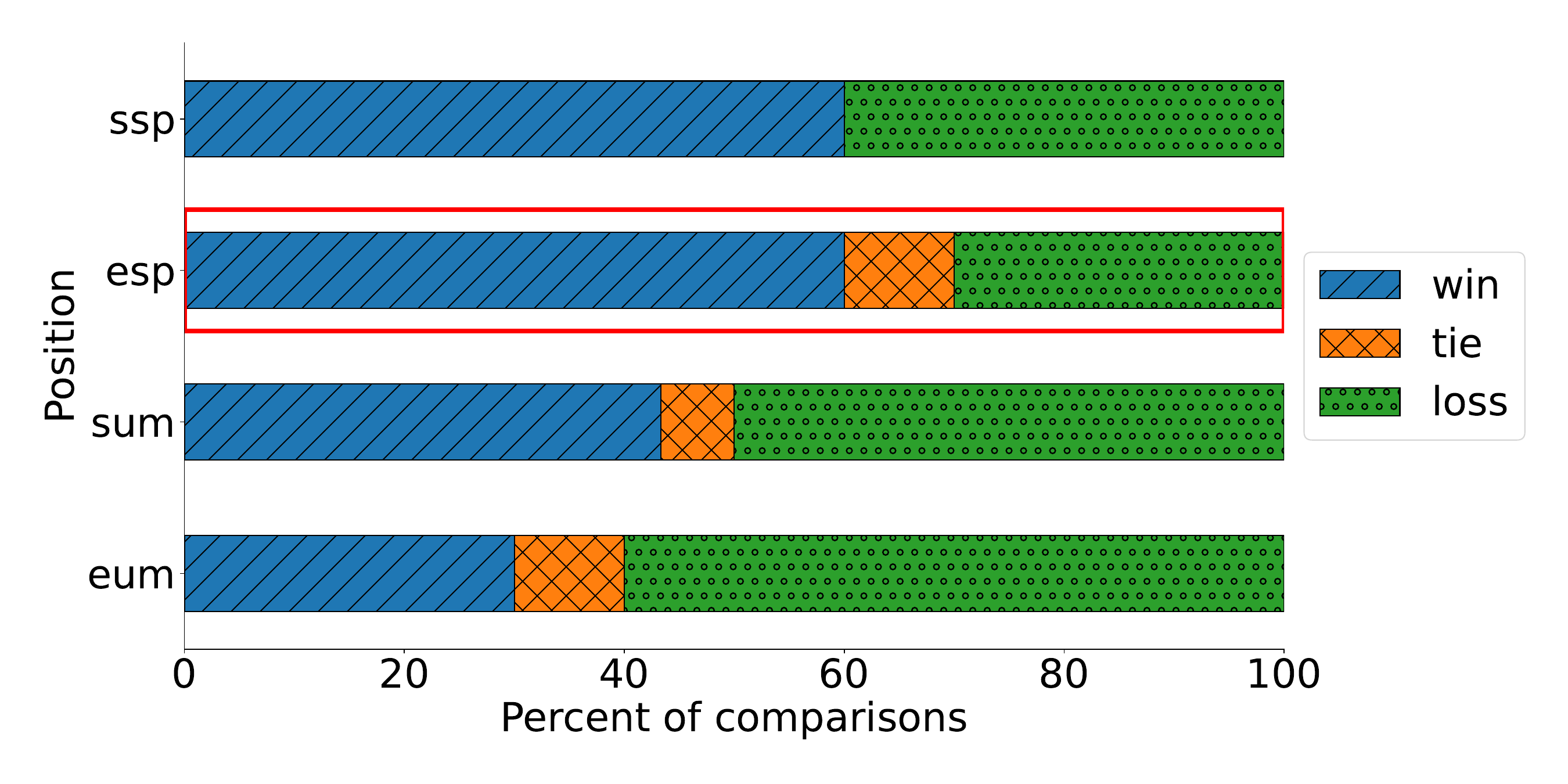}
    \caption{AG News results.}
    \label{fig:agnewsapd}
  \end{subfigure}
  \caption{Win–loss–tie analysis for \textsc{ARC} and \textsc{AG News} across all models.}
  \label{fig:models-side-by-side5}
\end{figure*}

\begin{figure*}[!httbp]
  \centering
  \begin{subfigure}[b]{0.48\linewidth}
    \includegraphics[width=\linewidth]{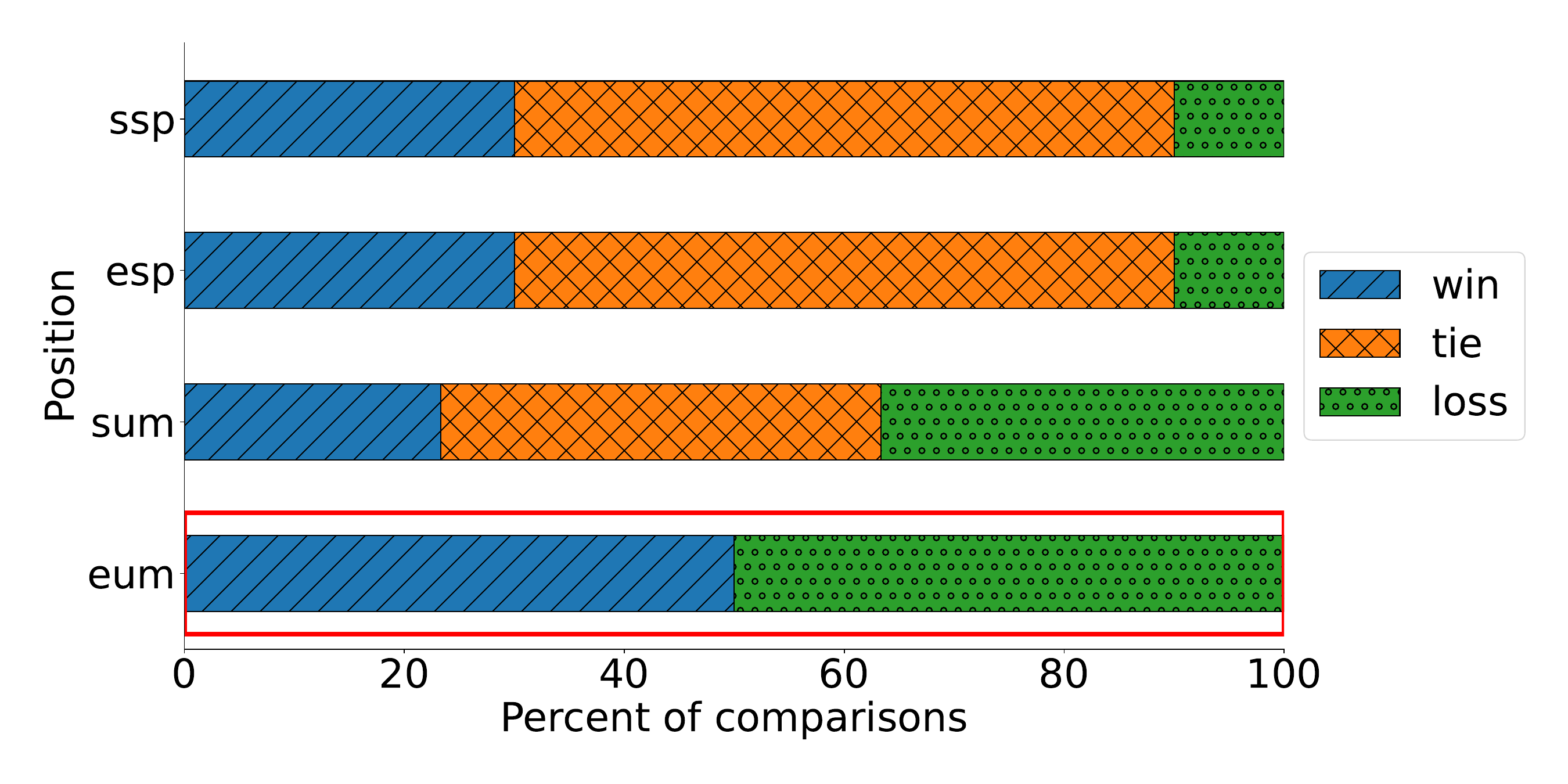}
    \caption{SQuAD results.}
    \label{fig:sqdapd}
  \end{subfigure}%
  \hfill
  \begin{subfigure}[b]{0.48\linewidth}
    \includegraphics[width=\linewidth]{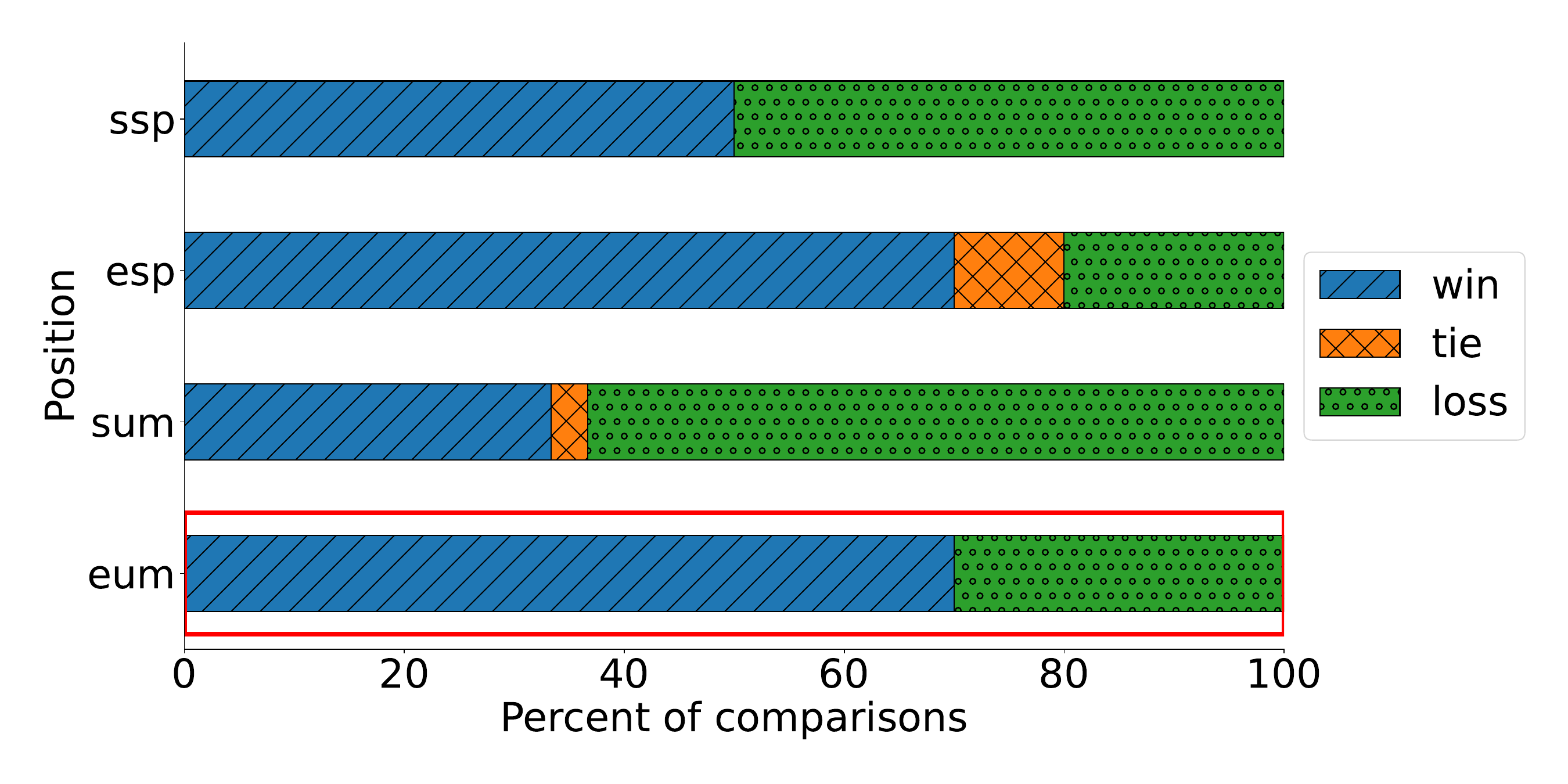}
    \caption{GSM8K results.}
    \label{fig:gsmapd}
  \end{subfigure}
  \caption{Win–loss–tie analysis for \textsc{SQuAD} and \textsc{GSM8K} across all models.}
  \label{fig:models-side-by-side6}
\end{figure*}

\begin{figure*}[!httbp]
  \centering
  \begin{subfigure}[b]{0.48\linewidth}
    \includegraphics[width=\linewidth]{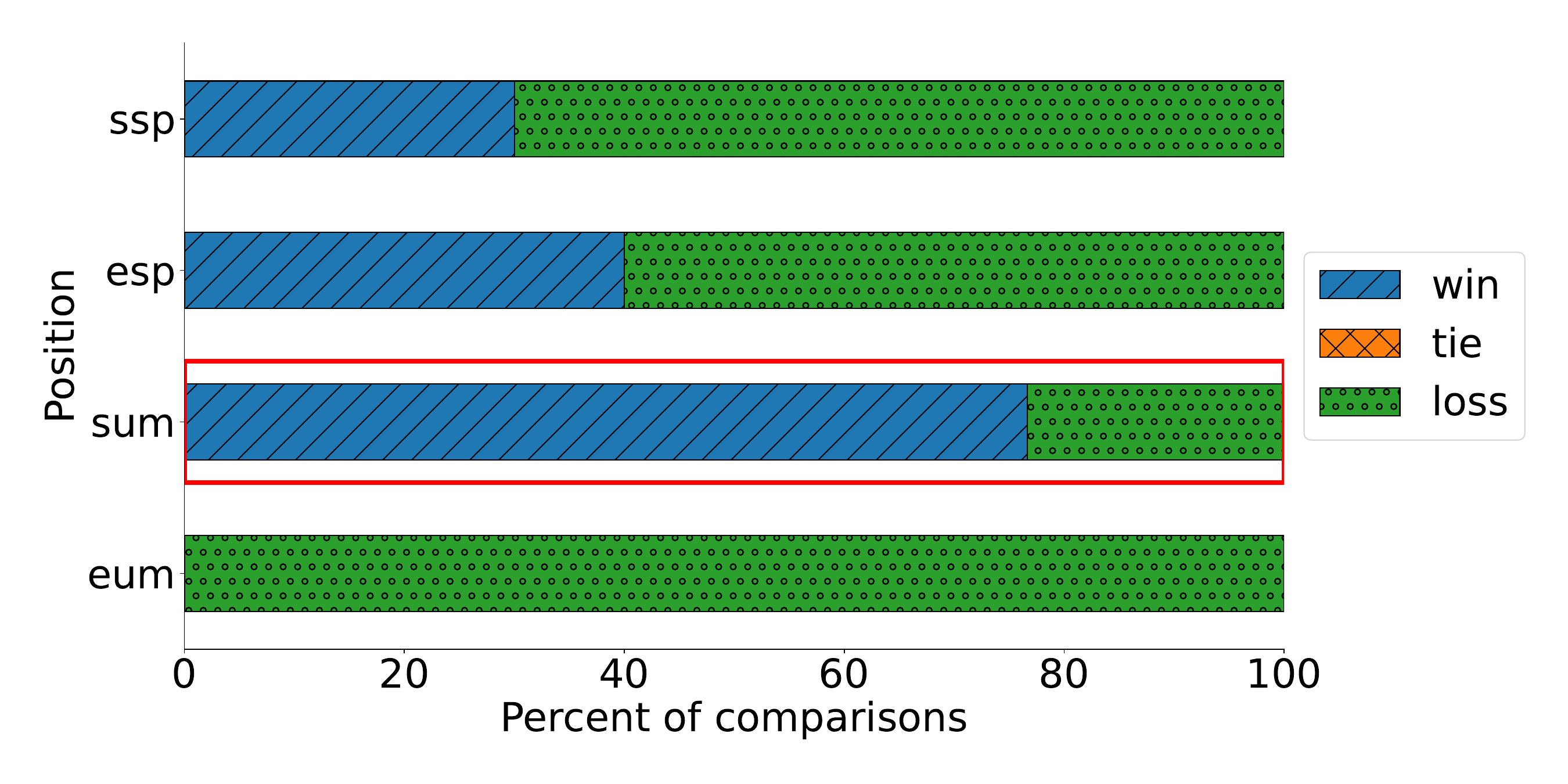}
    \caption{XSum results.}
    \label{fig:xsumapd}
  \end{subfigure}%
  \hfill
  \begin{subfigure}[b]{0.48\linewidth}
    \includegraphics[width=\linewidth]{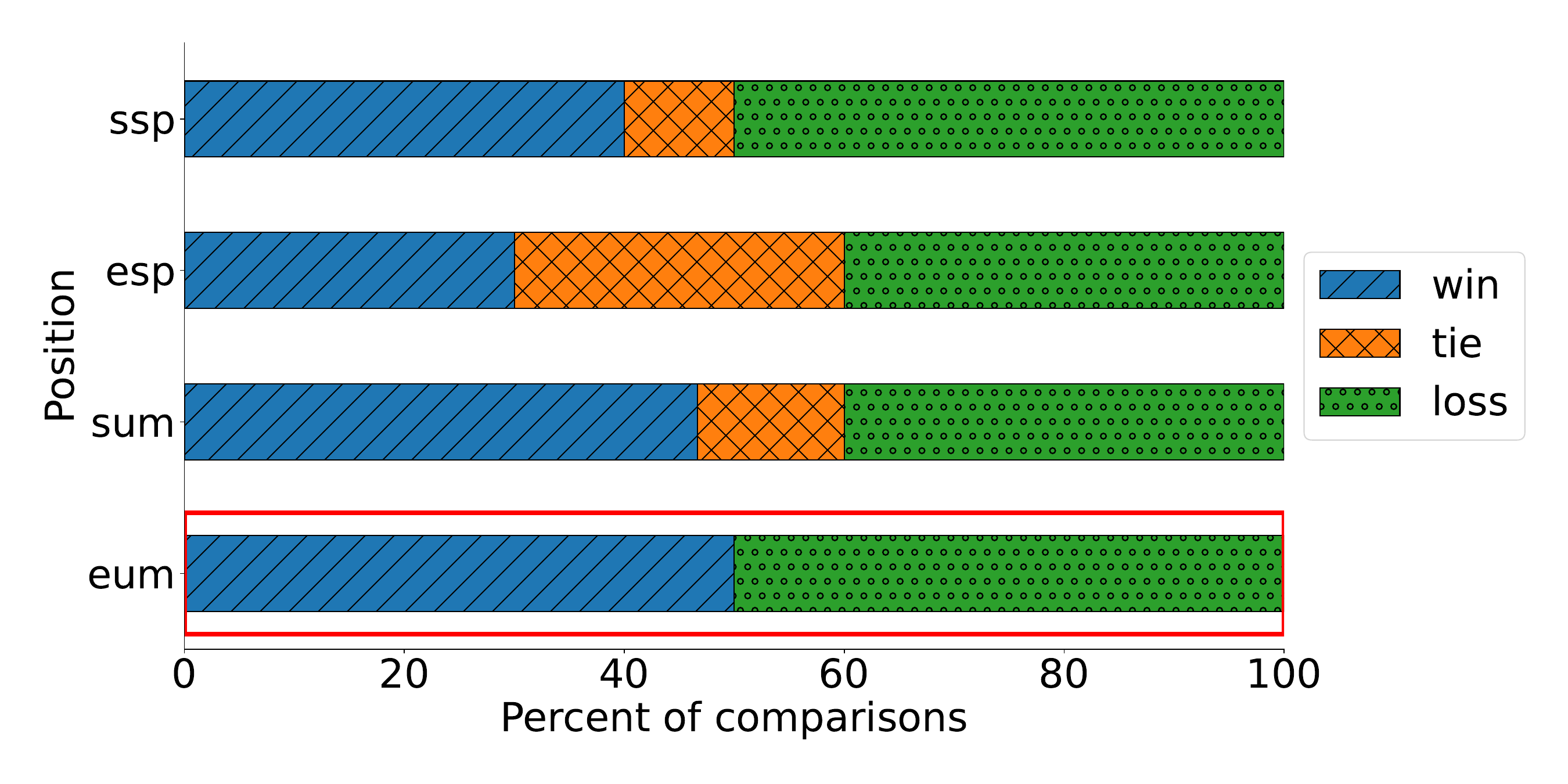}
    \caption{CNN/DailyMail results.}
    \label{fig:cnnapd}
  \end{subfigure}
  \caption{Win–loss–tie analysis for \textsc{XSum} and \textsc{CNN/DailyMail} across all models.}
  \label{fig:models-side-by-side7}
\end{figure*}

\paragraph{Task-Centric Analysis.}
To complement the model-centric win–loss breakdowns discussed above, we provide a task-centric perspective here. Figures~\ref{fig:models-side-by-side4} through~\ref{fig:models-side-by-side7} illustrate how frequently each demonstration position emerges as the best (or worst) across models for individual tasks. These visualizations confirm that no single position consistently dominates across tasks: while \texttt{ssp} often performs best on classification tasks like \textsc{MNLI} and \textsc{AG News}, positions like \texttt{esp} or \texttt{sum} sometimes outperform on reasoning or summarization tasks. This highlights the need for prompt position tuning tailored not just to model size but also to the task domain.

\begin{figure*}[!httbp]
  \centering
  \begin{subfigure}[b]{0.48\linewidth}
    \includegraphics[width=\linewidth]{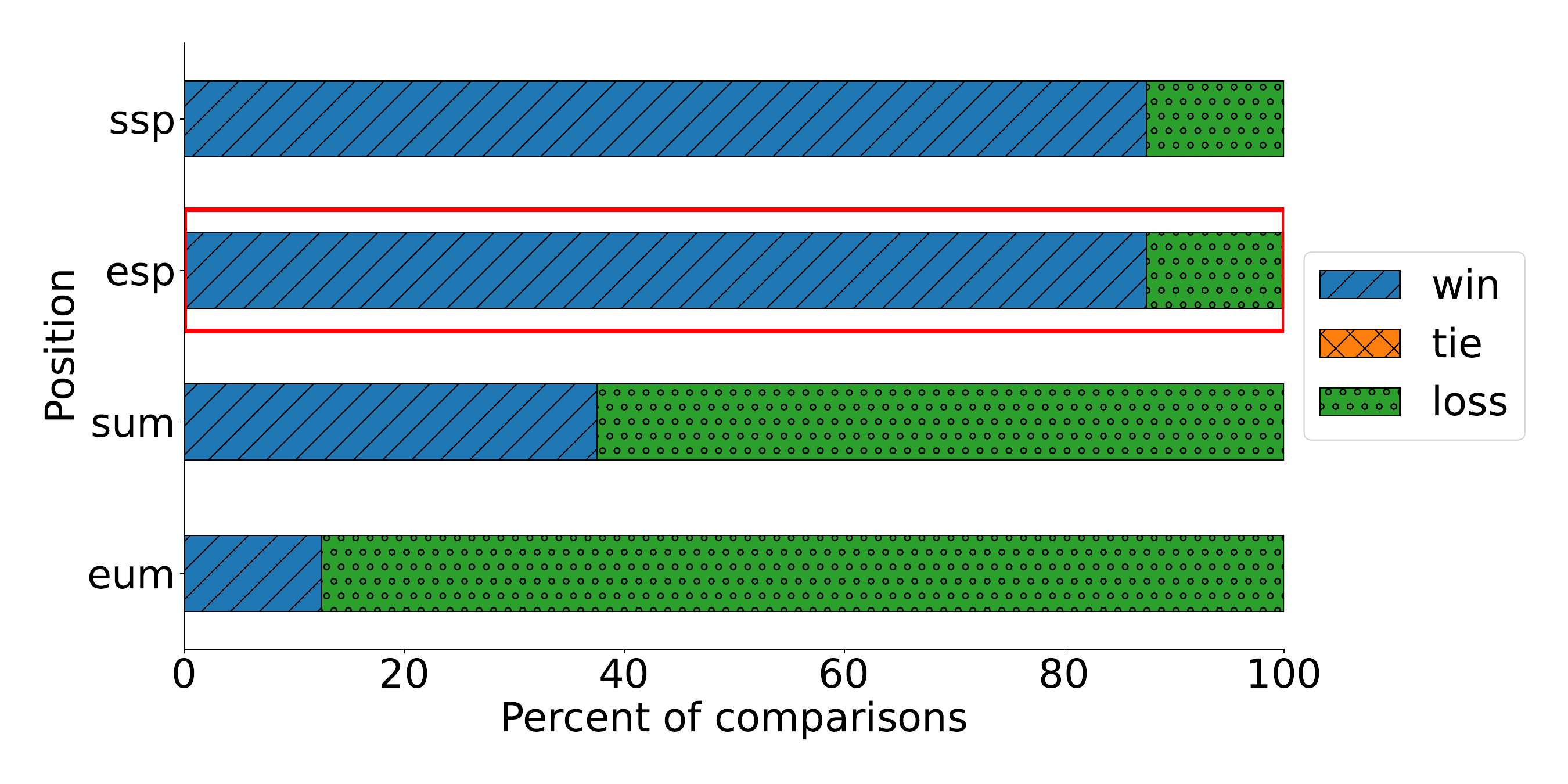}
    \caption{LLAMA3-3B results.}
    \label{fig:llama3-3b}
  \end{subfigure}%
  \hfill
  \begin{subfigure}[b]{0.48\linewidth}
    \includegraphics[width=\linewidth]{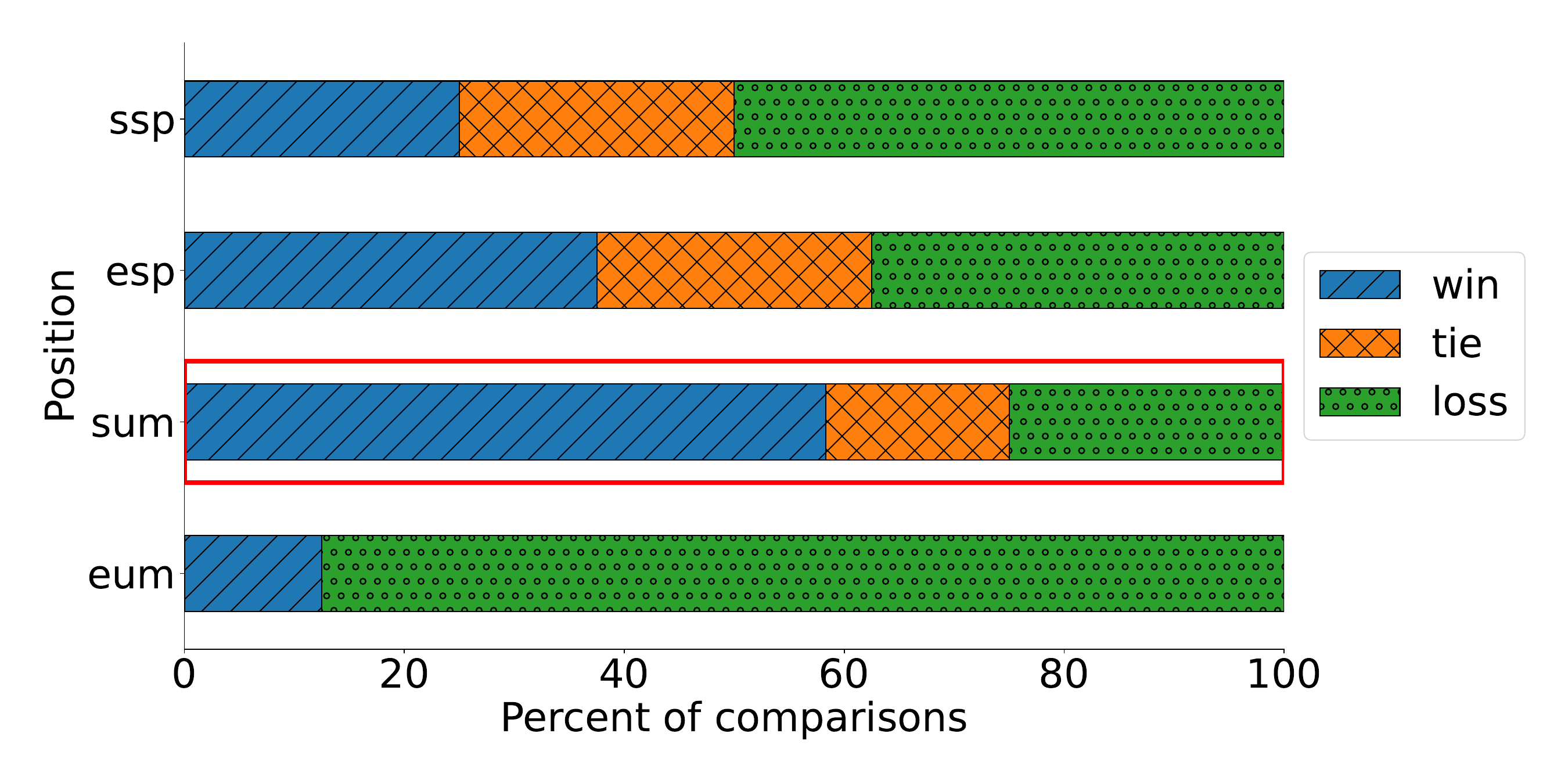}
    \caption{Qwen-7B results.}
    \label{fig:qwen-7b}
  \end{subfigure}
  \caption{Win–loss–tie analysis for \textsc{LLAMA3-3B} and \textsc{Qwen-7B} across all tasks}
  \label{fig:models-side-by-side1}
\end{figure*}



\begin{figure*}[!httbp]
  \centering
  \begin{subfigure}[b]{0.48\linewidth}
    \includegraphics[width=\linewidth]{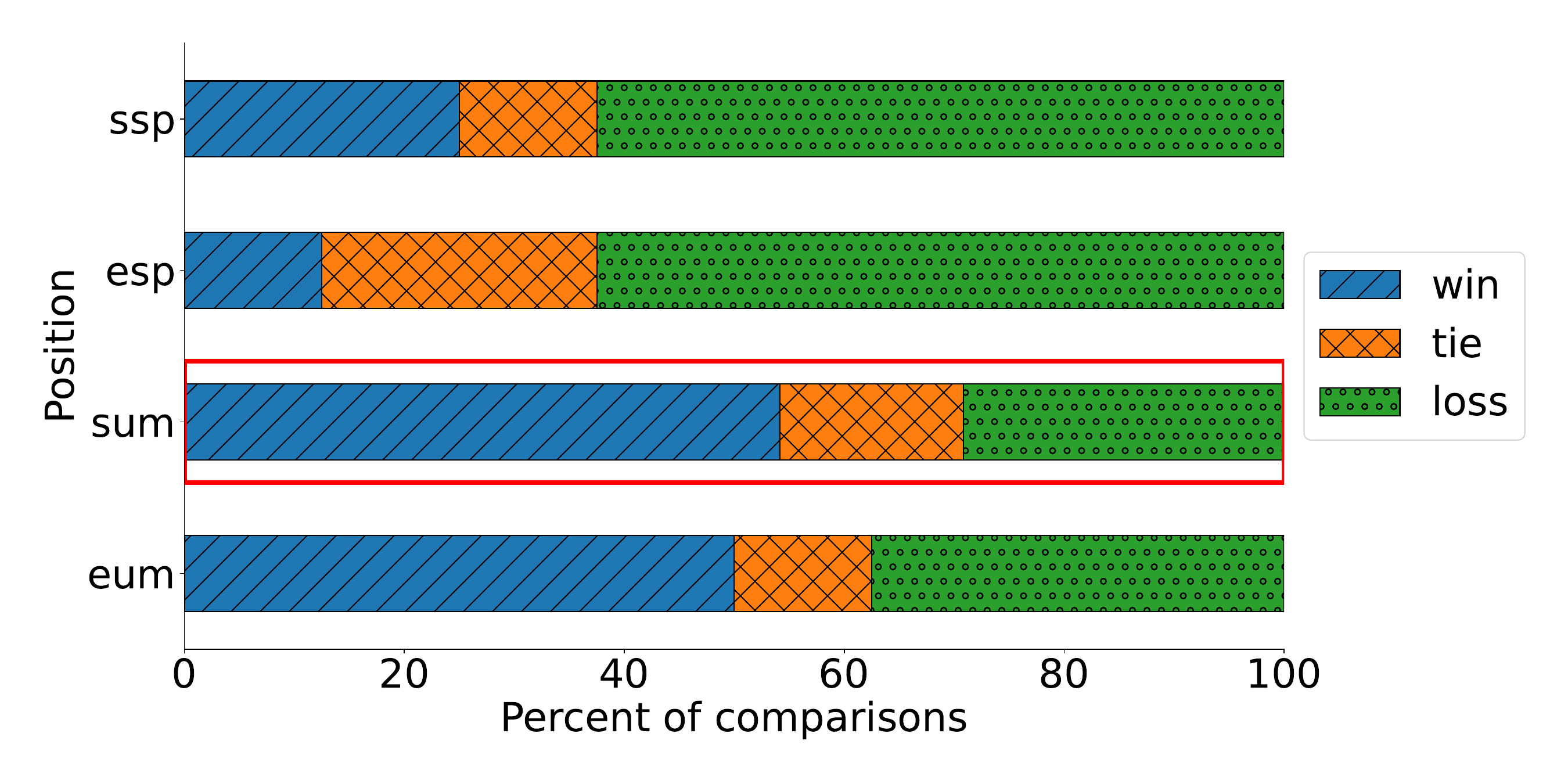}
    \caption{Qwen-72B results.}
    \label{fig:Qwen_72B}
  \end{subfigure}%
  \hfill
  \begin{subfigure}[b]{0.48\linewidth}
    \includegraphics[width=\linewidth]{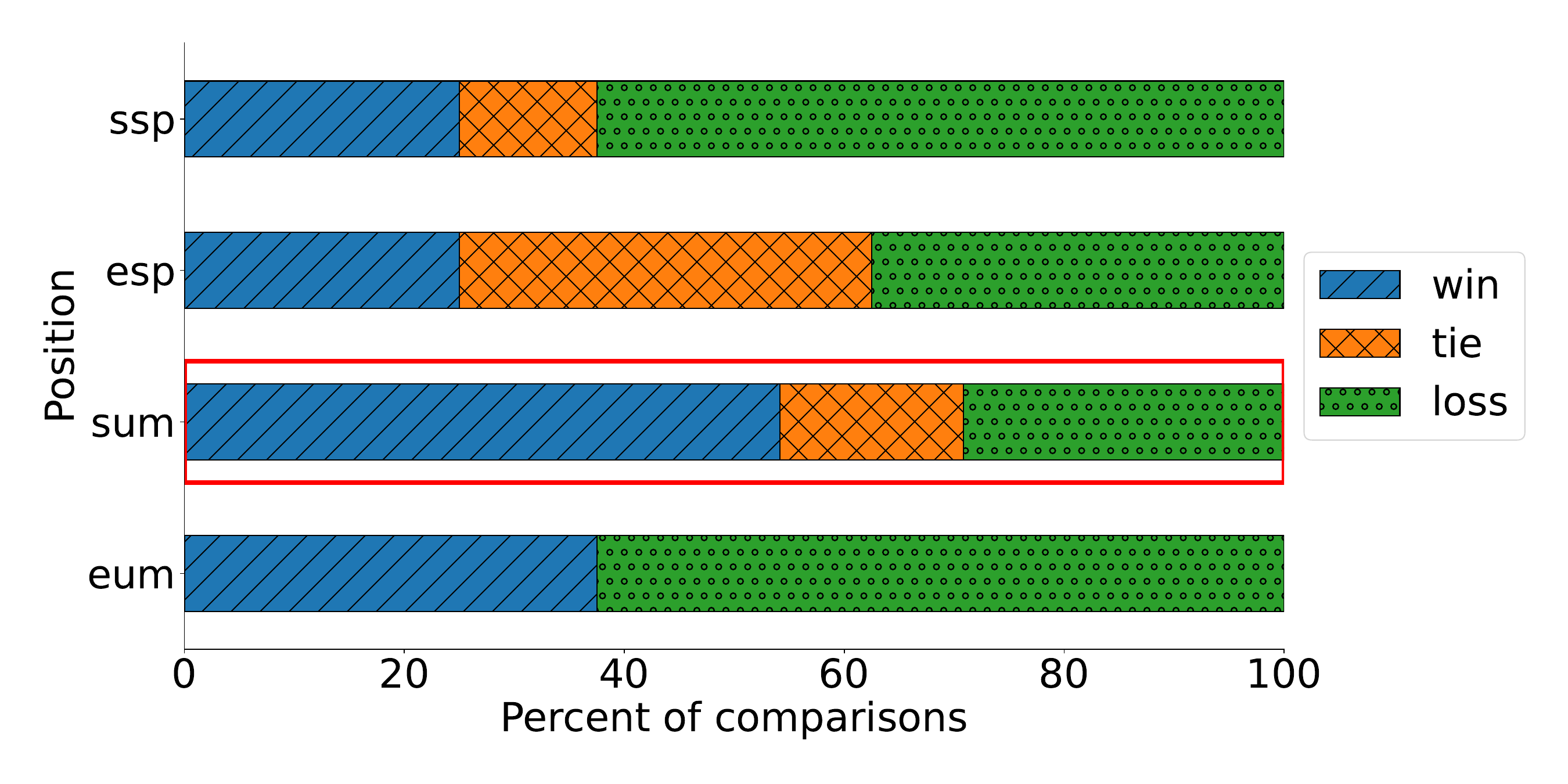}
    \caption{Mistral-7B results.}
    \label{fig:mist-7b}
  \end{subfigure}
  \caption{Win–loss–tie analysis for \textsc{Qwen-72B} and \textsc{Mistral-7B} across all tasks}
  \label{fig:models-side-by-side2}
\end{figure*}



\begin{figure*}[!httbp]
  \centering
  \begin{subfigure}[b]{0.48\linewidth}
    \includegraphics[width=\linewidth]{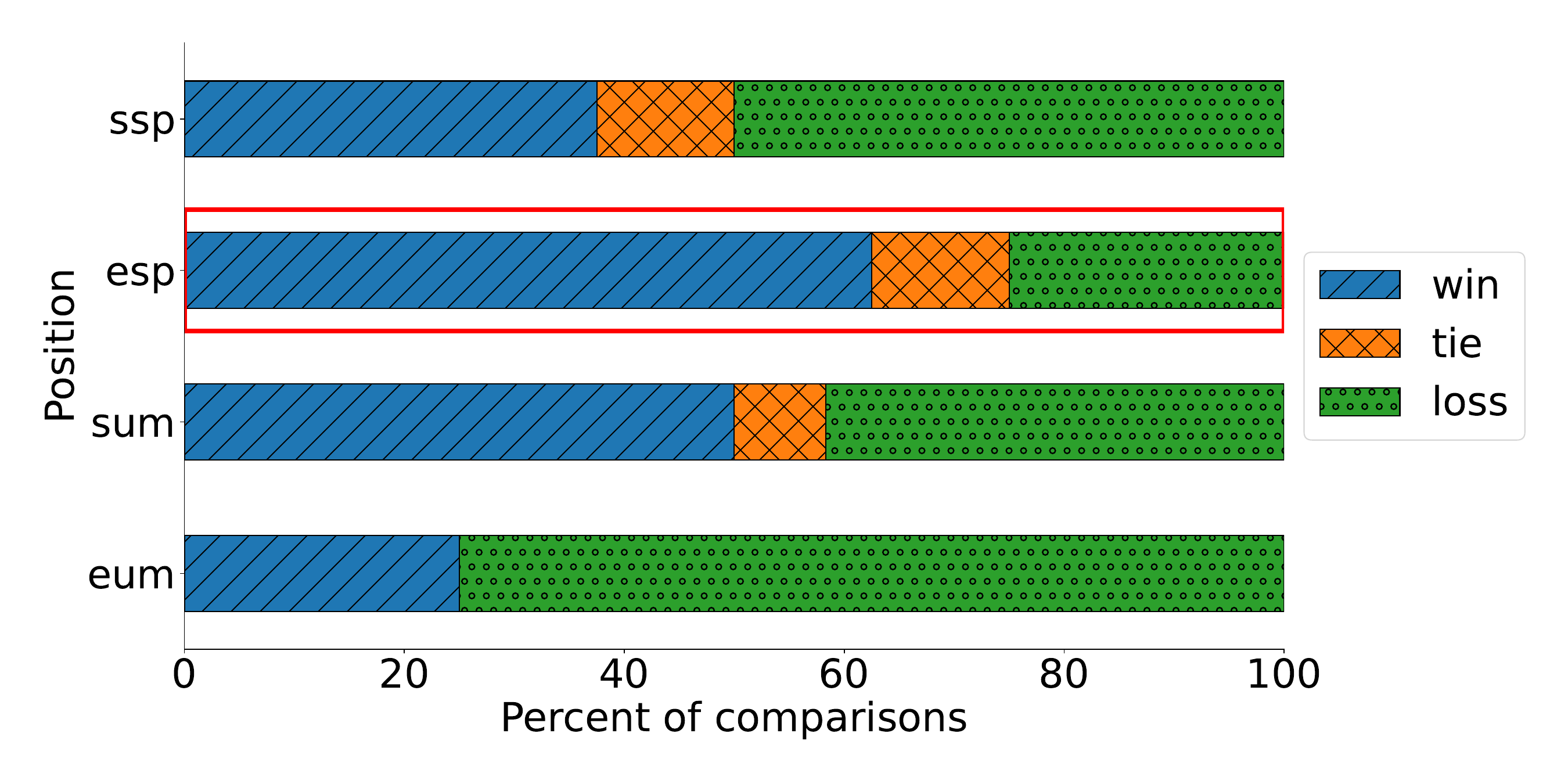}
    \caption{Mistral-8x7B results.}
    \label{fig:mist8x7}
  \end{subfigure}%
  \hfill
  \begin{subfigure}[b]{0.48\linewidth}
    \includegraphics[width=\linewidth]{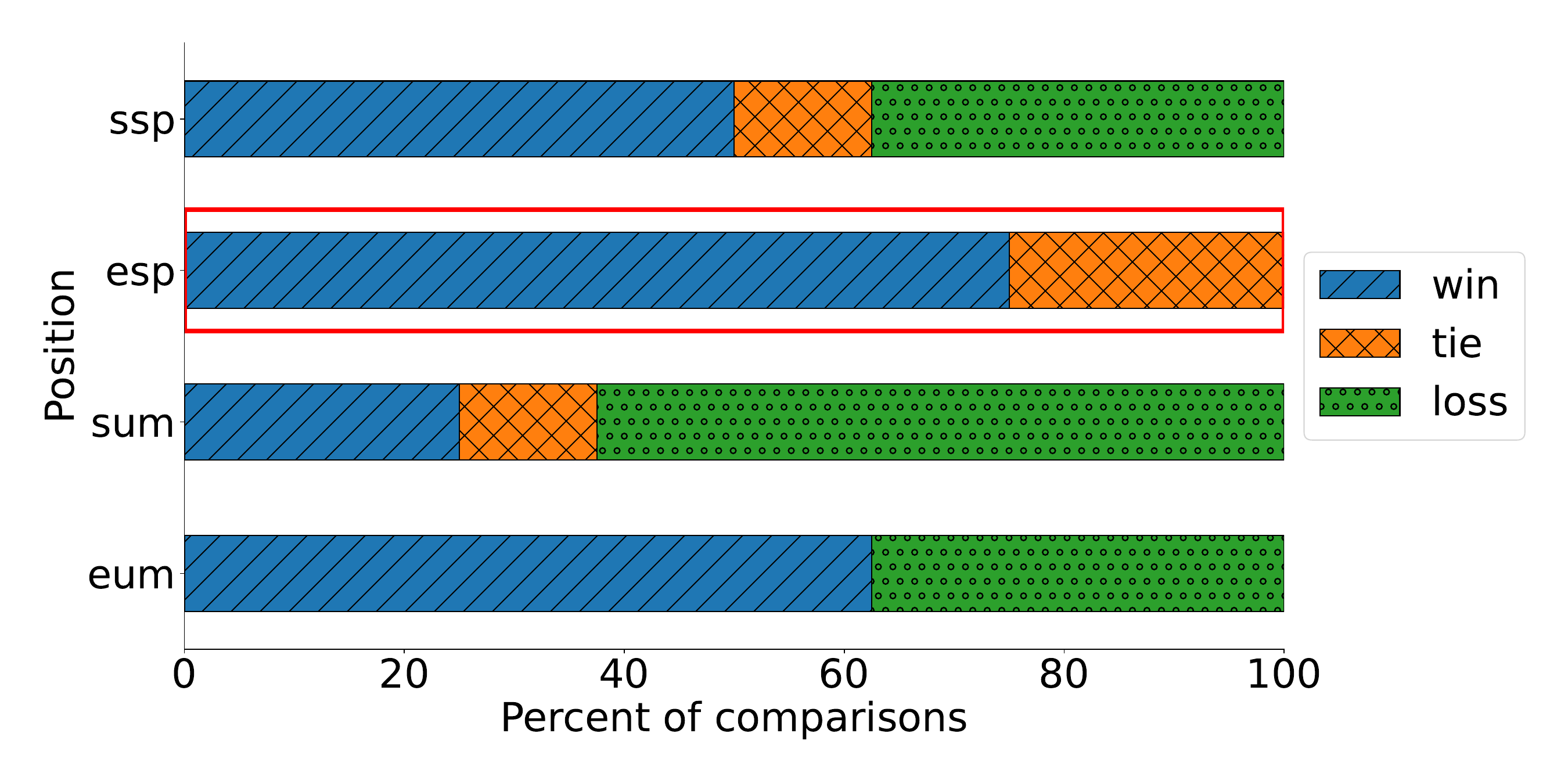}
    \caption{Cohere-32B results.}
    \label{fig:coh32b}
  \end{subfigure}
  \caption{Win–loss–tie analysis for \textsc{Mistral-8x7B} and \textsc{Cohere-32B} across all tasks}
  \label{fig:models-side-by-side3}
\end{figure*}





\clearpage
\subsection{Data Sampling}
For each benchmark we first sample 200 test examples (without replacement) from the official test split, using five different random seeds (42, 123, 456, 789, 1).  We also sample 5 in-context demonstration examples (without replacement) from the train split for each seed as our \texttt{DPP} set.  

\newpage
\subsection{Dataset-wise View of Prediction Volatility and Accuracy Gains}
To complement the main analysis, we present detailed per-task plots visualizing how different ICL \texttt{DPP's} affect model behavior across scale. For each dataset (\texttt{AG News, ARC, GSM8K, SQUAD and MNLI}), we report \textbf{(i)} the percentage of prediction changes when we switch from the default \texttt{DPP} (\texttt{sum}) to alternative  \texttt{DPP's} (\texttt{ssp}, \texttt{esp} and \texttt{eum}), and \textbf{(ii)} the signed percentage metric gain or loss when compared to the 0-shot baseline. 

\begin{figure*}[!httbp]
    \centering
    \includegraphics[width=\textwidth]{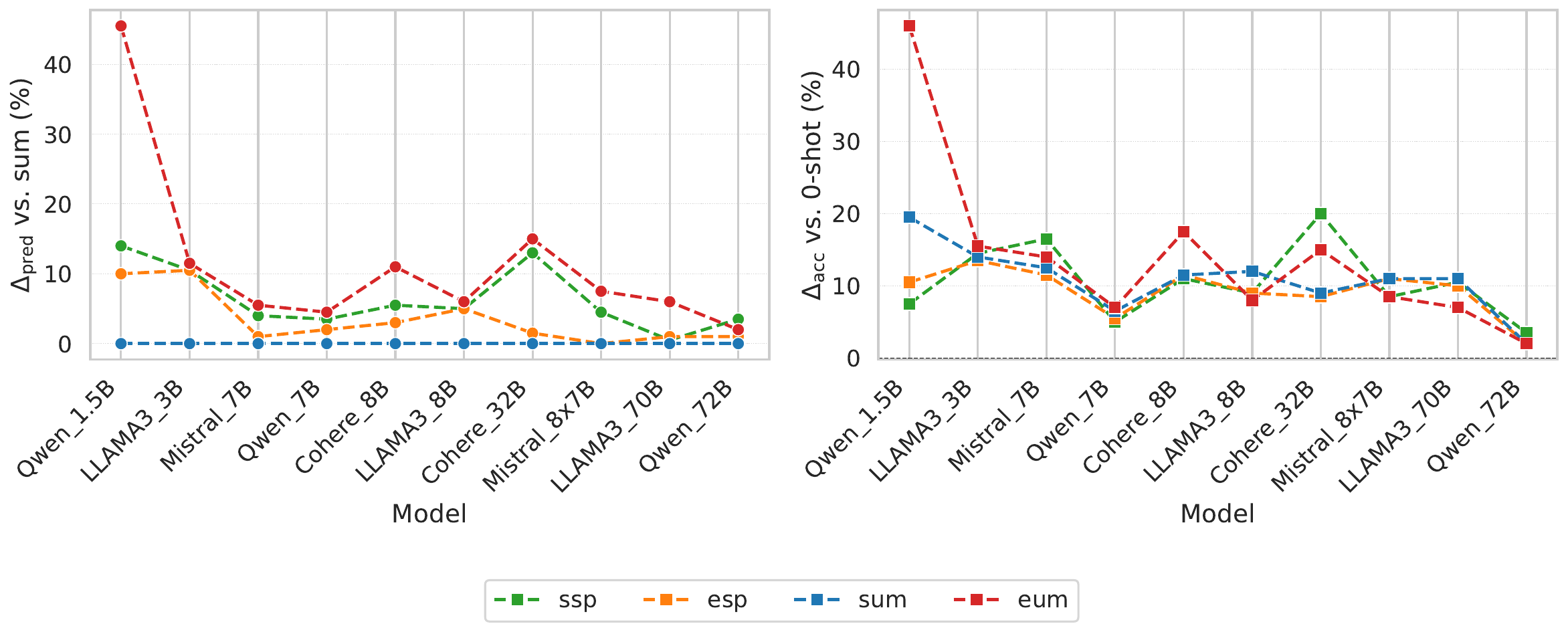}
    \caption{(\texttt{AG News}) \textbf{Left:} the percentage of predictions that change when switching from \texttt{sum} to other positions. \textbf{Right:} Accuracy change over the zero-shot baseline.}
\end{figure*}

\begin{figure*}[!httbp]
    \centering
    \includegraphics[width=\textwidth]{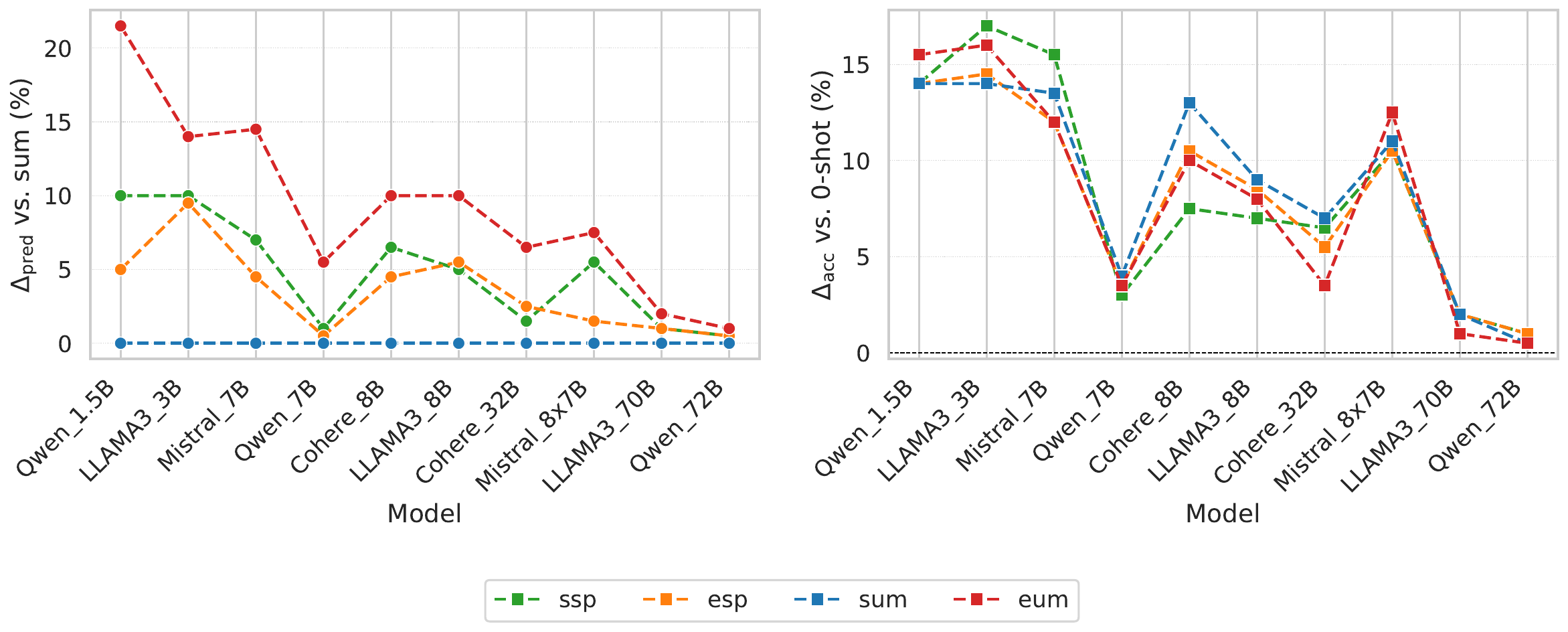}
    \caption{(\texttt{ARC}) \textbf{Left:} the percentage of predictions that change when switching from \texttt{sum} to other positions. \textbf{Right:} Accuracy change over the zero-shot baseline.}
\end{figure*}

\begin{figure*}[!httbp]
    \centering
    \includegraphics[width=\textwidth]{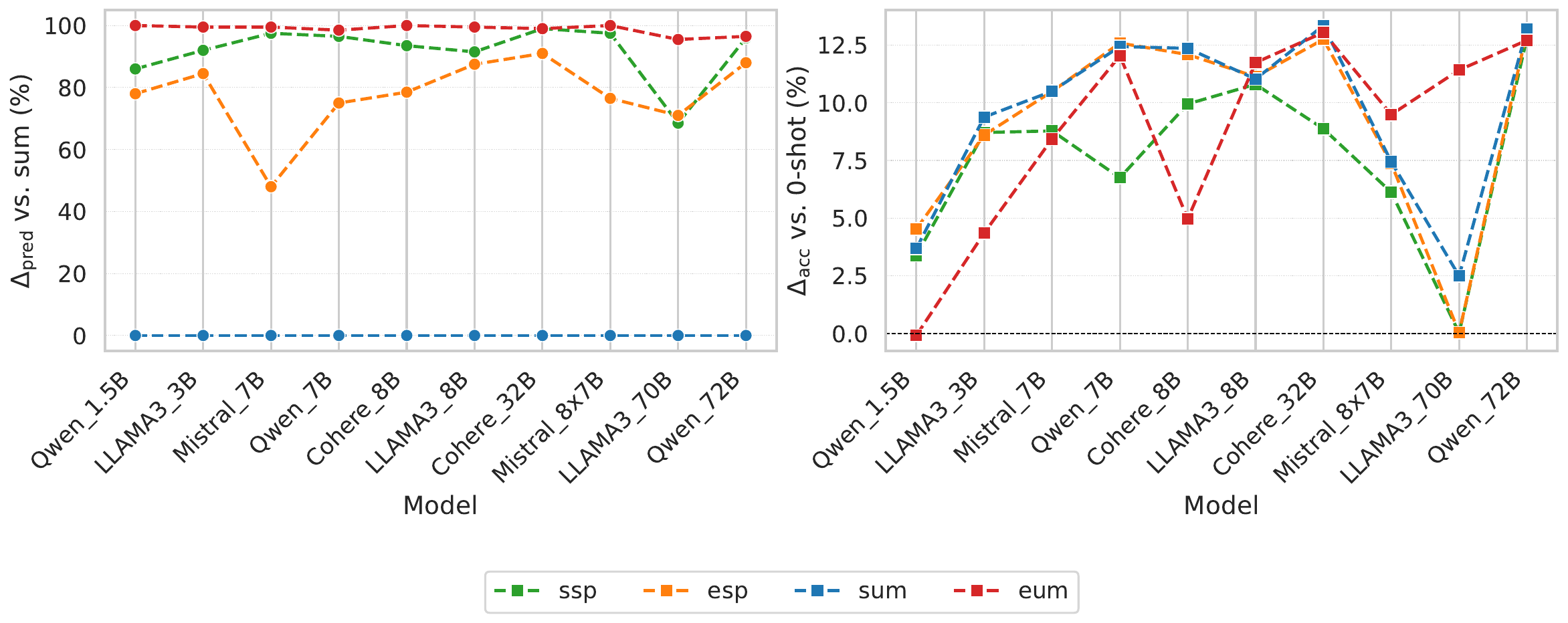}
    \caption{(\texttt{GSM8K}) \textbf{Left:} the percentage of predictions that change when switching from \texttt{sum} to other positions. \textbf{Right:} Accuracy change over the zero-shot baseline.}
\end{figure*}

\begin{figure*}[!httbp]
    \centering
    \includegraphics[width=\textwidth]{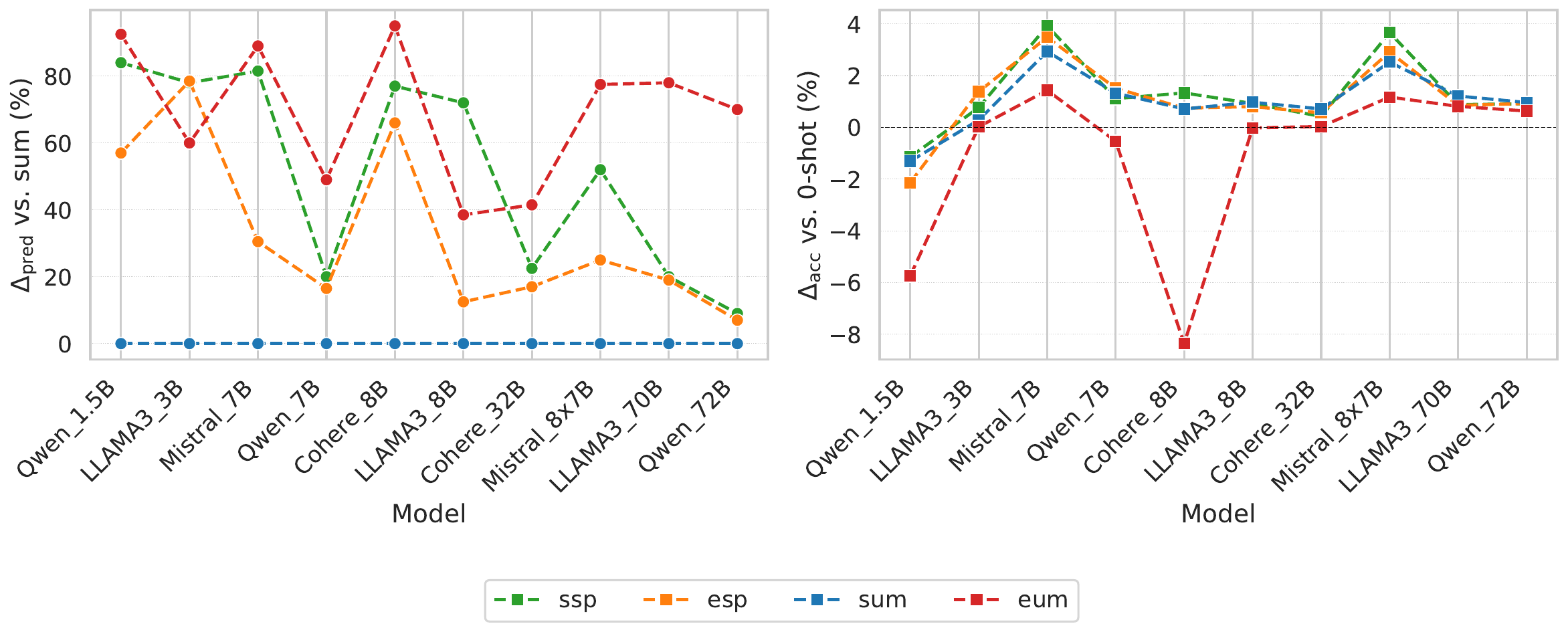}
    \caption{(\texttt{SQuAD}) \textbf{Left:} the percentage of predictions that change when switching from \texttt{sum} to other positions. \textbf{Right:} Accuracy change over the zero-shot baseline.}
\end{figure*}

\begin{figure*}[!httbp]
    \centering
    \includegraphics[width=\textwidth]{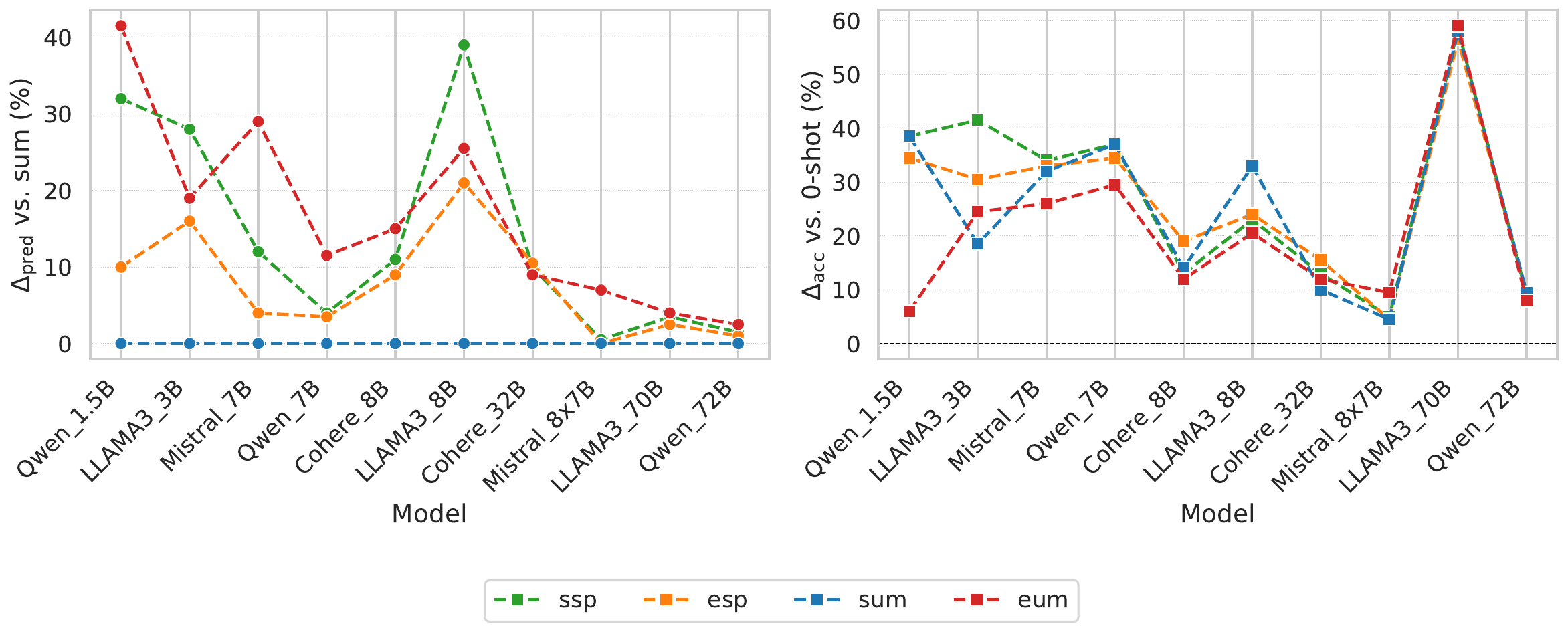}
    \caption{(\texttt{MNLI}) \textbf{Left:} the percentage of predictions that change when switching from \texttt{sum} to other positions. \textbf{Right:} Accuracy change over the zero-shot baseline.}
\end{figure*}

\begin{figure*}[!httbp]
    \centering
    \includegraphics[width=\textwidth]{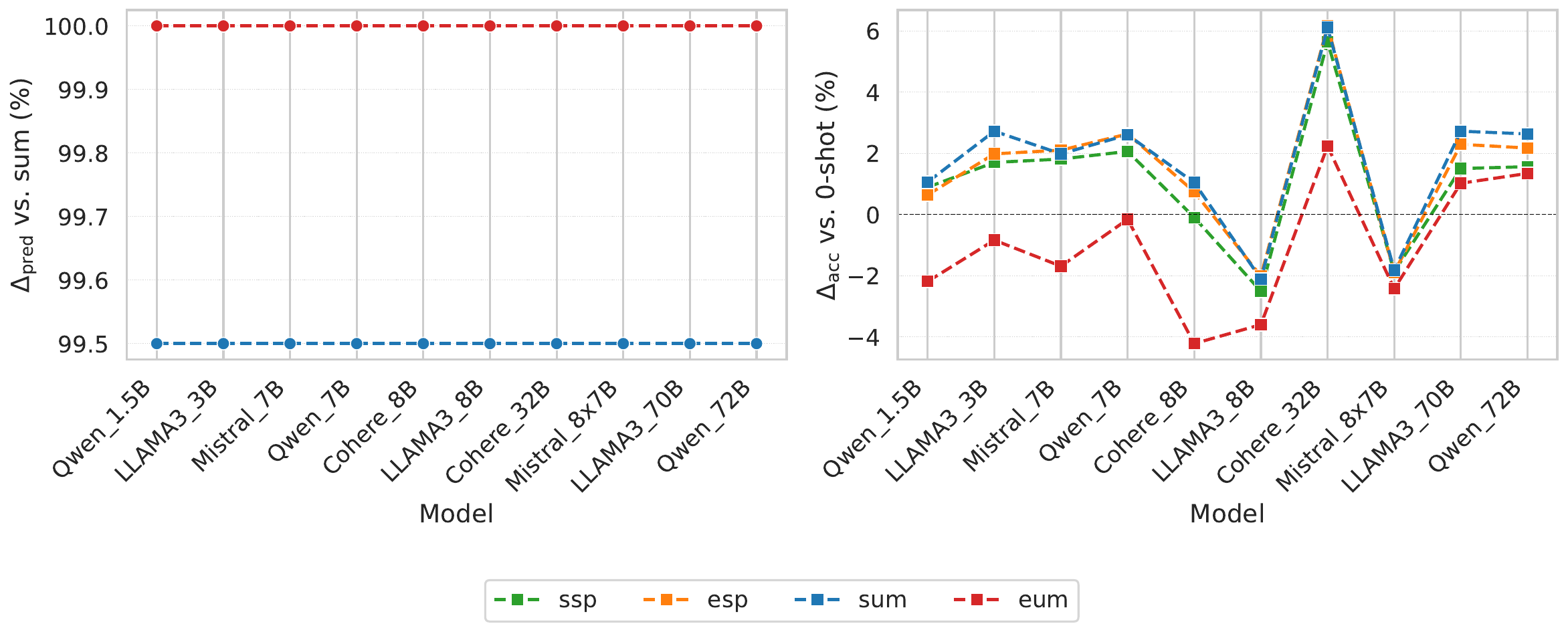}
    \caption{(\texttt{XSUM}) \textbf{Left:} the percentage of predictions that change when switching from \texttt{sum} to other positions. \textbf{Right:} Accuracy change over the zero-shot baseline.}
\end{figure*}


\end{document}